\RequirePackage[l2tabu]{nag}
\documentclass[a4paper,12pt,leqno,openbib]{memoir} 
\usepackage{datetime}
\usepackage{ifpdf}
\ifpdf
\fi
\ifdraftdoc 
	\usepackage{draftwatermark}				
	\SetWatermarkScale{0.3}
	\SetWatermarkText{\bf Draft: \today}
\fi
\newsubfloat{figure}
\newsubfloat{table}
\settrimmedsize{297mm}{210mm}{*}
\settypeblocksize{634pt}{448.13pt}{*} 
\setulmargins{4cm}{*}{*} 
\setlrmargins{*}{*}{1.5} 
\setmarginnotes{17pt}{51pt}{\onelineskip} 
\setheadfoot{\onelineskip}{2\onelineskip} 
\setheaderspaces{*}{2\onelineskip}{*} 
\checkandfixthelayout
\usepackage{fouriernc}
\usepackage{hhline}
\usepackage[T1]{fontenc}
\OnehalfSpacing 
\setsecnumdepth{subsection} 
\maxsecnumdepth{subsubsection}
\usepackage{calc,soul,fourier}
\makeatletter 
\newlength\dlf@normtxtw 
\newsavebox{\feline@chapter} 
\newcommand\feline@chapter@marker[1][4cm]{%
	\sbox\feline@chapter{%
		\resizebox{!}{#1}{\fboxsep=1pt%
			\colorbox{gray}{\color{white}\thechapter}%
		}}%
		\rotatebox{90}{%
			\resizebox{%
				\heightof{\usebox{\feline@chapter}}+\depthof{\usebox{\feline@chapter}}}%
			{!}{\scshape\so\@chapapp}}\quad%
		\raisebox{\depthof{\usebox{\feline@chapter}}}{\usebox{\feline@chapter}}%
} 
\newcommand\feline@chm[1][4cm]{%
	\sbox\feline@chapter{\feline@chapter@marker[#1]}%
	\makebox[0pt][c]{
		\makebox[1cm][r]{\usebox\feline@chapter}%
	}}
\makechapterstyle{daleifmodif}{

	\renewcommand\printchapternum{\null\hfill\feline@chm[2.5cm]\par}

} 

\chapterstyle{daleifmodif}
\makepagestyle{myvf} 
\makeoddfoot{myvf}{}{\thepage}{} 
\makeevenfoot{myvf}{}{\thepage}{} 
\makeheadrule{myvf}{\textwidth}{\normalrulethickness} 
\makeevenhead{myvf}{\small\textsc{\leftmark}}{}{} 
\makeoddhead{myvf}{}{}{\small\textsc{\rightmark}}
\newcommand{\clearemptydoublepage}{\newpage{\thispagestyle{empty}\cleardoublepage}}

\renewcommand\cleardoublepage{%
 \clearpage
 \ifodd\value{page}\else\stepcounter{page}\fi
}

\usepackage{afterpage}

\makeindex
\usepackage{import}
\usepackage{bm}
\usepackage[thinlines]{easytable}
\usepackage{amsmath, amssymb} 
\usepackage{lipsum}					
\usepackage{amsfonts} 					
\usepackage{stmaryrd}					
\usepackage{amssymb}					
\usepackage{amsthm}					
\usepackage{newlfont}					
\usepackage{layouts}					
\usepackage{graphics}					
\usepackage{longtable,rotating}			
\usepackage[applemac]{inputenc}			
\usepackage{colortbl}					

\usepackage{float}						
\usepackage{verbatim}					
\usepackage{upgreek }					
\usepackage{setspace}
\usepackage[noend]{algpseudocode}
\usepackage{algorithm} 
\usepackage[algo2e]{algorithm2e} 
\usepackage{latexsym}					
\usepackage{url}						
\usepackage{etex}						
\usepackage{fixltx2e}					
\usepackage[spanish,english]{babel}		
\usepackage{color}                    				
\usepackage{memhfixc}					
\usepackage{enumerate}					
\usepackage{footnote}					
\usepackage{microtype}					
\usepackage{rotfloat}					
\usepackage{alltt}						
\usepackage[version=0.96]{pgf}			
\usepackage{tikz}						
\usetikzlibrary{arrows,shapes,snakes,
		       automata,backgrounds,
		       petri,topaths}				
\newcommand{\pgftextcircled}[1]{                                                                    
    \setbox0=\hbox{#1}%
    \dimen0\wd0%
    \divide\dimen0 by 2%
    \begin{tikzpicture}[baseline=(a.base)]%
        \useasboundingbox (-\the\dimen0,0pt) rectangle (\the\dimen0,1pt);
        \node[circle,draw,outer sep=0pt,inner sep=0.1ex] (a) {#1};
    \end{tikzpicture}
}
\newcommand{\blackged}{\hfill$\blacksquare$}
\newcommand{\whiteged}{\hfill$\square$}
\newcounter{proofcount}

\let\oldsqrt\sqrt
\def\sqrt{\mathpalette\DHLhksqrt}
\def\DHLhksqrt#1#2{%
\setbox0=\hbox{$#1\oldsqrt{#2\,}$}\dimen0=\ht0
\advance\dimen0-0.2\ht0
\setbox2=\hbox{\vrule height\ht0 depth -\dimen0}%
{\box0\lower0.4pt\box2}}
\newcommand{\mycaption}[2][\@empty]{
	\captionnamefont{\scshape} 
	\changecaptionwidth
	\captionwidth{0.9\linewidth}
	\captiondelim{.\:} 
	\indentcaption{0.75cm}
	\captionstyle[\centering]{}
	\setlength{\belowcaptionskip}{10pt}
	\ifx \@empty#1 \caption{#2}\else \caption[#1]{#2}
}
\newcommand{\mysubcaption}[2][\@empty]{
	\subcaptionsize{\small}
	\hangsubcaption
	\subcaptionlabelfont{\rmfamily}
	\sidecapstyle{\raggedright}
	\setlength{\belowcaptionskip}{10pt}
	\ifx \@empty#1 \subcaption{#2}\else \subcaption[#1]{#2}
}
\usepackage{lettrine}
\newcommand{\initial}[1]{%
	\lettrine[lines=3,lhang=0.33,nindent=0em]{
		\color{gray}
     		{\textsc{#1}}}{}}
\theoremstyle{plain}

\theoremstyle{plain}

\theoremstyle{plain}
\theoremstyle{definition}

\theoremstyle{plain}

\theoremstyle{plain}

\theoremstyle{plain}

\begin{document}
\frontmatter
\pagenumbering{roman}
\begin{titlingpage}
\begin{SingleSpace}
\calccentering{\unitlength} 
\begin{adjustwidth*}{\unitlength}{-\unitlength}
\vspace*{13mm}
\begin{center}
\rule[0.5ex]{\linewidth}{2pt}\vspace*{-\baselineskip}\vspace*{3.2pt}
\rule[0.5ex]{\linewidth}{1pt}\\[\baselineskip]
{\Large Leveraging Robotic Prior Tactile Exploratory Action Experiences For Learning New Objects' Physical Properties
}\\[4mm]
\rule[0.5ex]{\linewidth}{1pt}\vspace*{-\baselineskip}\vspace{3.2pt}
\rule[0.5ex]{\linewidth}{2pt}\\
\vspace{6.5mm}
{\large By}\\
\vspace{6.5mm}
{\large\textsc{Di Feng}}\\
\vspace{11mm}

\includegraphics[scale=0.13]{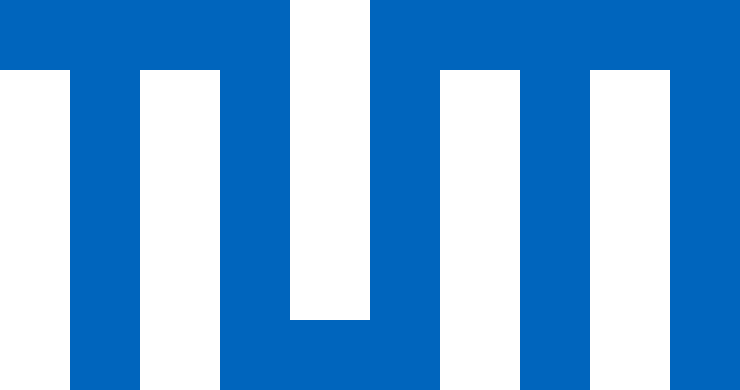} \ \ \ \ \ \ 
\includegraphics[scale=0.1]{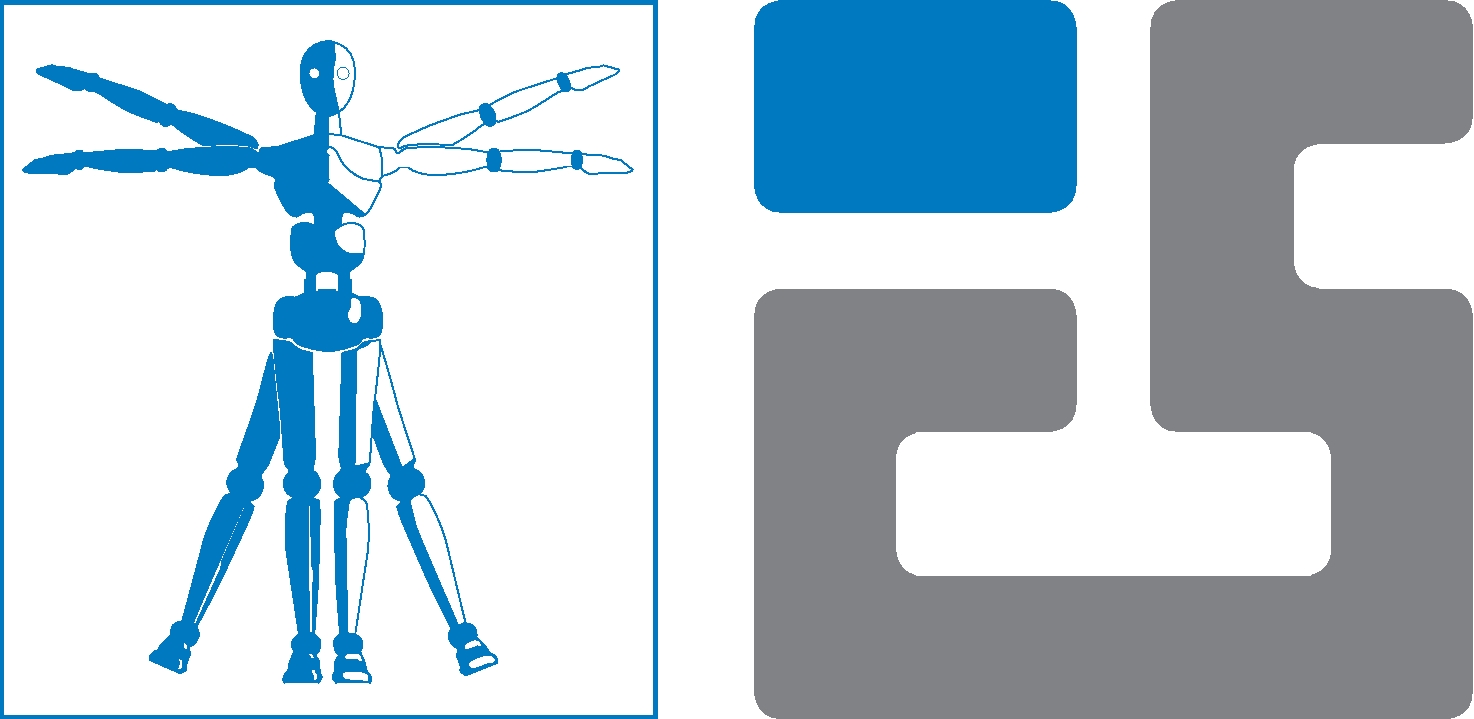}

\vspace{6mm}
{\large Prof. Gordon Cheng\\
Institute for Cognitive Systems\\
\textsc{Technische Universit\"At M\"unchen}}\\
\vspace{11mm}
\begin{minipage}{10cm}
A Master thesis submitted to the Technische Universit\"at M\"unchen in accordance with the requirements of the degree of \textsc{Master of Science} in the Faculty of Electrical and Computer Engineering.
\end{minipage}\\
\vspace{9mm}
{Examiner: Prof. Dr. Gordon Cheng \\
Supervisor: M.Sc. Mohsen Kaboli \\
Topic issued: 11/1/2016 \\
Date of submission: 10/04/2017
}
\vspace{12mm}
\end{center}
\end{adjustwidth*}
\end{SingleSpace}
\end{titlingpage}
\clearemptydoublepage
\begin{vplace}[0.7]
The tactual properties of our surroundings do not chatter at us like their colors; they remain mute until we make them speak. 

\ \ \ \ \ \ \ \ \ \ \ \ \ \ \ \ \ \ \ \ \ \ \ \ \ \ \ \ \ \ \ \ \ \ \ \ \ \ \ \ \ \ \ \ \ \ \ \ \ \ \ \ \ \ \ \ \ \ \ \ \ \ \ \ \ \ \ \ \ \ \ \ \ \ \ \ \ \ \ \ \ \ \ \ \ \ \ \ \ \ \ \ \ \ \ \ \ \ \ \ \ \ \ \ \ \ \ \ \ \ \ \ \ \    \textit{Katz (1925)} \cite{katz2013world}

\end{vplace}
\clearpage
\clearemptydoublepage
\chapter*{Abstract}
Reusing the tactile knowledge of some previously-explored objects helps us humans to easily recognize the tactual properties of new objects. In this master thesis, we enable a robotic arm equipped with multi-modal artificial skin, like humans, to actively transfer the prior tactile exploratory action experiences when it learns the detailed physical properties of new objects. These prior tactile experiences are built when the robot applies the pressing, sliding and static contact movements on objects with different action parameters and perceives the tactile feedbacks from multiple sensory modalities. Our method was systematically evaluated by several experiments. Results show that the robot could consistently improve the discrimination accuracy by over $10\%$ when it exploited the prior tactile knowledge compared with using no transfer method, and $25\%$ when it used only one training sample. The results also show that the proposed method was robust against transferring irrelevant prior tactile knowledge.
\clearpage
\clearemptydoublepage
\chapter*{Acknowledgements}

This master thesis was accomplished at the Institute for Cognitive Systems (ICS), Technische Universit\"at M\"unchen. I feel very thankful for the founder and director of the this institute, Prof. Dr. Gordon Cheng, for providing me the opportunity to learn and grow in this institute and for examining my master thesis.

I would like to express my great gratitude to my supervisor, Mohsen Kaboli, for his patient guidance and invaluable support for my research. I learned a lot from him, both scientifically and personally. Those experiences will be kept in my heart forever.

I would also like to thank my lab mate, Kunpeng Yao, for the projects we have worked together. He is a conscientious guy who always accomplishes the works nicely and ``ordentlich", from which I learned a lot.

I cannot imagine how I can finish my master thesis without the support of my friends. Many thanks to Yuliang Sun, Jiyang Wu and Zhiliang Wu, for helping me to collect the data in the experiments. Thanks to Canlong Ma, Daniel Wasoehrl, Ke Wang, Mingpan Guo, Xiang Yang for the happy time we have been together. During the tough time of my thesis, they added another ingredients to my life.

Special thanks to my girl friend Yi Fan for her uncountable supports and understanding, especially when I was depressed or busy till midnight. Her encouragement meant so much to me. I cannot find another girl better than her in my life. 

Finally, I would like to thank my parents and my grandma for their love throughout my whole life. They support me a lot to study in Germany, and push me to become a better person.

\clearpage
\clearemptydoublepage
\chapter*{List of Publications}
\begin{itemize}
    \item Mohsen Kaboli, \textbf{Di Feng}, Yao Kunpeng, Lanillos Pablo, and Gordon Cheng. A Tactile-Based Framework for Active Object Learning and Discrimination using Multimodal Robotic Skin. In \textit{IEEE Robotics and Automation Letters}, 2(4), 2143-2150, 2017.
    
    \item Mohsen Kaboli, \textbf{Di Feng}, and Gordon Cheng. Active tactile transfer learning for object discrimination in an unstructured environment using multimodal robotic skin. In \textit{International Journal of Humanoid Robotics}, accpeted 25 September 2017.
    
    \item Mohsen Kaboli, Kunpeng Yao, \textbf{Di Feng}, and Gordon Cheng. Tactile-based Active Object Recognition and Manifold Learning in an Unknown Workspace. In \textit{Autonomous Robots}, 1-30, 2018.
    
    \item \textbf{Di Feng}, Mohsen Kaboli, and Gordon Cheng. Active Prior Tactile Knowledge Transfer for Learning Tactual Properties of New Objects. In \textit{Sensors}, 2, 2018.
\end{itemize}

\clearpage
\clearemptydoublepage
%
\renewcommand{\contentsname}{Table of Contents}
\maxtocdepth{subsection}
\tableofcontents*
\addtocontents{toc}{\par\nobreak \mbox{}\hfill{\bf Page}\par\nobreak}

\clearemptydoublepage
\listoftables
\addtocontents{lot}{\par\nobreak\textbf{{\scshape Table} \hfill Page}\par\nobreak}
\clearemptydoublepage

\listoffigures
\addtocontents{lof}{\par\nobreak\textbf{{\scshape Figure} \hfill Page}\par\nobreak}
\clearemptydoublepage

\mainmatter
\let\textcircled=\pgftextcircled
\chapter{Introduction}\label{chapter:intro}
\initial{I}n this chapter, we first introduce the motivation and background of the master thesis. Then, we show the contributions and the organization of this thesis.
\section{Motivation} \label{sec:intro:motivation}
\subsection{Exploring Objects Using A Sense of Touch}
\begin{figure}[!hbtp]
	\centering
	\begin{minipage}{1\linewidth}
		\centering
		\includegraphics[width=0.75\textwidth]{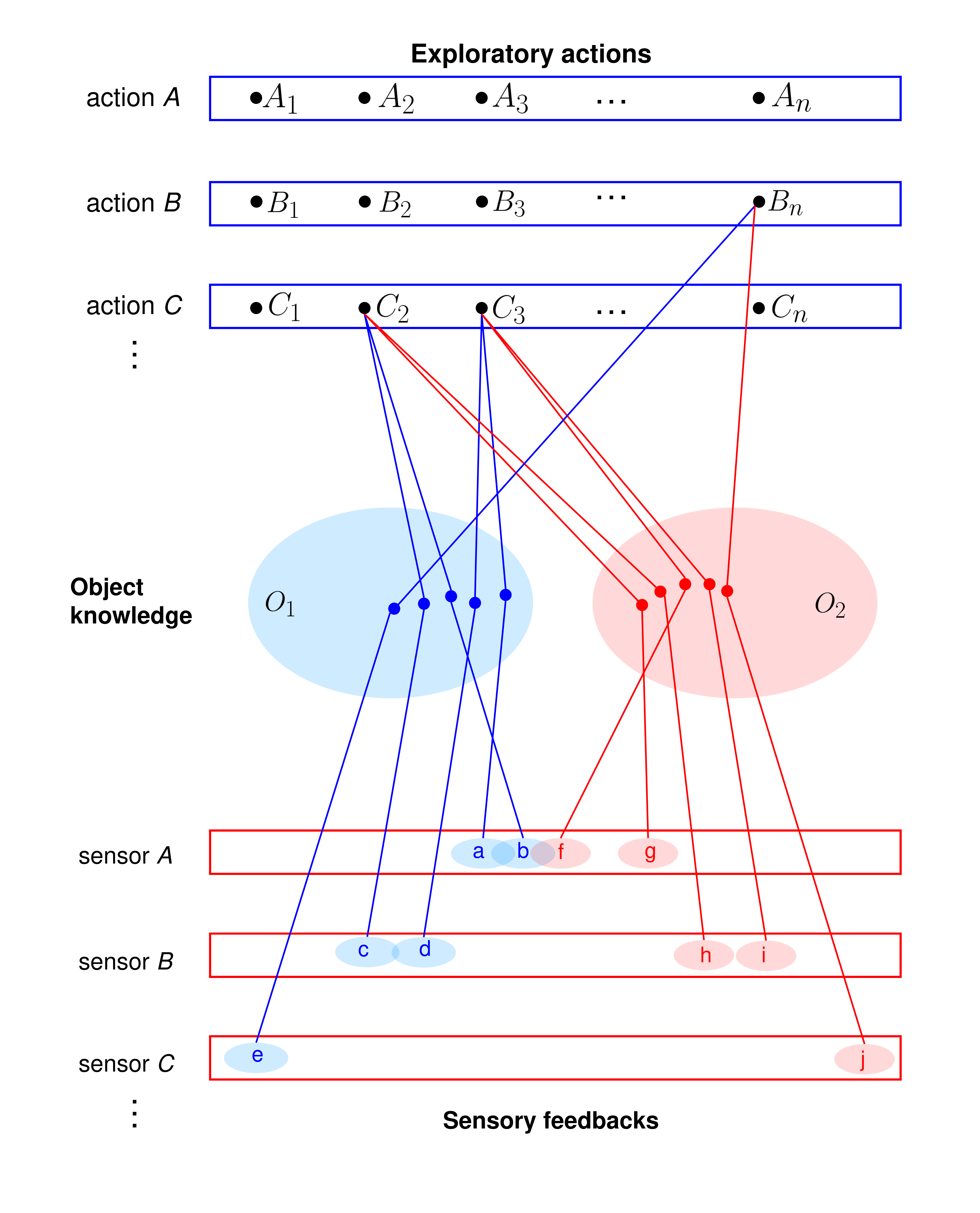}
	\end{minipage}
	\caption[An illustration for the relationship between exploratory actions and object tactile knowledge.]{An illustration for the link between exploratory actions and object tactile knowledge. The picture is reproduced and modified from \cite{loeb2014bayesian}. The tactile knowledge of two objects $O_1$ and $O_2$ (shown by blue and red dots) are built by applying exploratory actions ($A$, $B$, $C$ ...). Each exploratory action can be defined with different action parameters (e.g. $A$ contains $A_1$, $A_2$, ... $A_n$), and can result in multiple sensory feedbacks, shown by the shaded area with indices $a$ - $j$ (e.g. action $C$ can produce observations from sensor $A$ and $B$).   }\label{fig:action-object-pair}
\end{figure}

We humans perceive the tactual properties of an object (such as stiffness, textures, temperature, weight, etc.) by applying exploratory actions. Experimental psychologists have summarized six types of actions that we humans make to explore objects (also known as "Exploratory Procedure": EP) \cite{lederman1987hand}: (1) pressing to perceive object stiffness; (2) static contact to measure the temperature; (3) enclosure to roughly estimate object shape; (4) contour following to determine exact object shape; (5) lateral motion to sense object textural properties; (6) unsupported holding for estimating object weight. After applying different exploratory actions on an object, we can gain its different tactile information. Conversely, making the same exploratory action on different objects will produce different tactile observations \footnote{In the rest of this thesis, we will also refer to the tactile observations after applying an exploratory action as an \textit{instance}.}. Therefore, when we learn about an object (or, when we build the tactile knowledge of an object), we always link its physical properties with the exploratory actions that we apply on this object. As Fig.~\ref{fig:action-object-pair} demonstrates, applying exploratory actions $C_2$ and $B_n$ on object $O_1$ produces different tactile observations (signals $b$, $c$ and $e$); applying $C_2$ on $O_1$ and $O_2$ produces different observations (signals $b$, $c$ and $g$, $h$). Fig.~\ref{fig:action-object-pair} also shows that the object tactile knowledge that combines the exploratory actions and sensory feedbacks for both objects ($O_1$ and $O_2$) are stored as points and links (blue and red points in the shaded area).   

Over the past decades, various tactile sensors have been developed for the robotic systems, so that the robots with a sense of touch can explore objects by applying different exploratory actions (e.g. \cite{jamali2010material,chu2013using,fishel2012sensing}).

\subsection{How to Apply An Exploratory Action?}
\begin{figure}[!hbtp]
	\centering
	\begin{minipage}{1\linewidth}
		\centering
		\includegraphics[width=0.75\textwidth]{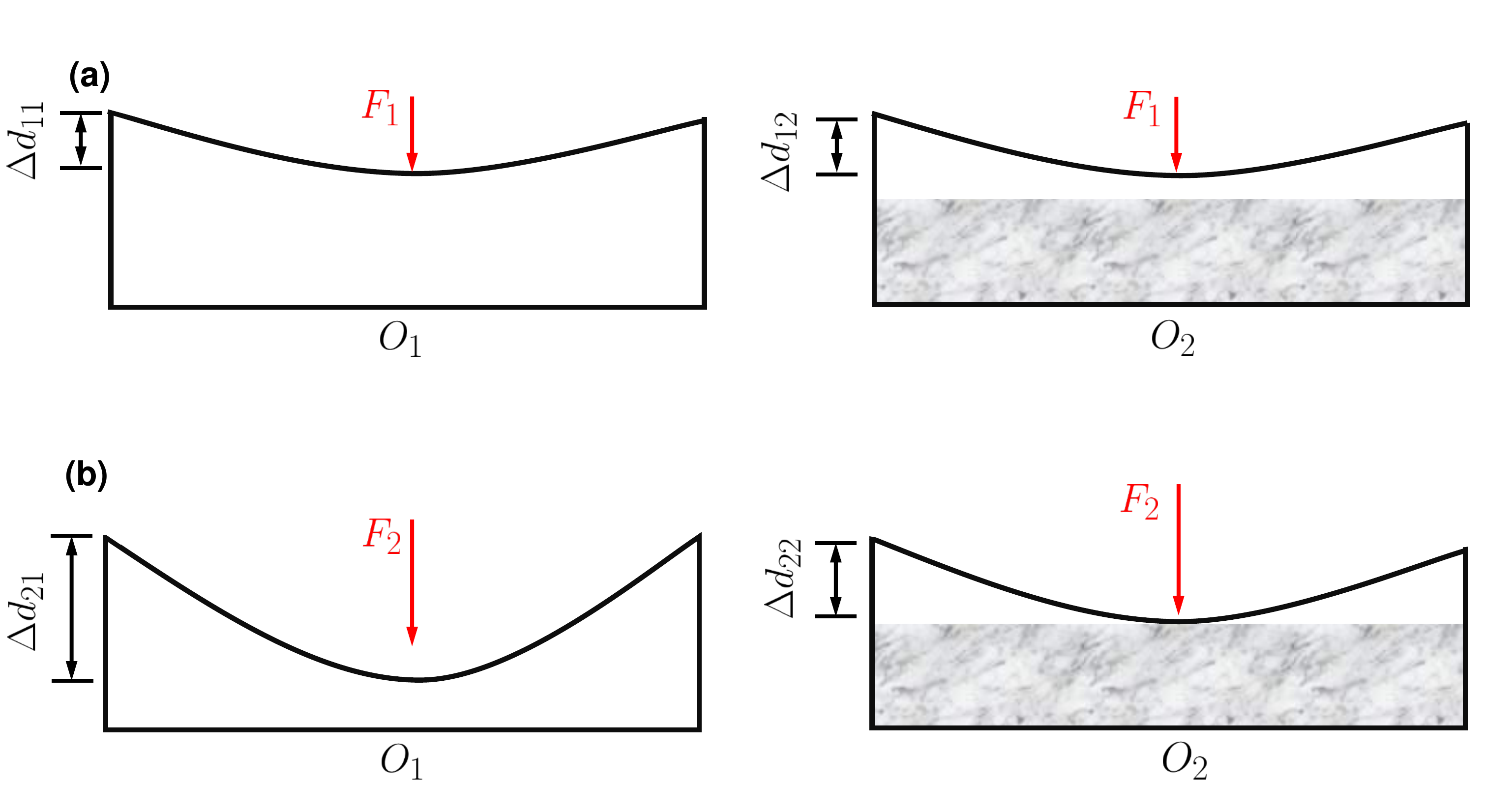}
	\end{minipage}
	\caption[An example to illustrate the effect of applying the pressing movement with different normal forces.]{An example to illustrate the effect of applying the pressing movement with different normal forces.}\label{fig:action_parameter_example}
\end{figure}

Besides different \textit{kinds} of exploratory actions, the tactile information we humans perceive from an object is also dependent on how we apply a \textit{specific} action. Consider an example of pressing on two objects $O_1$ and $O_2$, shown by Fig.~\ref{fig:action_parameter_example}. The object $O_1$ is made of soft sponge, whereas the object $O_2$ is a solid marble, covered by a sponge surface (the white area). When pressing our fingertips on both objects with a small normal force ($F_1$ in Fig.~\ref{fig:action_parameter_example}), we will recognize similar object deformations ($\Delta d_{11}$ and $\Delta d_{12}$). However, if we press with a larger normal force $F_2$, $O_1$ deforms much more than $O_2$ (see $\Delta d_{21}$ and $\Delta d_{22}$ in Fig.~\ref{fig:action_parameter_example}), since we have reached the harder part in $O_2$. A similar situation can also be found when we apply the sliding movement on the object surfaces. Depending on the sliding forces, velocities, and sliding directions, we will sense different textural properties. As a result, by applying different exploratory actions in different ways, we can build a \textit{detailed} knowledge of the object tactual properties which we call "tactile exploratory action experiences". 

Previously, some researchers have investigated the influence of action parameters that a robot employs when it applies exploratory actions on objects. For example, Fishel \textit{et al.} \cite{fishel2012bayesian} designed $36$ sliding movements based on the combinations of six forces and six speeds. They showed that the object discrimination accuracy was dependent on how the sliding movements were applied. Sinapov \textit{et al.} \cite{Sinapov2011Vibrotactile} enabled a humanoid robot to slide its fingertips with different speeds (slow, medium, fast) and directions (medial and lateral) on object surfaces. The results showed that by combining different sliding movements, the robot discriminated among surface textures with higher accuracy.

\subsection{Tactile Knowledge Transfer by Humans}
We humans learn about new objects in an active and incremental way. We actively select the most informative exploratory actions to interact with them. More importantly, we relate these new objects with the experiences of exploring the objects that we have previously encountered. By transferring the prior tactile knowledge, or prior tactile exploratory action experiences, we can largely reduce the amount of exploratory actions required to discriminate among new objects. In this way, we humans save a lot of time and energy, and recognize new objects with high accuracy.

As an example, suppose that there is a soft green kitchen sponge with a fine surface texture and a soft blue kitchen sponge with a rough surface. We can easily distinguish between these two sponges by the sliding movement. However, as the stiffness of both sponges is similar, we need to press more times to recognize their differences. Depending on the normal forces that are applied on the sponges, the pressing feedbacks are different. If we have experienced how to discriminate among kitchen sponges that have similar stiffness, we know how large a pressing force is informative. Such prior knowledge can help us to recognize new sponges more easily. 

Can robotic systems with a sense of touch actively transfer prior tactile exploratory action experiences to enhance the efficiency of learning the physical properties of new objects, like humans?
\section{Background}
The problem of transferring the robotic prior tactile knowledge has been rarely investigated. It was our works that introduced tactile transfer learning. Previously, Kaboli \textit{et al.} \cite{kaboli2015hand} developed a novel textural descriptor. Using the descriptor, a ShadowHand dexterous robotic hand equipped with BioTac sensors on its fingertips could efficiently discriminate among object surface textures. Later, we designed a transfer learning method \cite{kaboli2016re} so that the robotic hand could reuse the prior texture models from $12$ objects to learn $10$ new object textures. However, since only the sliding movement was applied, the robot could only transfer the object textural properties. In a later work \cite{di2017iros}, we proposed an active learning method with which an UR10 robotic arm with an artificial skin on its end-effector could not only apply sliding movement, but also pressing and static contact movements with fixed action parameters on objects to learn about their different physical properties (surface texture, stiffness, and thermal conductivity). Our active learning method enables the robot to efficiently select the exploratory actions to interact with objects. However, the robot still needs to learn from scratch given a new set of objects \footnote{For a more detailed introduction of the method proposed in \cite{di2017iros}, please see the related part in Sec.~\ref{sec:background:action_perception_loop}}.  

\section{Contribution and Organization of this Thesis}
\subsection{Contribution} 
In this master thesis, we focus on actively transferring the prior tactile exploratory action experiences to learn the detailed physical properties of new objects. These prior action experiences consist of feature observations (prior tactile instance knowledge) and observation models of prior objects (prior tactile model knowledge). They are built when a robotic arm equipped with a multi-modal artificial skin applies the pressing, sliding and static contact movements with different action parameters on objects. The feature observations are perceived from multiple sensory modalities. We call our algorithm Active Tactile Instance Knowledge Transfer (ATIKT). Using ATIKT, the robot has a "warm start" of the learning process. It applies fewer exploratory actions on the new objects and achieves higher discrimination accuracy. The master thesis is an extension to our previous works mentioned above.

\subsection{Organization of the Thesis} Chap.~\ref{chapter:background} presents the theoretical background and related works, including an introduction of transfer learning, kernel methods, Gaussian Process (GP) technique and robotic action perception loop. Chap.~\ref{chapter:system_description} introduces the robotic system used in this work, which consists of a UR10 robotic arm and a multi-modal artificial skin. Afterwards, in Chap.~\ref{chapter:action_perception}, we explain how the robot applies different exploratory actions with different action parameters on objects and how to extract the feature observations from different tactile sensors. Our proposed transfer learning algorithm is introduced in Chap.~\ref{chapter:method}, in which we show what, how, how much, and from where to transfer prior tactile exploratory action experiences. In Chap.~\ref{chapter:experiment}, we systematically evaluate our proposed transfer learning method with several experiments. In Chap.~\ref{chapter:conclusion}, we summarize the thesis and discuss potential future works.

\cleardoublepage
\let\textcircled=\pgftextcircled
\chapter{Background}\label{chapter:background}
\initial{T}he following chapter gives an overview of the theoretical background related to this master thesis. First, we introduce supervised learning scenario. Then, we put special focus on introducing transfer learning technique (TL). In this work, The Gaussian Process (GP) model with customized kernel is used to transfer the prior tactile instance knowledge. Therefore, after the introduction of TL, we introduce the kernel technique including its motivation and the kernel construction rules, and the Gaussian Process (GP) model with its applications in the transfer learning. Finally, we give the information about the tactile-based robotic action perception loop, with which a robot with a sense of touch can explore objects. We also summarize the methods proposed in \cite{di2017iros}, as the master thesis is an extension based on it.
\section{Supervised Learning} \label{sec:background:supervised_learning}
In machine learning technology, supervised learning aims to train a mathematical model that gives reliable prediction results given test data samples. To do this, the mathematical model is trained with the training data comprised of examples of the input feature vectors and their corresponding target vectors \cite{christopher2006pattern}. Common supervised learning models include Logistic Regression, Support Vector Machine (SVM), Gaussian Process (GP), and Artificial Neural Network (ANN), to name a few.

Formally, supervised learning describes the functional mapping $f: X \mapsto Y$ between the input dataset $X$ and the output dataset (or target) $Y$. Given a new sample $\mathbf{x}^\ast$, the function predicts the output denoted as $\mathbf{y}^\ast$, $\mathbf{y}^\ast  =f(\mathbf{x}^\ast)$. 

\begin{figure}[!htp]
	\centering
	\begin{minipage}{0.98\textwidth}
		\centering
		\subtop[Regression]{\includegraphics[width=0.49\linewidth]{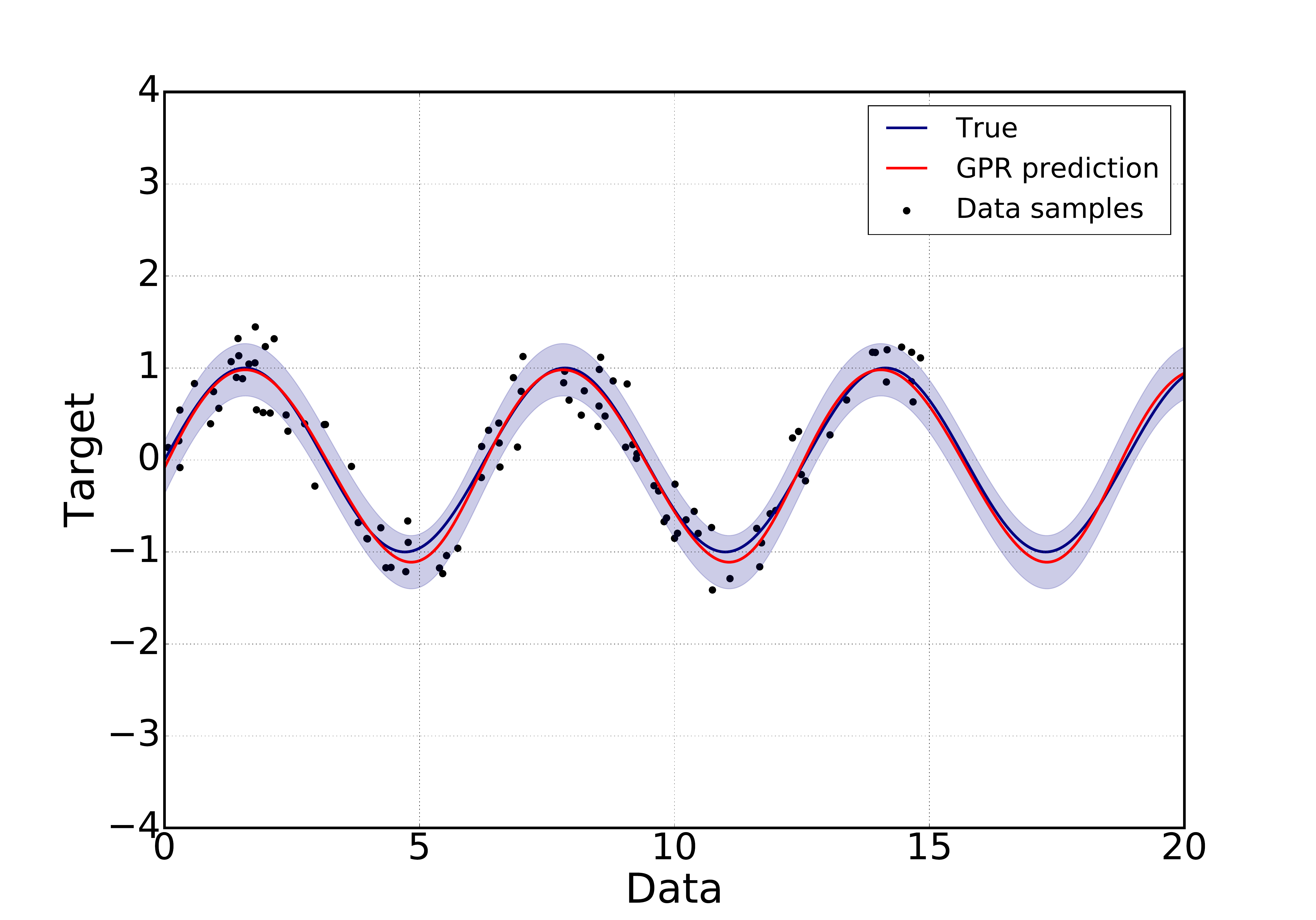}\label{fig:regression_example}}
		\subtop[Classification]{\includegraphics[width=0.49\linewidth]{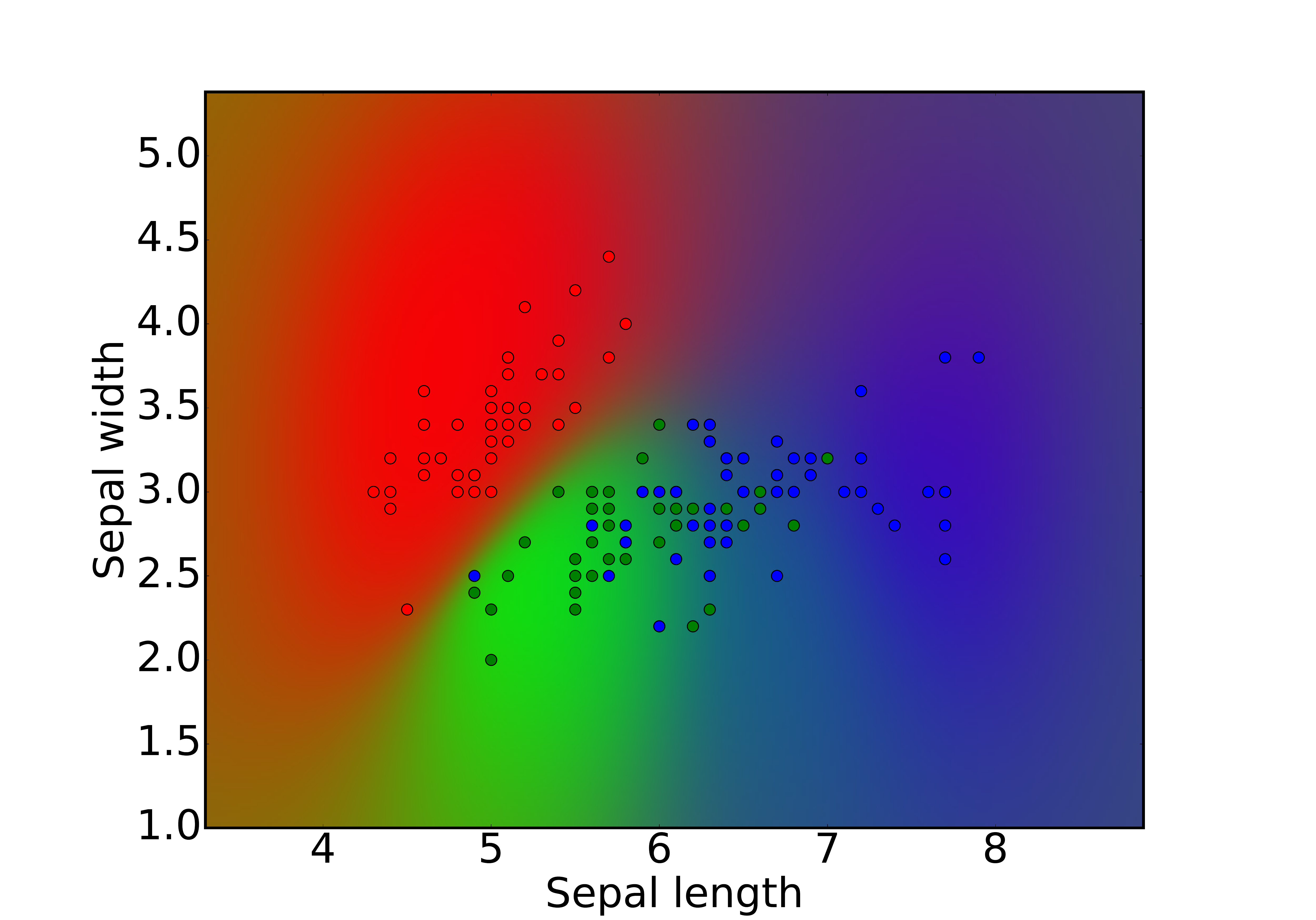}\label{fig:classification_example}}
	\end{minipage}
	\caption[Examples of regression and classification in supervised learning.]{Examples of regression and classification in supervised learning. \textbf{(a)}: The GP regression model is trained to fit the sinusoidal target function. \textbf{(b)}: The GP classification model is employed to classify three categories of iris flowers. The plots are modified from the sklearn website: \textit{http://scikit-learn.org/stable/modules/gaussian\_process.html}.
	}
\end{figure}

Based on the target $Y$, supervised learning can be divided into \textit{regression} and \textit{classification}. In the former problem, the desired output consists of one or more continuous variables \cite{christopher2006pattern}. Fig. \ref{fig:regression_example} shows an example of modeling the artificial dataset by GP regression (GPR). The dataset is generated by adding some noise to a sinusoidal target function (blue curve). As can be seen in the plot (Fig. \ref{fig:regression_example}), GPR correctly fits the target function (red curve), and provides reasonable confidence bounds to the prediction which is visualized by the blue shaded area. 

As for the classification problem, the target $Y$ contains of discrete values called labels. Each label represents a category, or class. The goal is to assign each input vector to one of a finite number of discrete categories \cite{christopher2006pattern}. Fig. \ref{fig:classification_example} illustrates the task of classifying three types of iris flowers, whose feature distributions are shown with red, green, and blue dots. The GP classification (GPC) model is employed to train the classifier. The prediction of the class region is demonstrated by the color. The brightness of color shows the confidence of the prediction, i.e. a light color means high confidence, while a dark color low confidence.  
\section{Transfer Learning}\label{sec:background:TL}
\subsection{Introduction}
\subsubsection{Motivation}\label{sec:background:TL-Motivation}
Most of the supervised learning algorithms assume that the training data and the test data share the same feature representations and are drawn from the same distributions. Moreover, the performance of algorithms is highly dependent on the quality and the number of training samples \cite{vapnik2013nature}. In many real world applications, however, collecting labeled training samples are costly and tedious. This problem has hindered the applicability of machine learning methods in practice. For instance, in order to train a classifier to discriminate among objects based on their textural properties, we need to collect sufficient training samples by sliding the tactile sensors on object surfaces. On one hand, executing the sliding action is time consuming, because one needs to adjust the position of the sensors during sliding in order to  accurately control the contact force. On the other hand, it is very costly to slide the tactile sensors on objects many times, because the sensors can be easily destroyed, especially when the surface textures are hard and rough. These constraints limit the number of training samples the robot can obtain for object textures, and thus limit the performance of an object texture classifier.  

To tackle with the bottleneck mentioned above, many methods have been developed aiming at using only a small number of labeled training samples to learn a model that can achieve a greater prediction accuracy to the test data. For instance, the \textit{active learning} technique usually assumes that the training samples a machine learner will learn can be selected and labeled by an oracle (usually human annotator) from a large unlabeled data pool. By iteratively querying the oracle to label the most informative (or most valuable) training sample and assigning it into the training dataset, an active machine learner can perform better with fewer labeled training samples. However, since there may be limited budget for querying the oracle, the training dataset is still insufficient to train an accurate model \cite{aggarwal2014data}.

\begin{figure}[!hbtp]
	\centering
	\begin{minipage}{1\linewidth}
		\centering
		\includegraphics[width=1\textwidth]{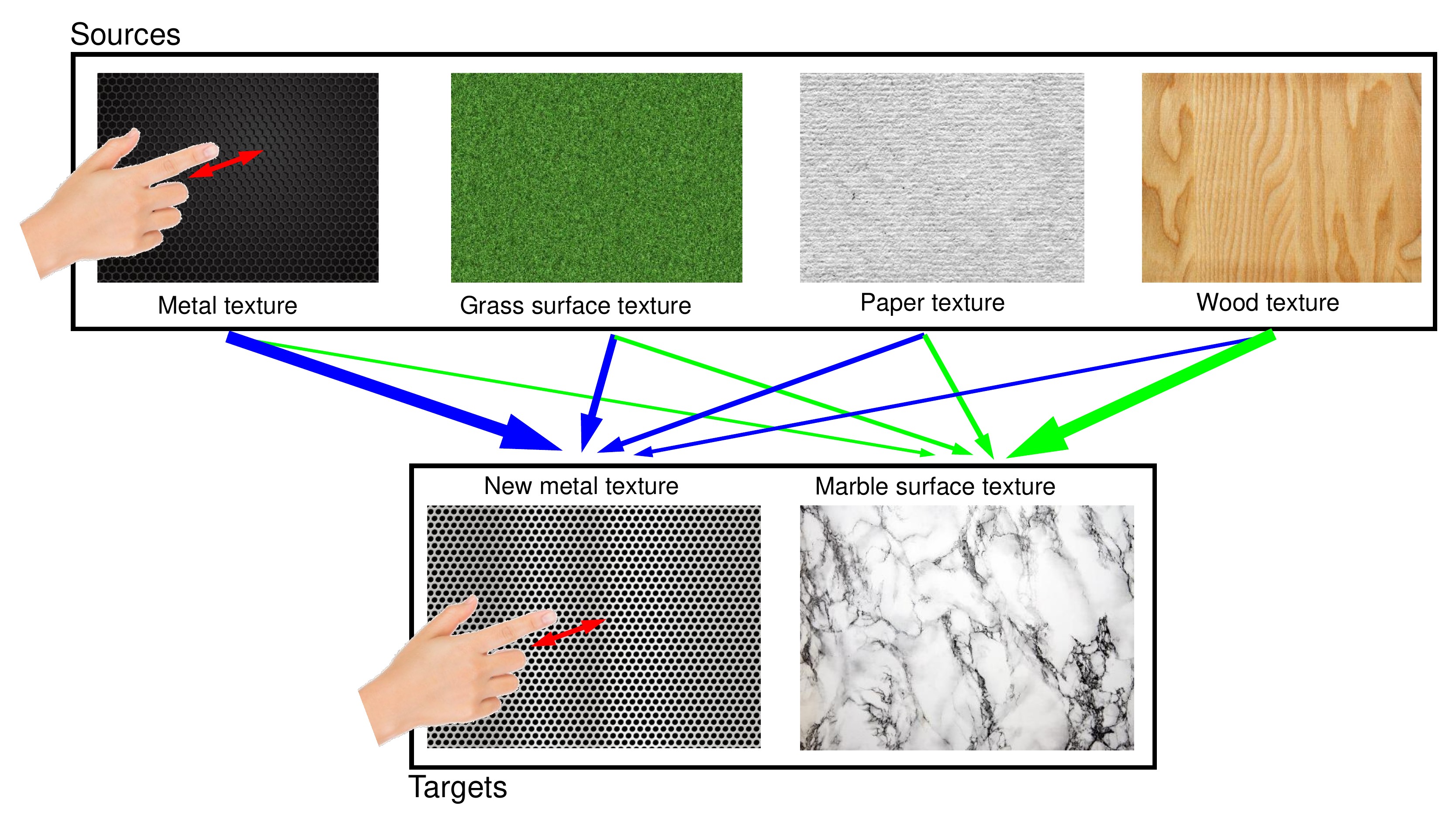}
	\end{minipage}
	\caption[An example for transfer learning scenario.]{An example to illustrate transfer learning scenario. The arrow thickness indicates the object relatedness.}\label{fig:TL_tactile_eg}
\end{figure}

Instead, \textit{transfer learning} aims at leveraging the prior knowledge from the related tasks which the machine has learned previously, in order to improve the performance for the current task. The previous tasks may come from different domains and have different data distribution or feature representations with the current task. 
The transfer learning scenario can be easily observed in our daily life. For example, when a group of Chinese students learn German as a foreign language, those students who have a good knowledge of English can learn faster than those who do not have any experience in learning a foreign language, since the words, logic as well as the grammar between German and English share many similarities. Another example of the transfer learning in the field of machine learning can be illustrated by the task of learning different surface textures (Fig. ~\ref{fig:TL_tactile_eg}). Suppose that a system have already slid on different surface materials (metal surface, grass surface, paper and wood textures) and built the prior knowledge of textures. Now the system is tasked to learn new surface textures (new metal texture and wood texture). Since the new metal texture is similar to the old metal texture, a texture learner can reuse some prior texture knowledge from the old metal texture to increase the learning speed of new metal texture. Furthermore, grass surface texture, paper texture and wood texture share less similarity with the new metal texture, therefore, little prior knowledge can be leveraged. On the contrary, when learning about the marble surface texture, more prior knowledge can be used from the wood texture instead of the metal texture, as both of the surface materials (marble and wood) are smooth. By reusing the prior texture knowledge, a system can learn new surface textures with fewer sliding movements. 

\subsubsection{Terminology and Definition}
We define domain $D$ as a combination of a feature space $X$ and a marginal probability distribution $P(\mathbf{x})$, $\mathbf{x}\in X $ , $D = \{X,P(\mathbf{x})\}$. We also define task $T = \{Y,f(\cdot)\}$, where $Y$ is the label space, and $f(\cdot)$ is the predictive function that maps the instance $x \in X$ to the label $y \in Y$. In transfer learning, usually there exist a single or multiple source domains $D_S$ and a target domain $D_T$. The source domains are where the prior knowledge comes from; target domain $D_T$ is the domain of the new task that the system is required to learn. The main idea for transfer learning is to use the knowledge from some related domains (source domains) to help a machine learner to achieve a better performance in the target domain. It can be defined as follows (\cite{aggarwal2014data}):\\
\\
\textbf{Definition} \textit{Given a source domain $D_S$ and learning task $T_S$, a target domain $D_T$ and learning task $T_T$ , transfer learning aims to help improve the learning of the target predictive function $f_T(\cdot)$ in $D_T$ using the knowledge in $D_S$ and $T_S$, where $D_S \neq D_T$ , or  $T_S \neq T_T$.} 

\subsubsection{Three Ways of Improving The Performance of a Machine Learning Model}
\begin{figure}[!hbtp]
	\centering
	\begin{minipage}{1\linewidth}
		\centering
		\includegraphics[width=0.6\textwidth]{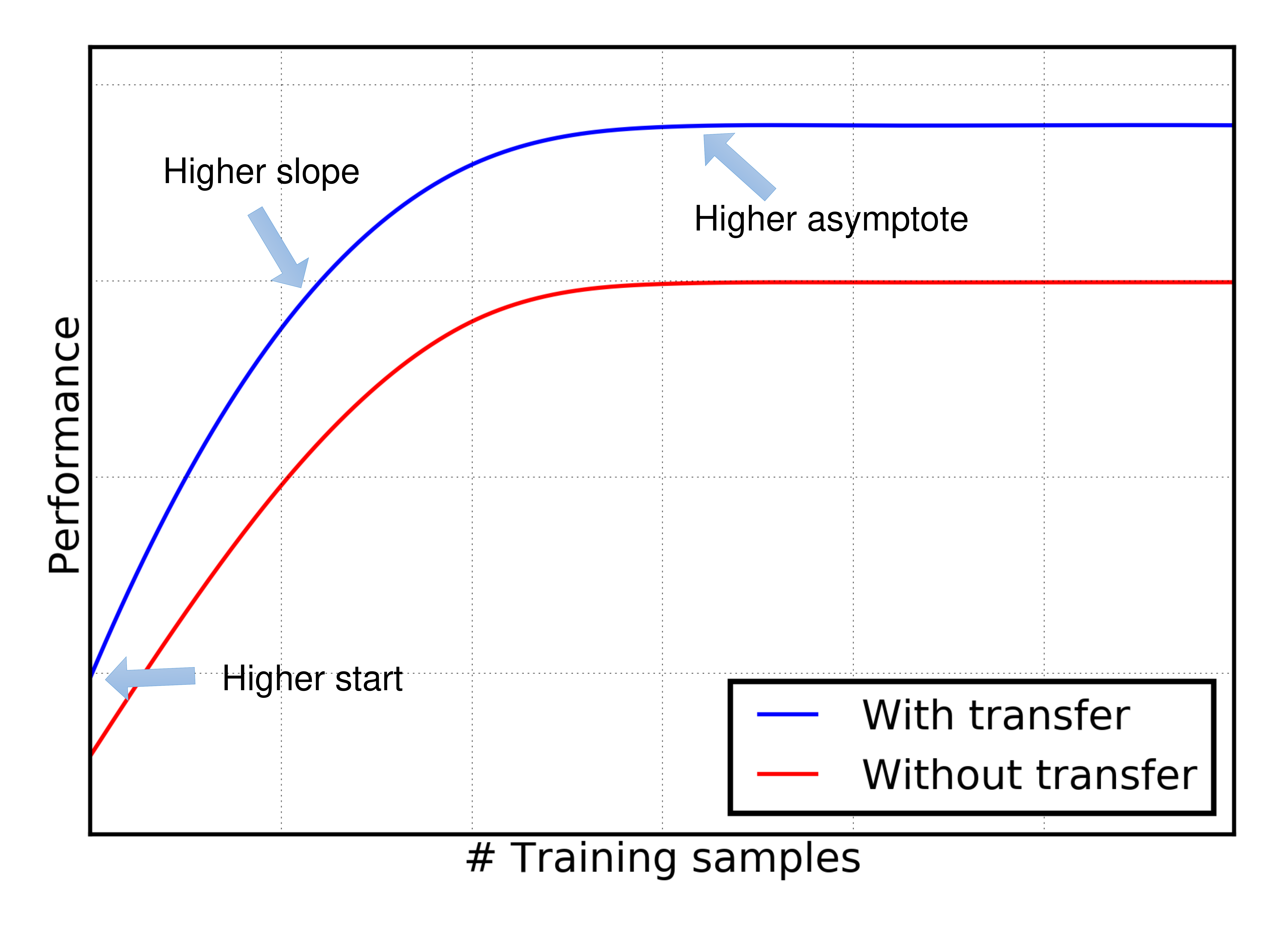}
	\end{minipage}
	\caption[Three ways transfer learning helps to improve the learning process.]{Three ways transfer learning helps to improve the learning process (Figure reproduced from \cite{olivas2009handbook}).}	\label{fig:transfer_learning}
\end{figure}

Transfer learning can boost the learning process in three ways \cite{olivas2009handbook} (see Fig. \ref{fig:transfer_learning}): (1). higher start: since a transfer learner can borrow the prior knowledge to help initialize the learning process of the new task, the initial performance of a model with transfer learning is often much higher than without transfer learning. This behaviour is often referred to as one-shot learning \cite{tommasi2010safety}; (2). higher slope: the learning performance grows faster; (3). higher asymptote: when the learning process becomes stable, i.e. there are sufficient training samples, the model with transfer learning can achieve a higher accuracy than without transfer learning.

\subsection{Research Issues of Transfer Learning} \label{Sec:TL-research_issues}
When designing transfer learning algorithm, several issues need to be addressed, namely what to transfer, how to transfer, how much to transfer, and from where to transfer. \\

\noindent \textbf{What to transfer} tackles with the issues of which part of knowledge can be transferred from the source domain to the task domain \cite{pan2010survey}. The prior knowledge can be represented by the training instances (instance-based transfer), common features across domains (feature-based transfer), the parameters in the models (parameter-based transfer) or the relationships between samples (relational-information-based transfer), to name a few. Some of the prior knowledge is task-specific, and thus is useless when learning the new task. Some knowledge is common and can be shared by different domains and tasks, therefore they can help to improve the learning performance of the new tasks.\\

\noindent \textbf{How to transfer} asks how to use the prior knowledge mentioned above to develop the learning algorithms. Different representations of prior knowledge leads to different transfer strategies (briefly summarized in Sec.~\ref{Sec:TL-methodology}).\\

\noindent \textbf{How much to transfer} asks how much the prior knowledge can be transferred to the target domain. Since the \textit{relatedness} (or \textit{transferability}) between sources and targets are different, we want to transfer more information when sources are highly related to the target (e.g. in Fig.~\ref{fig:TL_tactile_eg} wood texture and marble surface texture), and less when they share less commonality (e.g. in Fig.~\ref{fig:TL_tactile_eg} wood texture and new metal texture). In the extreme case, when the source domain and the target domain are not related, brute-force transfer may even hurt the performance of learning in the target domain, which is called \textit{negative transfer} \cite{pan2010survey}. In this case, knowledge transfer should be stopped.\\

\noindent \textbf{From where to transfer} asks the question of how many sources are needed to knowledge transfer. Suppose a situation where there exists many sources for a target task. On one hand, we can select only one source to support the target. However, this method does not fully exploit all the available prior knowledge. On the other hand, we may use a combination of all sources to help the learning process in the target domain. However, using all sources may largely increase the computational cost.  Therefore, coping with "from where to transfer" is an important issue when designing transfer learning techniques.\\

\subsection{Methodology}\label{Sec:TL-methodology}
In the previous section (Sec.~\ref{Sec:TL-research_issues}), we briefly introduced the problem of "what to transfer" and "how to transfer". Now let us talk about the methods w.r.t these issues in detail. Most of the transfer learning approaches deal with binary classification ($Y=\{-1,+1\}$), and the feature spaces between source domain and target domain are related (homogeneous transfer)\cite{aggarwal2014data}. In this setting, there are several ways to transfer the prior knowledge.\\

\noindent \textbf{Instance-based Approach}. The idea is to reuse training samples in the source domain for the task domain. It assumes that the source domain and target domain share the same feature representations, and the data distributions are similar \cite{aggarwal2014data}. The instance-based approach first re-weights or re-samples training samples from the source domain, and then uses them to learn the new task. \\

\noindent \textbf{Feature-based Approach}. 
If there is little feature overlapping between source domain and target domain, directly transferring the instances from source domain is inappropriate. In this regard, we can learn a new feature representation, such that the source domain data are mapped to the new features that can be reused for the target domain. Formally, this approach aims at learning the mapping $\phi(\cdot)$, with which the difference between two domain after transformation (i.e. $\{\phi(x_{S_i})\}$ and $\{\phi(x_{T_i})\}$) can be reduced \cite{aggarwal2014data}. For instance, in \cite{li2012multi}, an transfer learning algorithm was proposed for simultaneously training different text classification systems. Since different text classification problems have some commonality, e.g. product sentiment prediction system for electronics, books, or furniture in Amazon.com \cite{li2012multi}, the author proposed to jointly learn the classifiers in different domain together. First, they decomposed the text feature into a common latent feature and a domain-specific feature. The former was attained by applying Spectral Feature Alignment. Then, they encoded the training samples that comes from different domains and are represented by common features into the normal model loss function.\\

\noindent \textbf{Parameter-based Approach}. Unlike the instance and the feature based transfer learning which are based in the data level, model based approach extracts the knowledge from the model parameters in the source domain. This method is motivated by the fact that "a well-trained source model has captured a lot of structure, which can be transferred to learn a more precise target model \cite{aggarwal2014data}". For example, in order to transfer the prior texture models built by the Least Square Support Vector Machine model (LSSVM), Kaboli \textit{et al.} \cite{kaboli2016re} proposed to use the model parameters $\mathbf{\hat{w}}$ from the prior texture models as a bridge that links between source models and target models. First, $\mathbf{\hat{w}}$ was incorporated into the cost function of the LSSVM for new texture models. By optimizing the cost function, $\mathbf{\hat{w}}$ helped to improve the learning performance of new object textures. As for the Gaussian Process Classification (GPC) model, the parameters in the kernel function can be used to transfer (introduced in Sec.~\ref{sec::background::GP_model}). 
\\

\noindent\textbf{Relational-information-based Approaches}. This approach assumes that data samples share similar relationship between each other in the source domain and in the target domain. Thus, the idea is to map the relational knowledge between two domains. This approach can be regarded as a higher level of knowledge transfer, and can be easily observed in our daily life. As an example, the relationship between a professor and PhD students are somehow similar to that between a project manager and project engineers.
\\

\subsection{Other Research Issues of Transfer Learning}
\subsubsection{Heterogeneous Knowledge Transfer}
Besides homogeneous transfer learning, in some other problems, the feature space or the label space between source domain and target domain are different. In this setting, the transfer learning methods aim at finding the relationship of features (or labels) across different domains \cite{aggarwal2014data}. For example, how to transfer the prior knowledge about texts to the web image categorization is a heterogeneous feature transfer problem. 

\subsubsection{Active Transfer Learning}\label{subsubsec::background::ATL}
As mentioned in the Section related to the TL motivation (Sec.~\ref{sec:background:TL-Motivation}), both transfer learning and active learning aim at learning a reliable model for the tasks of classification or regression with minimal necessary training samples. It is a direct idea to combine active learning with transfer learning in order to further reduce training samples or human supervision effort. For example, both Shi \textit{et al.} \cite{shi2008actively} and Saha \textit{et al.} \cite{saha2011active} proposed active transfer learning frameworks. First, the classifier in the source domain was adapted to the target domain. Then, a "hybrid oracle" was constructed which could either directly label the new unlabeled instance by the adapted source classifier, or use generic active learning methods to quest the human oracle.

\section{Kernel Method} \label{sec:background:kernels}
\subsection{From Linear Classifier to Non-linear Classifier}
\begin{figure}[!htp]
	\centering
	\begin{minipage}{0.98\textwidth}
		\centering
		\subtop[Linear separable classification ]{\includegraphics[width=0.49\linewidth]{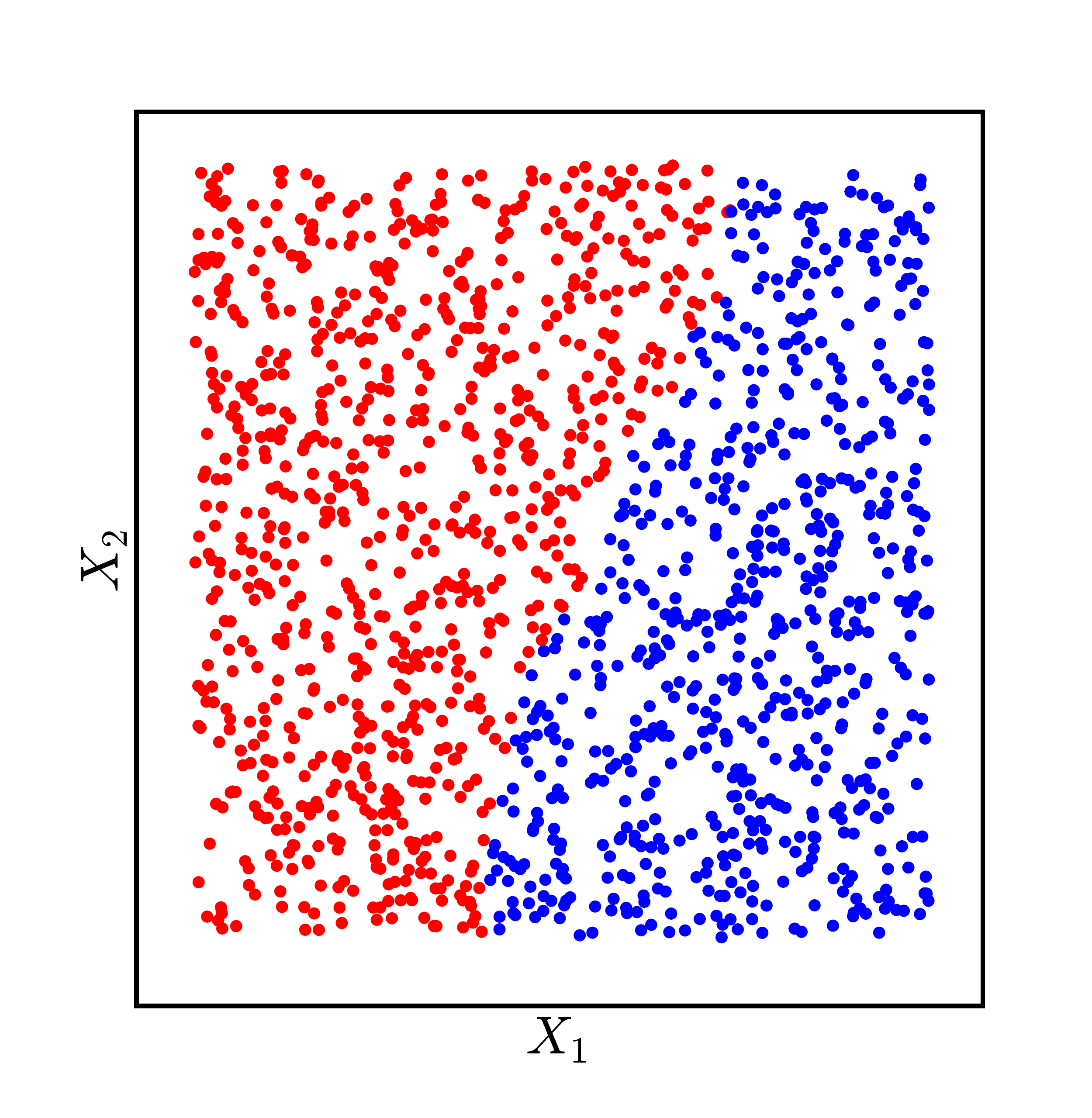}\label{fig:linear_classification_problem}}
		\subtop[Non-linear separable classification]{\includegraphics[width=0.49\linewidth]{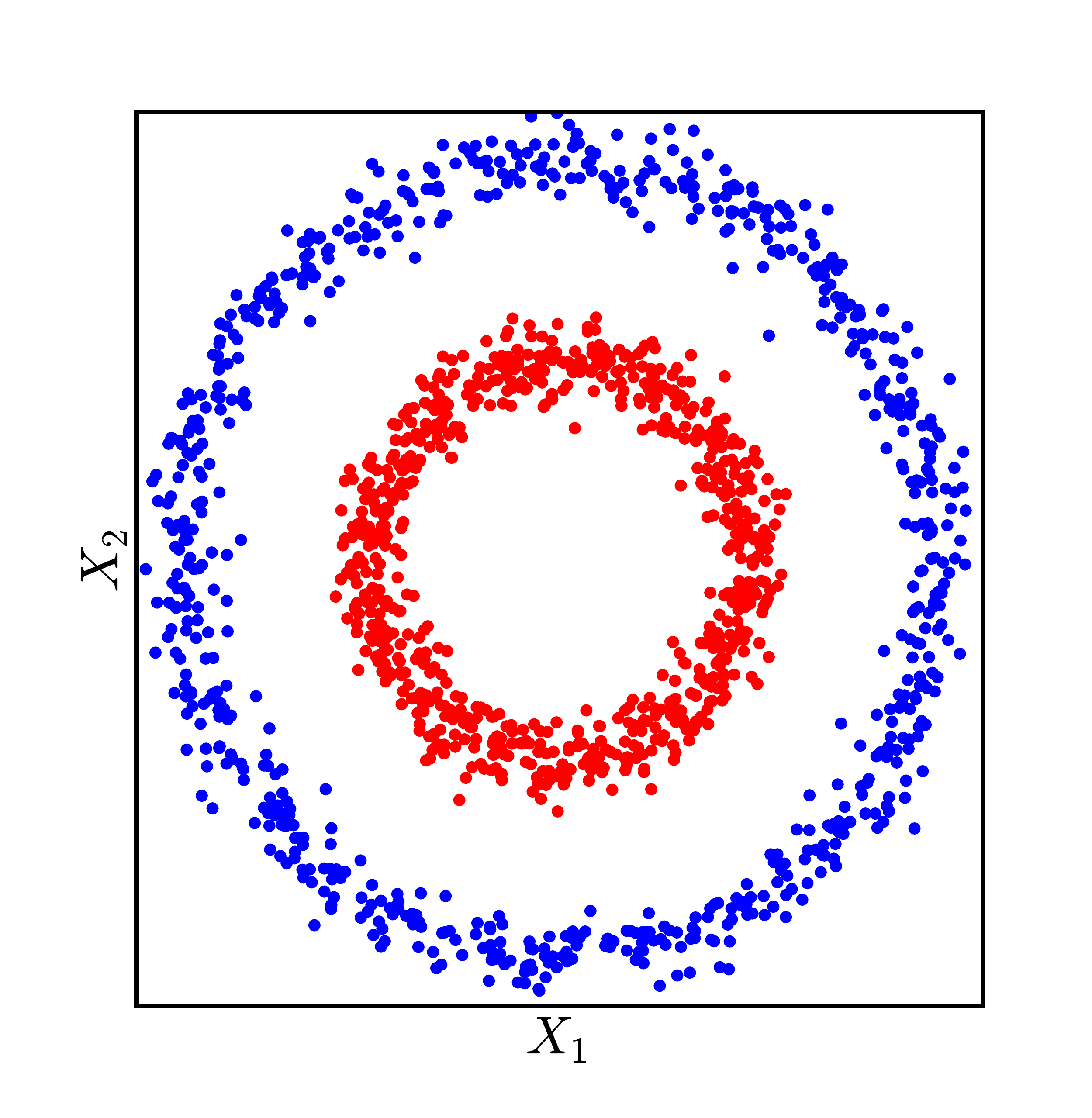}\label{fig:non_linear_classification_problem}}
	\end{minipage}
	\caption[Linear separable and non-linear separable classification problems .]{Linear separable and non-linear separable classification problems. \textbf{(a)} the red and blue classes can be easily classified by a linear function.  \textbf{(b)} The margins of the two classes are not linear separable, therefore a non-linear classifier is needed.}\label{fig:linear-nonlinear-classification-problem}
\end{figure}

We focus on classification problems. 
In many situations, classes are linearly separable by \textit{linear classifiers}. These classifiers are learned to find appropriate hyper-planes that can separate the space of training samples. In binary classification problem (i.e. $Y=\{-1,+1\}$), a linear classifier can be described by:
\begin{equation}
f(\mathbf{x}) = \text{sign}(\mathbf{w}^{T}\mathbf{x}+b)
\end{equation}\label{eq:linear_classifier}

\noindent where $\mathbf{w}$ is the normal vector to the hyper-plane and $b$ is the offset. Fig.~\ref{fig:linear_classification_problem} illustrates the feature distributions of a two-class toy dataset. It can be easily seen that a linear classifier can solve such a problem. 

However, in many cases (e.g. Fig.~\ref{fig:non_linear_classification_problem}), linear classifiers are not sophisticated enough to describe the complex data distributions. In this scenario, we can transform the linear classifiers to \textit{non-linear classifiers} by mapping the original observations $\mathbf{x}$ to a higher dimensional feature space with the transformation function: $\phi(\mathbf{x})$. Accordingly, Eq.~\ref{eq:linear_classifier} can be  modified as:
\begin{equation}
f(\mathbf{x}) = \text{sign}(<\mathbf{w},\phi(\mathbf{x})> + b)
\end{equation}\label{eq:non-linear_classifier}
where $<\cdot,\cdot>$ is the scalar product.

According to the representer theorem from \cite{scholkopf2001generalized}, the normal vector $\mathbf{w}$ can be expressed as a linear combination of the observations in a high dimensional space:
\begin{equation}
\mathbf{w} = \sum_{i=1}^{n}\alpha_i\phi(\mathbf{x}_i).
\end{equation}\label{eq:representer-theroem}
Therefore, Eq.~ \ref{eq:linear_classifier} can be further modified as:
\begin{equation}
f(\mathbf{x}) = \text{sign}(\sum_{i=1}^{n}\alpha_i<\phi(\mathbf{x}_i),\phi(\mathbf{x})> + b).
\end{equation}\label{eq:non-linear_classifier-with-phi}

\subsection{Kernel Trick}
Instead of finding the explicit transformation function $\phi(\cdot)$, we can find a kernel (or kernel function): $K:X \times X\xrightarrow{} \mathbb{R}$ that computes the scalar product: $K(\mathbf{x}_i,\mathbf{x}_j) = <\phi(\mathbf{x}_i),\phi(\mathbf{x}_j)>$. This is often referred to as "kernel trick". In this way, we only need to design an appropriate kernel and incorporate it into the linear classification model to solve more complex non-linear problems. The introduction of kernels plays a very important role in many supervised learning models, such as SVM and GP. There are two big advantages of kernels:
\begin{itemize}
\item[1).] Kernels can be regarded as a similarity measurement between two observations. An appropriate kernel gives us more insight of the data structure. 
\item[2).] Kernels can be regarded as a normalization tool to compare different signals which may have different physical meanings or different feature representations. 
\end{itemize}

\subsection{Kernel Construction}\label{subsec:background:kernel_contruction}
\noindent \textbf{Mercer Condition.} The Mercer condition \cite{vapnik2013nature} provides us the basic rule of kernel construction. It states that if a \textit{kernel matrix} is positive semi-definite, there always exists a corresponding mapping $\phi(\cdot)$. The kernel matrix is built by computing the kernel values between each two observations from the dataset. Formally, a kernel is positive semi-definite, when the following equation satisfies for all vector $\mathbf{g}$ \cite{vapnik2013nature}:
\begin{equation}
<\mathbf{g},K\mathbf{g}> = \mathbf{g}^{T} \cdot K\mathbf{g} = \sum_i^N \sum_j^N g_i K_{ij} g_j \geq 0.
\end{equation}\label{eq:mercer-condition}

When building a new kernel, we need to ensure that the kernel matrix is positive semi-definite.\\

\noindent \textbf{Linear Kernel Combination.}
By linearly combining different kernel matrices, we can construct a complex kernel that captures more information of the dataset. As we will see in Chapter \ref{chapter:method}, this technique is a simple but powerful tool to fuse different sensor signals together. A linearly combined kernel can be expressed as:
\begin{equation}
K' = \gamma^{(1)} K^{(1)} + \gamma^{(2)} K^{(2)} + ... + \gamma^{(m)} K^{(m)} + ... + \gamma^{(M)} K^{(M)}
\end{equation}\label{eq:kernel-combination}

\noindent where $\gamma_m \geq 0$. In the following, we will prove that the kernel $K'$ satisfies the Mercer condition.\\

\noindent \textbf{Proof:}

\noindent The scalar product in Eq.~\ref{eq:mercer-condition} can be rewritten as:
\begin{align}
<\mathbf{g},K'\mathbf{g}> = \mathbf{g}^{T} \cdot K'\mathbf{g} = \sum_i^N \sum_j^N g_i K'_{ij} g_j \\
= \sum_i^N \sum_j^N g_i (\sum_m^{M} \gamma^{(m)} K^{(m)})_{ij} g_j \\
= \sum_i^N \sum_j^N \sum_m^{M} (\gamma^{(m)} g_i K^{(m)}_{ij} g_j). 
\end{align}\label{eq:prove-linear-kernel-combination1}
Since each basic kernel satisfies the Mercer Condition, i.e. $g_i K^{(m)}_{ij} g_j \geq 0$, and $\gamma^{(m)}$ is an non-negative value, we have: 
\begin{align}
<\mathbf{g},K'\mathbf{g}>
= \sum_i^N \sum_j^N \sum_m^{M} (\ \ \underbrace{\gamma^{(m)}_{\ }}_{\geq 0}\ \ \cdot  \underbrace{g_i K^{(m)}_{ij} g_j}_{\geq 0}\ \ ) \geq 0. 
\end{align}\label{eq:prove-linear-kernel-combination2}

\section{Gaussian Process} \label{sec::background::GP_model}
The Gaussian Process (GP) model is a kernel-based supervised learning method (Sec.~\ref{sec:background:supervised_learning}). There are several motivations of using GP: (1). GP is a probabilistic model that gives the observation prediction with probability (in other words, empirical confidence). Therefore, the GP model is suitable for the problems that need Bayesian inference. (2). The GP model can use different kinds of kernels, depending on problem domains. Its biggest drawback is the quickly decreasing performance and computational efficiency, when the feature dimension or the number of training samples grow (curse of dimensionality).  
\subsection{The Model}
The GP model describes the mapping between the observation set $X$ and the output $Y$ by: $X \xrightarrow{f} Y$. It assumes that there exists an underlying function $h:$ $X \xrightarrow{h} \mathbb{R}$ such that given $h(\mathbf{x}_i)$, the output $y_i$ is conditionally independent from the input $\mathbf{x}_i$ with a so-called \textit{noise model} $p(y_i|h(\mathbf{x}_i))$ \cite{rasmussen2006gaussian}.

The latent function $h(\mathbf{x})$ is assumed to be sampled from a high-dimensional gaussian distribution called GP prior \cite{rasmussen2006gaussian}: $h(\mathbf{x})\thicksim\mathcal{GP}(m(\mathbf{x}),K(\mathbf{x},\mathbf{x}'))$, where each sample $h(\mathbf{x_i})$ is a random variable. The mean function $m(\mathbf{x})$ and the covariance function $K(\mathbf{x},\mathbf{x}')$ are defined by:
\begin{align}
m(\mathbf{x}) =\mathbb{E} [\mathbf{x}], \\
K(\mathbf{x},\mathbf{x'}) = \mathbb{E} [ (h(\mathbf{x})-m(\mathbf{x})) (h(\mathbf{x'})-m(\mathbf{x'})) ].
\end{align}
The convariance function is also called kernel function: $K:X \times X\xrightarrow{} \mathbb{R}$ which describes the similarity between two observations. It models the assumption that when the observations are similar, the function outputs should also be similar.

Given a new observation $\mathbf{x}^{\ast}$, the GP model provides a probabilistic prediction estimate $p({y}^{\ast}|\mathbf{x}^{\ast},X_{train},Y_{train})$ by marginalizing over the latent function values:
\begin{align}
p({y}^{\ast}|\mathbf{x}^{\ast},X_{train},Y_{train}) = \int_{\mathbb{R}} p({y}^{\ast}|h^{\ast})p(h^{\ast}|\mathbf{x}^{\ast},X_{train},Y_{train}) d h^{\ast} \\
= \int_{\mathbb{R}} p({y}^{\ast}|h^{\ast})(\int_{\mathbb{R}}p({h}^{\ast}|\mathbf{x}^{\ast},\mathbf{h})p(\mathbf{h}|X_{train},Y_{train}) d \mathbf{h}) d h^{\ast}
\end{align}
where $\mathbf{h} =\{h(\mathbf{x}_i)\}_{i=1}^{n}$ is the latent function values of the training set $X_{train}$.

Incorporating the noise model and the assumption that all samples are drawn i.i.d into the Bayes rule yields \cite{rasmussen2006gaussian}:
\begin{align}
p(\mathbf{h}|X_{train},Y_{train}) \propto 
p(\mathbf{h}|X_{train})p(Y_{train}|X_{train},\mathbf{h})\\
= p(\mathbf{h}|X_{train})\prod_{i}p(y_i|h(\mathbf{x}_i)).
\end{align}
$p(\mathbf{h}|X_{train})$ is drawn from the GP prior. Depending on the output $y_i$, the noise model $p(y_i|h(\mathbf{x}_i))$ can take diffrent forms for the task of regression and classification.
\subsection{Gaussian Process Regression}
GP regression (GPR) links the output $y$ and the latent function $h$ with a Gaussian noise model:
\begin{align}
p(y_i|h(\mathbf{x}_i)) = \mathcal{N}(h_i,\sigma^{2}).
\end{align}
The joint distribution between the output value at point $\mathbf{x}^{\ast}$ and all the training samples and the training samples (denoted as $\mathbf{y} =\{y_i)\}_{i=1}^{n}$) can be written as \cite{rasmussen2006gaussian}: 
\begin{align}
\begin{bmatrix}
\mathbf{y} \\ h^{\ast}
\end{bmatrix} \sim \mathcal{N}(\mathbf{0},
\begin{bmatrix}
K(X_{train},X_{train})+\sigma^{2}I & K(X_{train},\mathbf{x}^{\ast})\\
K(\mathbf{x}^{\ast},X_{train}) &
K(\mathbf{x}^{\ast},\mathbf{x}^{\ast})
\end{bmatrix}.
\end{align}
The inference of $y^{\ast}$ is in closed form:
\begin{align}
\bar{y}^{\ast}(\mathbf{x}^{\ast}) = K(\mathbf{x}^{\ast},X_{train})^{T}[K(X_{train},X_{train})+\sigma^{2}I]^{-1}\mathbf{y}
\end{align}
\begin{align}\label{equ:gp_variance}
\mathbb{V}(y^{\ast}(\mathbf{x}^{\ast}) = K(\mathbf{x}^{\ast},\mathbf{x}^{\ast}) -K(\mathbf{x}^{\ast},X_{train})^{T}[K(X_{train},X_{train})+\sigma^{2}I]^{-1}K(X_{train},\mathbf{x}^{\ast})+\sigma^{2}.
\end{align}
\subsection{Gaussian Process Classification}
In case of classification, $Y$ is the target set which contains integers indicating the labels of the input data. In this case, the Gaussian noise model is not always fitted to the nature of discrete label. Instead, other noise models such as cumulative Gaussian or sigmoid function are applied. Exact inference is not tractable, and efficient approximation methods are needed. such as Laplace Approximation and Expectation Propagation   \cite{rasmussen2006gaussian}. 

For multiclass-classification problem, we can either use the multi-label inference, or the one-vs-all (OVA) method. In the latter case, a binary GP classifier (GPC) whose output label is converted to $\{-1,+1\}$ is trained for each of the $N$ labels: $f_n(\cdot)$. Given a new sample $\mathbf{x^{*}}$, each binary classifier predicts the observation probability of its label $p(y_n|\mathbf{x^{*}})$. The sample is assigned to the class with the largest prediction
probability:
\begin{align}
y^{*} = \arg\max_{y_n\in{Y}}\ p(y_n|\mathbf{x^{*}}).
\end{align}
\subsection{Parameters Tuning}\label{subsec:background:gpc_optimization}
The main parameters in GP are kernel parameters, denoted here as $\mathbf{\theta}$. There are many ways to find the optimal $\mathbf{\theta}$ in the GP framework. One way is to maximize the log-marginal likelihood $\text{log}(p(\mathbf{y}|X_{train}))$. In the GPR problem, the log-likelihood can be expressed in closed form \cite{rasmussen2006gaussian}:

\begin{multline}
  \text{log}(p(\mathbf{y}|X_{train},\mathbf{\theta})) = -\frac{1}{2}\mathbf{y}^T(K(X_{train},X_{train})+\sigma^2 I)^{-1}\mathbf{y}\\
  -\frac{1}{2}\text{log}|K(X_{train},X_{train})+\sigma^2 I|-\frac{n}{2}\text{log}(2\pi).
\end{multline}

Another method is the Leave One Out (LOO) cross validation (CV) method. It is the sum of the predictive log probability when leaving out each of the training sample once. The optimal parameters can be found which maximize the LOO value \cite{rasmussen2006gaussian}.
\subsection{Transfer Learning Using Gaussian Process}
In one of the pioneering works about the transfer learning applications in GP framework, Lawrence \textit{et al.} \cite{lawrence2004learning} showed that the hyper-parameters trained jointly by the data from multiple tasks could improve the learning process for each task. Urtasun \textit{et al.} \cite{urtasun2008transferring} assumed that there was a latent feature space shared by many tasks. They first transformed features from each task using the Gaussian Process Latent Variabel Model (GPLVM). Then, the source samples in the new feature space were used to boost the learning process in the target tasks. Bonilla \textit{et al.} \cite{bonilla2007multi} modeled the relationship among tasks by modifying the kernel function as a Kronecker production of the task-related kernels and sample-related kernels. They showed that jointly optimizing all the tasks was highly beneficial for each task (symmetric multi-task learning). Chai \textit{et al.} \cite{chai2009generalization} adapted this method to "asymmetric" multi-task learning in which the source task was used to improve the target task. This work considered only one source task, and the task relationship was simply controlled by a correlation parameter. Rodner \textit{et al.} \cite{rodner2010one} extended this "asymmetric" multi-task learning method for categorization of web images.

\section{Tactile-based Robotic Action-Perception Loop} \label{sec:background:action_perception_loop}
\begin{figure}[!htp]
	\centering
	\begin{minipage}{1\linewidth}
		\centering
		\includegraphics[width=0.6\textwidth]{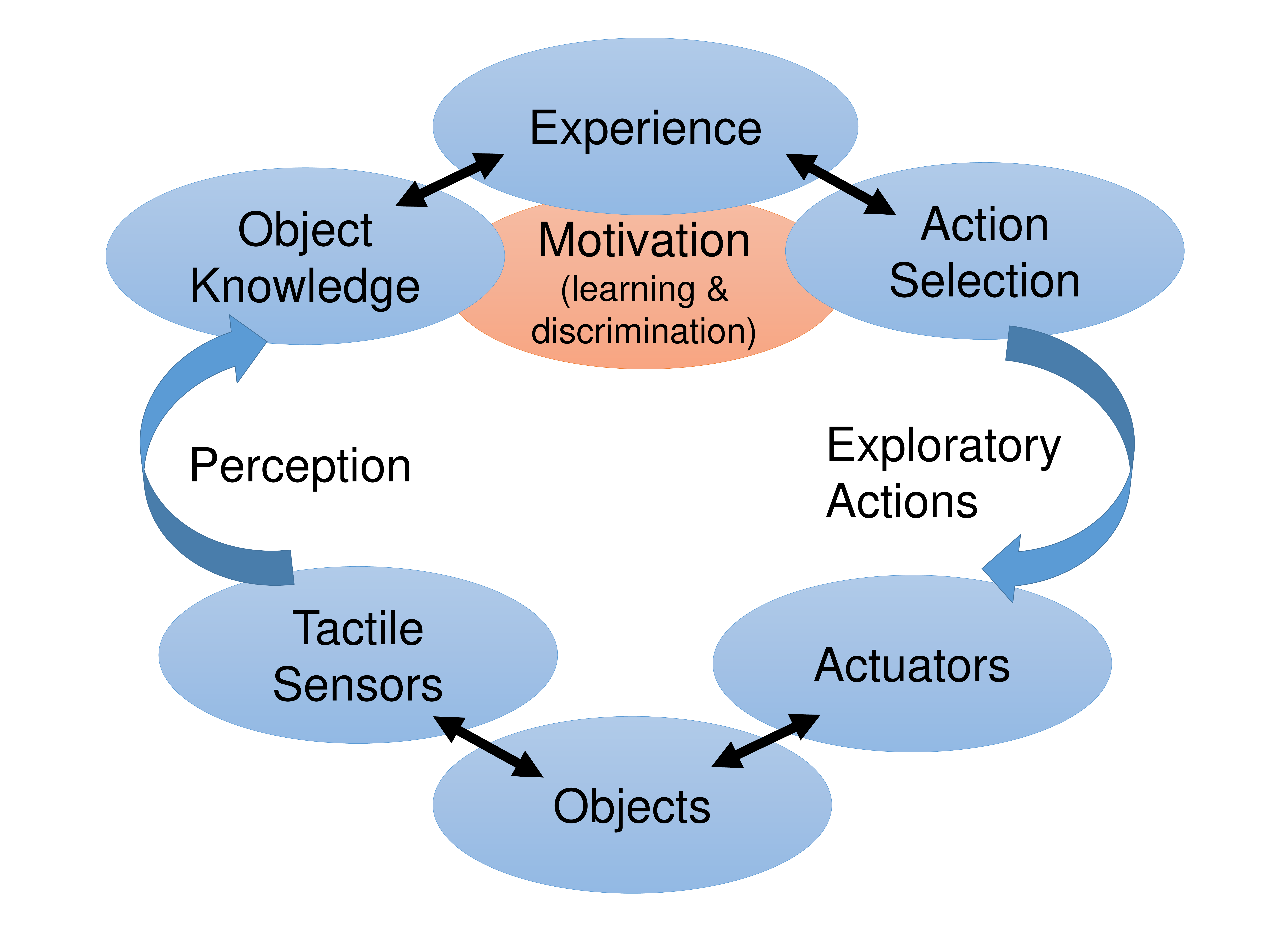}
	\end{minipage}
	\caption[Robotic action perception loop for tactile sensing.]{Robotic action perception loop for tactile sensing (reproduced and modified from \cite{loeb2014bayesian}).}\label{fig:action_perception_loop}
\end{figure}

Equipped with tactile sensors, the robotic system can interact its sensory part with objects using different exploratory actions. Correspondingly, after applying different actions on the objects, the robot perceives different sensory feedbacks, based on which objects can be learned or distinguished from each other. The task of active object exploration can be achieved with the paradigm of "action-perception loop", demonstrated in Fig.~\ref{fig:action_perception_loop}. Following this loop, the robotic system iteratively touches objects in order to perceive feature observations. Then, based on the prior knowledge of objects and sensory feedbacks, the robot decides the next exploratory action to apply on objects. This procedure iterates until the robot reaches its target (e.g. discriminating among objects with a certain probability threshold). Examples of using this action-perception loop in the problem of tactile-based active object exploration can be found in \cite{di2017iros,lepora2013active,xu2013tactile,schneider2009object,tanaka2014optimal}.

In the following, we briefly summarize our work that follows the action-perception loop to realize an active object learning task (\cite{di2017iros}), since this master thesis is based on it.

Previously, the robots collected training samples in an uniform and offline manner, in order to build the object observation models to discriminate among objects \cite{lepora2013active,fishel2012bayesian,xu2013tactile,schneider2009object}. However, since some objects are easily confused with each other, while some others are discriminant, uniformly collecting training samples is data inefficient and time-consuming. In this regard, in \cite{di2017iros} we proposed an active learning method so that the robot can build reliable object observation models online with fewer training samples

In this work, the robot can press an object to sense its object stiffness, slide to attain its object textural property, and build static contact to measure  its thermal conductivity. To efficiently construct the object observation models (i.e. to learn about objects) with as few training samples as possible, the robot iteratively selected the next object to explore and the next physical property to perceive \footnote{This process corresponds to the action selection step in the action perception loop (see Fig.~\ref{fig:action_perception_loop}).}. The object observation models were built based on the one versus all Gaussian Process Classification model \cite{di2017iros}. At each learning iteration, the classification competences of the GPC models were estimated to guide the next round of training sample generation. As we can see in Sec.~\ref{sec::method::active_learning}, when updating the prior tactile exploratory action experiences proposed in this master thesis, we use similar approach to collect new feature observations.  

\cleardoublepage
\let\textcircled=\pgftextcircled
\chapter{Robotic System Description}\label{chapter:system_description}

\begin{figure}[!htp]
	\centering
	\begin{minipage}{1\linewidth}
		\centering
		\includegraphics[width=1\textwidth]{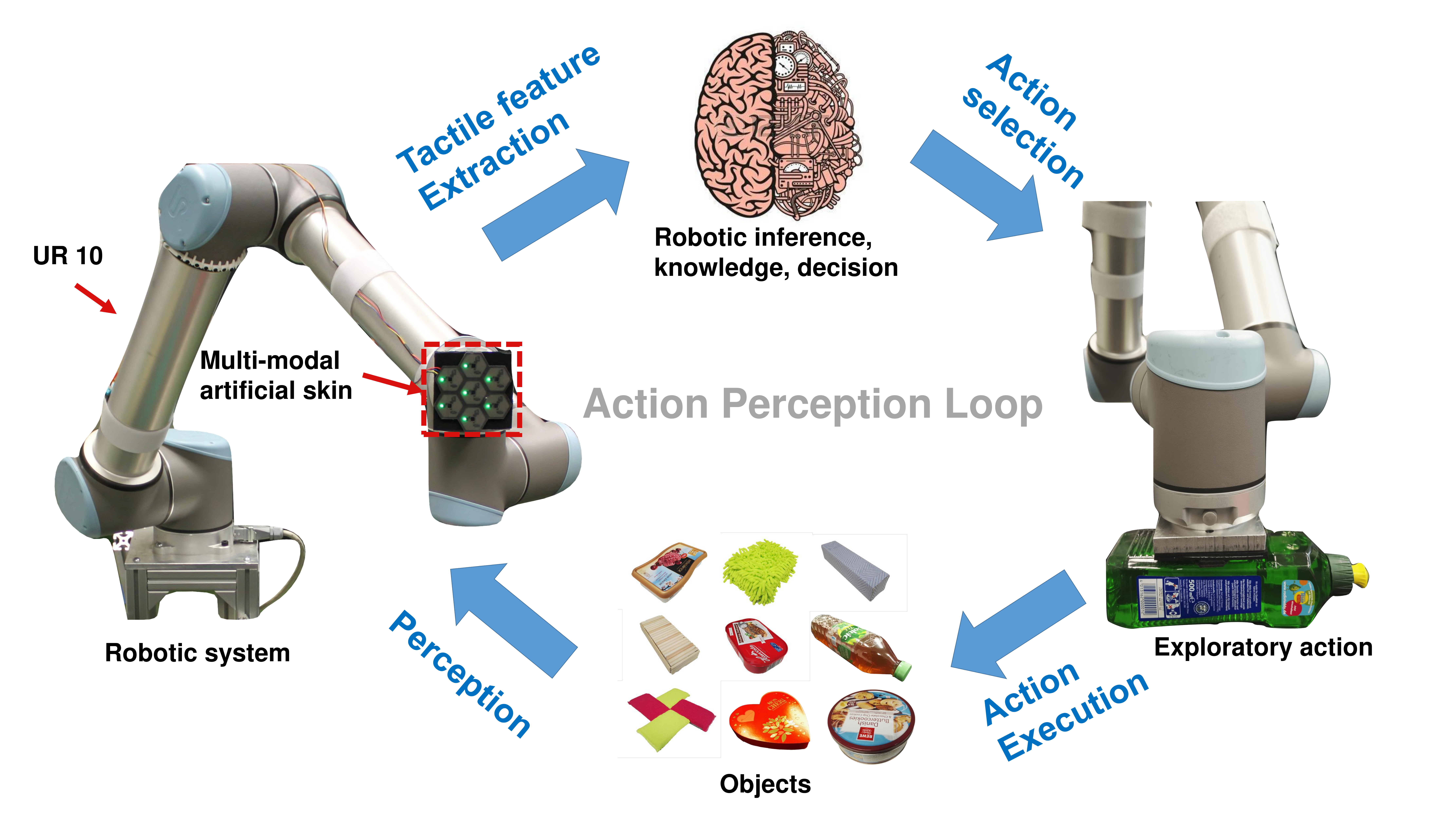}
	\end{minipage}
	\caption[Robotic system.]{The robotic system and the action-perception loop. The UR10 robot is equipped with an artificial skin on its end-effector. By following the action perception loop, the robot senses the object physical properties so that it can learn about or discriminate among objects.}\label{fig:experimental-setup}
\end{figure}

\initial{I}n this work, an Universal Robot (UR10) is equipped with an artificial skin with $7$ skin-cells on its end-effector. Following the \textit{robotic action-perception loop} (Fig.~\ref{fig:experimental-setup}) which was introduced in  Sec.~\ref{sec:background:action_perception_loop}, the robotic system applies different exploratory actions with different action parameters to transfer the prior tactile knowledge for learning the detailed physical properties of new objects. In this chapter, we first introduce the UR10 robotic arm, including its technical information and control information. Afterwards, we introduce the artificial skin with a brief description of its sensing modalities and the technical information. Finally, we illustrate how to control UR10 and the artificial skin to realize the action-perception loop (Fig.~\ref{fig:experimental-setup}).

\section{Universal Robot 10}\label{sec:system_description:ur10}
The UR10 robotic arm is one of the three main products (the other two: UR3 and UR5) from the company Universal Robots. It is the biggest robots among UR3 and UR5, with $6$ joints, $10$ kilos lifting capability, $28$ kilos weights, $1300\ \text{mm}$ working radius and $+/- 0.1\ \text{mm}$ repetitions \cite{UR10_tech}. The UR robots are collaborative robots, as they can work next to the personnel without safety guarding. 

The UR10 controller is developed with the PolyScope grphical user interface, which can be either programmed directly with touchpad or by script programming. It offers the communication protocol of TPC/IP, Profinet, Modbus TCP and Ethernet Socket \cite{UR10_tech}. UR10 controller provides servers to send robot state data (e.g. real-time joint space positions and end-effector positions) and receive URScript commands \cite{UR10_tech}.

There are mainly two ways of controlling the UR10 robot: (1). Control Box. UR10 is provided with a control box, with $10$ digital and $2$ analog I/O ports \cite{UR10_tech}. The robotic movements can be directly controlled via the control box. (2). ROS package. UR10 can also be remotely controlled based on Robot Operating System (ROS) framework via socket connection. Official packages have been released for the communication with UR10 controllers, controlling the robotic arm via MoveIt package \footnote{\textit{http://moveit.ros.org/}.}, and visualization based on \textit{rviz} \footnote{\textit{http://wiki.ros.org/rviz}.}. 

\section{Multi-modal Artificial Skin}\label{sec:system_description:skincell}
\subsection{A Brief Introduction}
\begin{figure}[!htp]
	\centering
	\begin{minipage}{1\linewidth}
		\centering
		\includegraphics[width=0.85\textwidth]{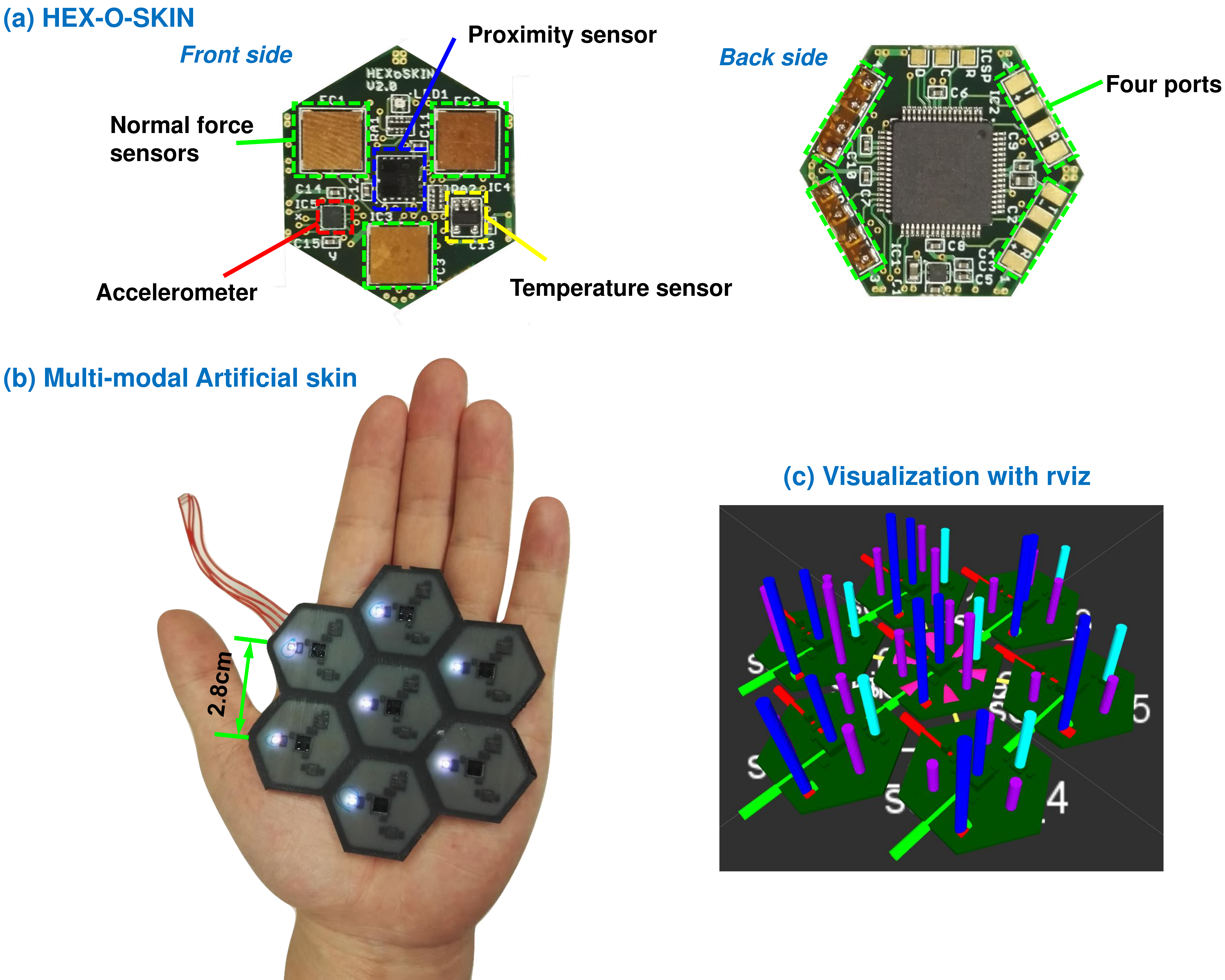}
	\end{minipage}
	\caption[Multi-modal artificial skin.]{Multimodal artificial skin and HEX-O-SKIN (skincell). \textbf{(a).} the front and the back sides of the skincell. \textbf{(b).} the artificial skin which consists of seven skincells. \textbf{(c).} the rviz visualization of the artificial skin using rviz toolbox under the ROS framework (red, green and blue bars show the accelerometer signals in x,y,z axes; the lila bar shows the proximity sensing signals.).  }\label{fig:artificial-skin}
\end{figure}

To enable robotic systems to perform more human-like behaviours, it is necessary a variety of sensing modalities, so that they can have more interactions with the environment. In this master thesis, we use an artificial skin made by seven active tactile modules called "HEX-O-SKIN", or "skincell". Each skincell is a small hexagonal printed circuit board equipped with off-the-shelf sensors (temperature sensor, proximity sensor, accelerometers and normal force sensors) \cite{mittendorfer2011humanoid} (Fig.~\ref{fig:artificial-skin}). In this way, robots with this artificial skin can emulate the human tactile sensing of temperature, light touch, vibrations, and force. 
\subsection{Sensing Modalities and Their Technical Information}\label{sec:system:artificial_skin_sensing}
Our artificial skin has seven skincells. Each skincell is equipped with one proximity sensor, one temperature sensor, one accelerometer and three normal force sensors (Fig.~\ref{fig:artificial-skin}). There are in total seven proximity sensors, seven temperature sensors, seven accelerometers and 21 normal force sensors. Their technical information is summarized in  Tab.~\ref{tab:sensors_skin}.

\begin{table}[!htp]
    \centering
    \begin{tabular}{ c c c c c c } 
     \hhline{======}
     Sensing modality & Sensor & Range & Accuracy & Resolution & \# per cell\\ 
    
     Proximity & VCNL4010 & $200 \text{mm}$ & N.A. & $0.25$lx & $1$ \\
     Acceleration & BMA250 & $\pm2 \text{g}$ & $256 \text{LSB/g} $ & $3.91$ mg & $1$\\
     Temperature & LM71 & $-40 \ - 150^\circ C$ & $ \pm1.5^\circ C $ & $31.25 \text{m}^\circ C$ & $1$ \\
    Normal force & customized & $>10\text{N}$ & $0.05\text{N}$ &  N.A. & $3$\\
    \hhline{======}
    \end{tabular}   
    \caption[Technical information of sensors in the artificial skin.]{Technical information of sensors in the artificial skin (\cite{kaboli2014humanoids}).}
    \label{tab:sensors_skin}
\end{table}

The proximity sensors use the optical reflection from the infrared signals to estimate its distance to objects.Their sensory feedbacks are strong when the distance is small, but degrades drastically as the distance increases. Therefore, we can use the proximity sensing to emulate the lightest touch from the human skin \cite{mittendorfer2011humanoid}, which is especially useful in the motion control of robotic systems. It is noteworthy to mention that since the reflections of infrared signals are influenced by colors and materials, the distance estimates from proximity sensors are highly dependent on the object surface properties.  

The temperature sensing provides useful information about the environment. When contacting the temperature sensor with objects, it measures the change of temperature, with which the object thermal conductivity can be inferred. 

Accelerometers can be used to sense vibrations and orientations. The vibrations that are generated when a robot slides its artificial skin on different object surfaces can be used to identify the textures \cite{Kaboli-TRO2017} or estimate slippage \cite{Kunpeng-Hu2017,Kaboli-Humanoid16}.  

The normal force sensors provide us the information of contact pressure. It can be either used for robust human-robot-interaction, or to detect object hardness. 
\subsection{Controlling the Artificial Skin}
The artificial skin provides the standard ROS interface with rviz visualization. Fig.~\ref{fig:rosgraph} shows the ROS graph when driving the artificial skin. The rviz helps to visualize the tactile signals in real-time (Fig.~\ref{fig:artificial-skin}).
\section{Controlling the Robotic System via ROS}
\begin{figure}[!htp]
	\centering
	\begin{minipage}{1\linewidth}
		\centering
	\includegraphics[width=0.98\textwidth]{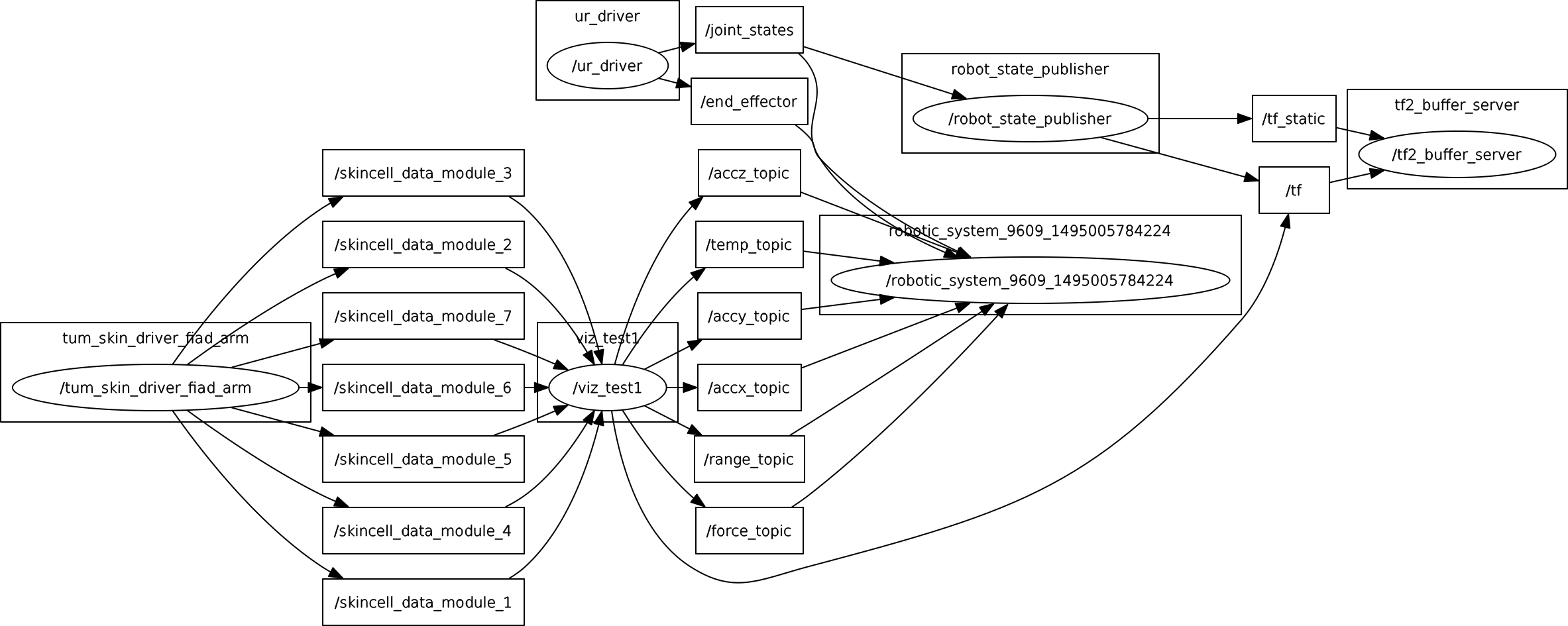}
	\end{minipage}
	\caption[ROS graph for controlling the robotic system.]{ROS graph for controlling the robotic system.}\label{fig:rosgraph}
\end{figure}

In this work, UR10 and the artificial skin were remotely controlled by PC based on the ROS framework and Python scripts. The ROS graph is shown in Fig.~\ref{fig:rosgraph}. The tactile signals (proximity sensing, temperature, acceleration and normal force) were pre-processed by the rosnode \textit{vis\_test1} and were published by the ros-node \textit{tum\_skin\_driver\_fiad\_arm}. The ros-node \textit{ur\_driver} published the end-effector position and joint positions of UR10 in 3D cartesian coordinate. The main ros-node (\textit{robotic\_system\_driver}) subscribed the ros-topics both from the artificial skin and the robotic arm. It analyzed the tactile feedback and controlled the next exploratory action that the robot would apply on objects.\\

\noindent \textbf{Controlling UR10}. In this work, UR10 is controlled based on the end-effector position in the Cartesian coordinate. We use \textit{movel(\ )} function from the URScript commands \cite{UR10_tech} to realize the linear movement of the UR10 end-effector. Given the target end-effector position $\mathbf{p}_t=[x_t,y_t,z_t]$, the robot compares it with the current end-effector position $\mathbf{p}_c=[x_c,y_c,z_c]$ and decides whether the target position is reached or not. The target position is reached when the robot current position to it was less than $1$mm in each axis. Alg.~\ref{alg:ur_10_control} demonstrates how the robot was controlled. 

\begin{algorithm}[!htp]
	\caption{UR linear position control}
	\label{alg:ur_10_control}
	\SetKwInOut{Input}{Input}
	\SetKwInOut{Output}{Output}	
	
	\Input{$\mathbf{p}_t=[x_t,y_t,z_t]$ \Comment{End-effector target position}} 
	\textbf{Initialization}: $\mathbf{p}_c \gets positionSubscriber()$ 
	\Comment{Get the end-effector current position $\mathbf{p}_c=[x_c,y_c,z_c]$.}\;
	
	\While{$|x_t-x_c|>1$mm or $|y_t-y_c|>1$mm or $|z_t-z_c|>1$mm} 
	{
	    $movel(\mathbf{p}_t)$ \Comment{Linear moving the end-effector to the target position}\;
	    $\mathbf{p}_c \gets positionSubscriber()$ \Comment{Update the current end-effector position}\;
	}
\end{algorithm}
\cleardoublepage
\let\textcircled=\pgftextcircled
\chapter{Exploratory Action And Perception}\label{chapter:action_perception}
\initial{I}n this chapter, we first illustrate how the robotic system can apply the pressing, sliding, and static contact movements with different action parameters to attain the sensory feedbacks (Sec.~\ref{sec:action_perception:exploratory_action}). Afterwards, we explain how the feature observations from these sensory feedbacks are extracted (Sec.~\ref{sec:action_perception:perception}). 
\section{Exploratory Action Definition} \label{sec:action_perception:exploratory_action}
When applying different exploratory actions on an object, the robot can perceive its different physical properties. When applying the \textit{same} action but with different action parameters, the robot can perceive different feature observations of a physical property. Therefore, the robot can build object tactile knowledge by applying different exploratory actions with different action parameters. In this work, we consider three types of actions: \textit{pressing} (denoted as $P$), \textit{sliding} (denoted as $S$), and \textit{static contact} (denoted as $C$). Note that as each type of exploratory action can be defined with different action parameters, in the rest of the this thesis, we will consider a type of exploratory action with different action parameters as \textit{different} exploratory actions. 

Formally, we define $N_a$ number of exploratory actions as $A=\{\mathbf{\alpha}_n^{\bm{\theta}_n}\}_{n=1}^{N_{\mathbf{\alpha}}}$, where $\bm{\theta}_n$ is the action parameters that define "how" the robot can apply the exploratory action. We further define $\bm{\theta}=\{\bm{\theta}_P,\bm{\theta}_S,\bm{\theta}_C\}$, where $\bm{\theta}_P,\bm{\theta}_S$ and $\bm{\theta}_C$ represent the action parameters for the pressing, sliding, and static contact movements respectively.

\subsection{Pressing, Sliding, and Static Contact} \label{subsec:action_perception:actions}
\subsubsection{Pressing}
\begin{figure}[!htp]
	\centering
	\begin{minipage}{1\linewidth}
		\centering
		\includegraphics[width=1\textwidth]{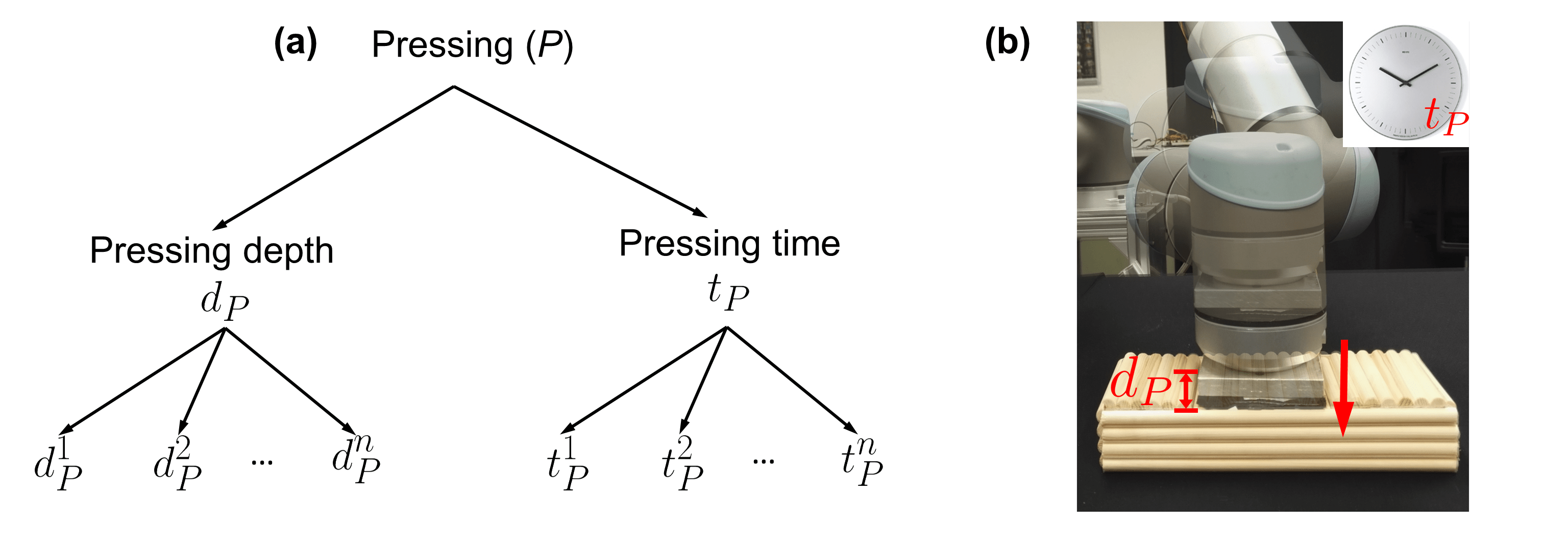}
	\end{minipage}
	\caption[Visualization of the pressing movement.]{\textbf{(a)} the action parameters related to the pressing movement. \textbf{(b)} visualization of the pressing movement.} \label{fig:pressing}
\end{figure}

A robotic system presses against object surfaces in order to perceive stiffness. The pressing movement consists of pressing until a depth of $d_{P}$ and holding the artificial skin for $t_{P}$ seconds, i.e. $\bm{\theta}_P = [d_P,t_P]$. During the pressing, the multi-modal artificial skin can record the normal force feedbacks from each normal force sensor: $\mathbf{F}_{n_f,n_s} = \{F_{n_f,n_s}^{m}\}_{m=1}^{t_P\cdot f_s}$ in order to measure the object stiffness. $N_f$ is the number of normal force sensors in one skincell (in our case $N_f=3$), and $N_s$ is the number of skincells in the artificial skin (in our case $N_s=7$). Moreover, it can also record the temperature feedbacks from each temperature sensor, for the purpose of attaining the objects' thermal conductivity: $\mathbf{T}_{n_t,n_s} = \{T_{n_t,n_s}^{m}\}_{m=1}^{t_P\cdot f_s},\ n_t = 1,..., N_t$, with $N_t$ being the number of temperature sensors in one skincell (in our case $N_t=1$). $f_s$ is the sampling rate of the artificial skin, and $m$ is the sampling time step. The pressing movement is visualized in Fig.~\ref{fig:pressing}. An signal sequence of normal force during pressing is demonstrated in Fig.~\ref{fig:sensor_signals} (a).

\subsubsection{Sliding}
\begin{figure}[!htp]
	\centering
	\begin{minipage}{1\linewidth}
		\centering
		\includegraphics[width=1\textwidth]{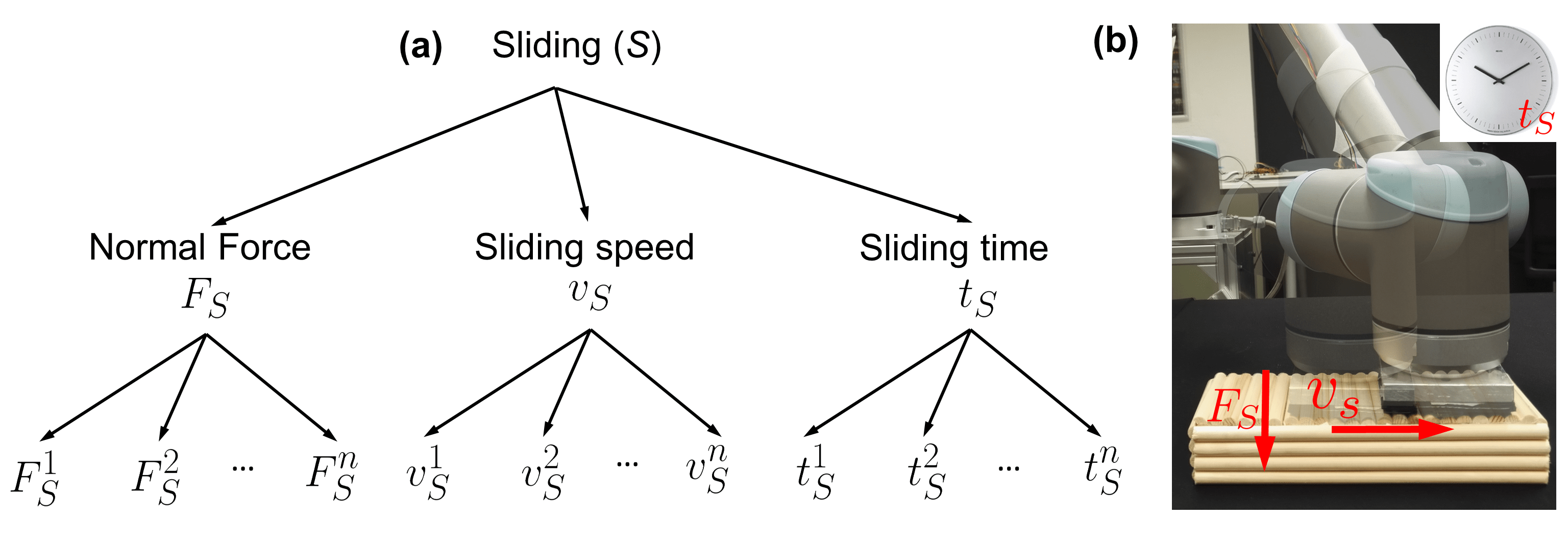}
	\end{minipage}
	\caption[Visualization of the sliding movement.]{\textbf{(a)} the action parameters related to the sliding movement. \textbf{(b)} visualization of the sliding movement.} \label{fig:sliding}
\end{figure}

A robot slides the artificial skin on object surfaces and perceives textural properties. To do this, the robot first builds a contact with objects with a normal force of $F_S$, then it linearly slides on the objects with a speed of $v_S$ for $t_S$ seconds, $\bm{\theta}_S = [F_S,v_S,t_S]$. During sliding, the robot collects the outputs of accelerometers (in three axes: $x,y,z$): $\mathbf{a}^{(x)}_{n_a,n_s} = \{a^{(x),m}_{n_a,n_s}\}_{m=1}^{t_S\cdot f_s}$, $ \mathbf{a}^{(y)}_{n_a,n_s} = \{a^{(y),m}_{n_a,n_s}\}_{m=1}^{t_S\cdot f_s}$, $ \mathbf{a}^{(z)}_{n_a,n_s} = \{a^{(z),m}_{n_a,n_s}\}_{m=1}^{t_S\cdot f_s}$.
Then it combines these signals together: $\mathbf{a} = \{\mathbf{a}_{n_a,n_s}\}_{n_a=1,n_s=1}^{N_a,N_s}$; $\mathbf{a}_{n_a,n_s} = [\mathbf{a}^{(x)}_{n_a,n_s},\mathbf{a}^{(y)}_{n_a,n_s},\mathbf{a}^{(z)}_{n_a,n_s}], n_a = 1,..., N_a$, where $N_a$ is the number of accelerometers in one skincell (in our case $N_a=1$). Besides, the change of temperature during sliding is also collected as an extra information $\mathbf{T}_{n_t,n_s} = \{T_{n_t,n_s}^{m}\}_{m=1}^{t_S\cdot f_s}$. Fig.~\ref{fig:sliding} illustrates the sliding movement. An exemplary signal sequence of accelerometers is shown in Fig.~\ref{fig:sensor_signals} (b).

\subsubsection{Static Contact}
The object thermal cues can be attained by applying static contact movement: the robot presses its sensory part against the object surface until a depth of $d_C$ and maintains the contact for $t_C$ seconds. The normal force feedbacks and temperature feedbacks are recorded: $\mathbf{F}_{n_f,n_s} = \{F_{n_f,n_s}^{m}\}_{m=1}^{t_C\cdot f_s}$, $\mathbf{T}_{n_t,n_s} = \{T_{n_t,n_s}^{m}\}_{m=1}^{t_C\cdot f_s}$, i.e. $\bm{\theta_C} = [d_C,t_C]$.

\begin{figure}[!htp]
	\centering
	\begin{minipage}{1\linewidth}
		\centering
		\includegraphics[width=1\textwidth]{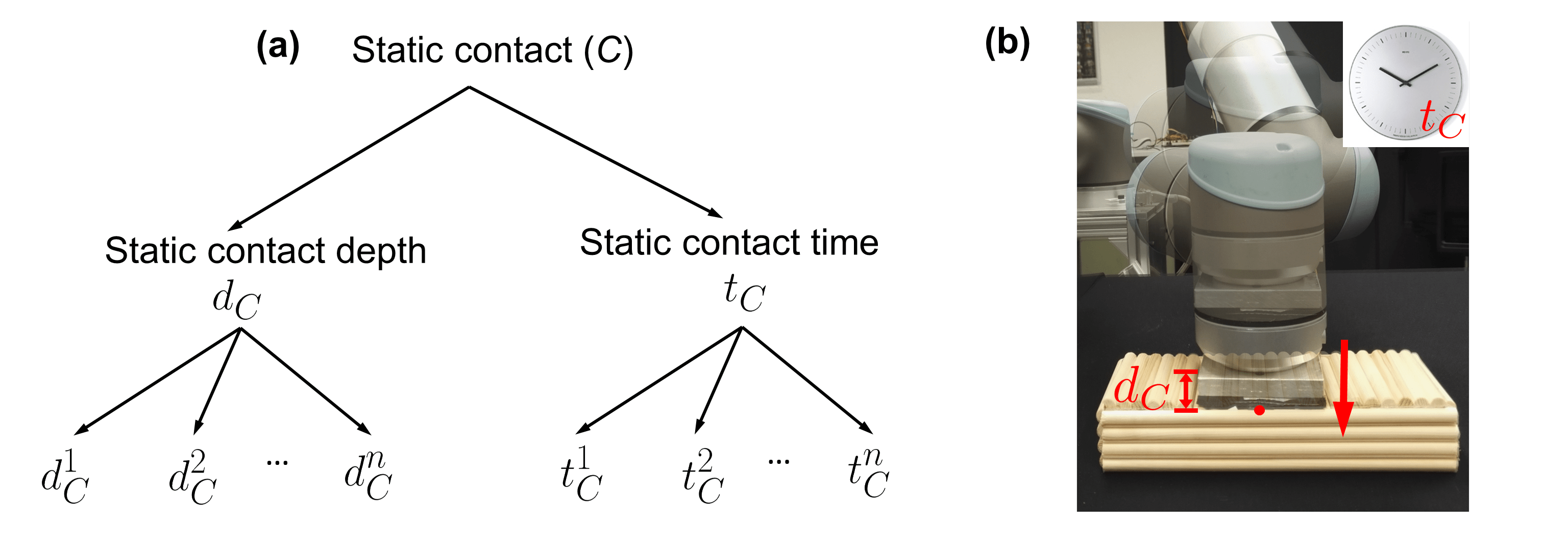}
	\end{minipage}
	\caption[Visualization of the static contact movement.]{\textbf{(a)} the action parameters related to the static contact movement. \textbf{(b)} visualization of the static contact movement.} \label{fig:static_contact}
\end{figure}

\begin{figure}[!htp]
	\centering
	\begin{minipage}{1\linewidth}
		\centering
		\includegraphics[width=1\textwidth]{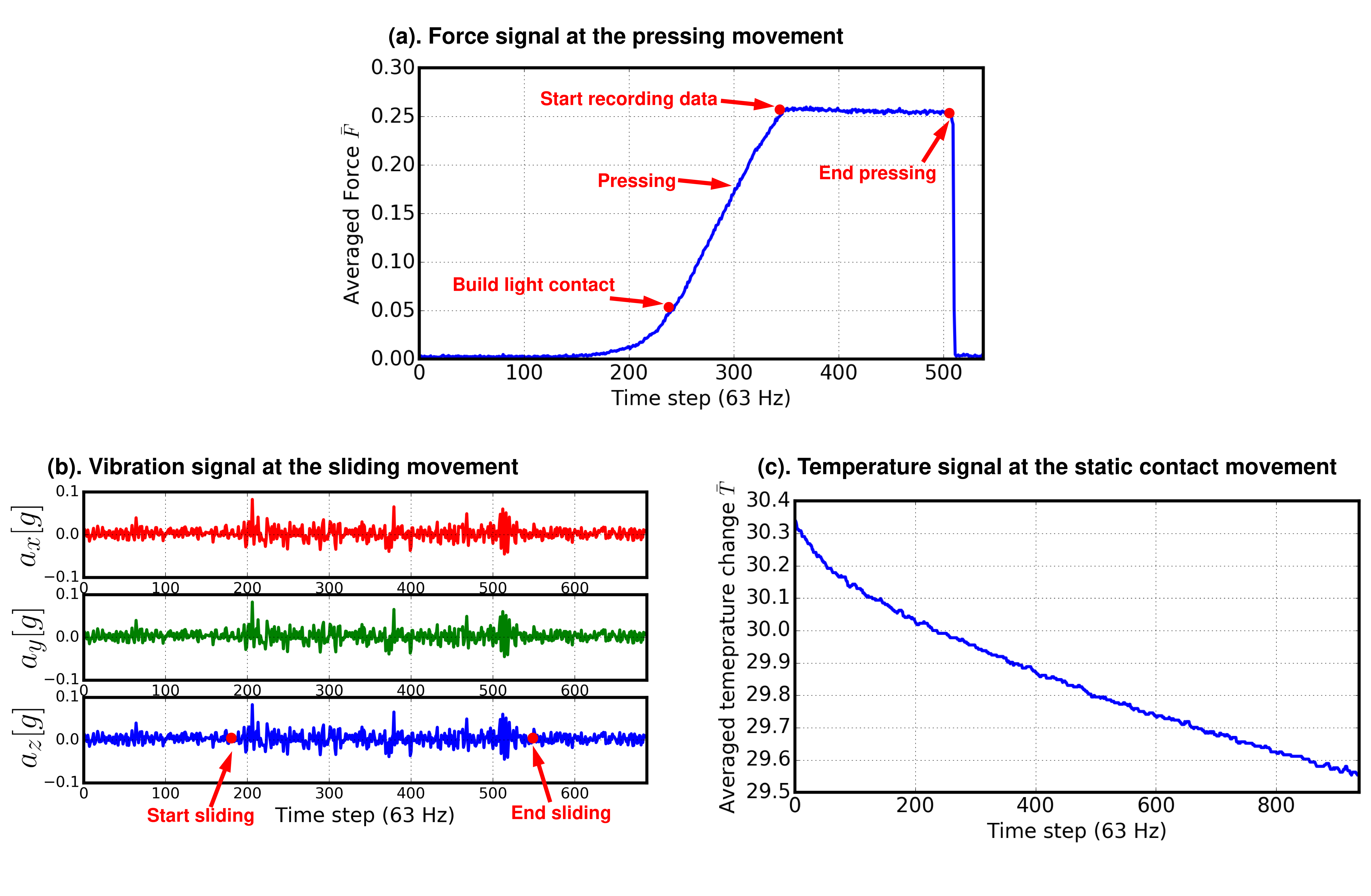}
	\end{minipage}
	\caption[Visualization of sensor signals.]{The sensor signals perceived during applying the exploratory actions. \textbf{(a)} The normal force signal, when the robot built a light contact as the sum of all normal force sensor values exceeded $0.05$ N, pressed its artificial skin $2$ mm deep and held for $3$ s. \textbf{(b)} The acceleration signals in three directions ($x,y,z$), as the robot slid the object surface with horizontal speed of $1$ cm/s for $5$ s, the contact force was set to be $0.2$ N. \textbf{(c)} An example of the change of temperature signal during static contact. The robot pressed its artificial skin $2$ mm deep and recorded the data for $15$ s.} \label{fig:sensor_signals}
\end{figure}

Fig.~\ref{fig:static_contact} shows the static contact movement. An example of the temperature signals is shown in Fig.~\ref{fig:sensor_signals} (c).

\newpage
\section{Object Physical Properties Perception}\label{sec:action_perception:perception}
In this section, we illustrate how a robotic system can perceive multiple object physical properties and extract corresponding tactile features.

\subsection{Stiffness}
We use the averaged normal force value that are calculated from all normal force sensors and time steps as an indicator for the object stiffness. For the pressing movement with pressing time steps $t_P\cdot f_s$, object stiffness can be estimated by $\bar{F}$, with:
\begin{equation}
   \bar{F} = \frac{1}{t_P\cdot f_s} \frac{1}{N_f} \frac{1}{N_s}
\sum\nolimits_{m=1}^{t_P\cdot f_s}
\sum\nolimits_{n_f=1}^{N_f} \sum\nolimits_{n_s=1}^{N_s}  F_{n_f,n_s}^{m}.
\end{equation}

\subsection{Textural Property}
In \cite{kaboli2014humanoids}, a set of feature descriptors was proposed to describe the object textural property, namely activity ($Act$), mobility ($Mob$), and complexity ($Comp$). Given a sequence of vibration signals during the sliding movement $x(m)$ ($m=1,2,3...,M$), the feature descriptors were calculated by:  
\begin{align}
    Act(x) = \frac{1}{M} \sum\nolimits_{m=1}^M {(x(m) - \bar{x})}^2, \\
    Mob(x) = \sqrt{\frac{Act(\frac{dx(m)}{dm})}{Act(x)}}, \\
    Comp(x) = \frac{Mob(\frac{dx(m)}{dm})}{Mob(x)}.
\end{align}

In this work, we employ the same textural feature extraction method used in \cite{di2017iros}. The vibration signals in three directions $x,y,z$ from each accelerometer in the artificial skin are used to calculate the \textit{averaged} activity, mobility and complexity, denoted as $Act(\mathbf{a})$, $Mob(\mathbf{a})$, $Comp(\mathbf{a})$. We also take the relationships between different signals into account. Given two sequence of signals $x(n)$ and $y(n)$, their linear correlation can be computed as:
\begin{equation}
   Lcorr(x,y) \ =  \frac{\sum_{m=1}^{M} { (x(m) - \bar{x})(y(m)-\bar{y})}} {{\sigma(x)\sigma(y)}}. 
\end{equation}
We compute the averaged linear correlation of accelerometer signals between different directions ($xy$,$yz$,$xz$), denoted as $Lcorr(\mathbf{a})$.
The final descriptor of textural feature combines activity, mobility, complexity and linear correlation together \cite{di2017iros}:
\begin{equation}
[Act(\mathbf{a}),Mob(\mathbf{a}),Comp(\mathbf{a}),Lcorr(\mathbf{a})].
\end{equation}

\subsection{Thermal Conductivity}
We use the data-driven approach to extract the features that describe the object thermal cues \cite{bhattacharjee2015material}. To do this, we first calculate the average temperature sequence from all temperature sensors:
\begin{equation}
   \bar{\mathbf{T}} = \sum\nolimits_{n_t=1}^{N_t} \sum\nolimits_{n_s=1}^{N_s} \frac{\mathbf{T}_{n_t,n_s}}{N_t\cdot N_s}.
\end{equation}
We then calculate its gradient at each time step as: $\nabla \bar{\mathbf{T}}$, and combine it with the average temperature sequence: $[\bar{\mathbf{T}},\nabla \bar{\mathbf{T}}]$. To avoid the curse of dimensionality, we further reduce this combination to $10$ dimensions via the Principle Component Analysis (PCA) method and use it as the final feature to describe the object thermal conductivity. 

Tab~\ref{tab:action+signal}. summarizes the exploratory actions, the sensory feedbacks and the corresponding tactile features.

\begin{table}[!htp]
    \centering
    \begin{tabular}{  c c c c c c  } 
    \hhline{======}
     Exploratory actions & Action Parameters ($\bm{\theta}$) & Sensory feedbacks & Features\\ 
    \hline
    Pressing & $d_P$, &  $\mathbf{F}_{n_f,n_s}$,  & $\bar{F}$, \\
    & $t_P$. & $\mathbf{T}_{n_t,n_s}$. & $[\bar{\mathbf{T}},\nabla \bar{\mathbf{T}}]$.\\
    \hline
     & &  & $Act(\mathbf{a})$,\\
    Sliding &$F_S$, & $\mathbf{a}^{(x)}_{n_a,n_s}$,  $\mathbf{a}^{(y)}_{n_a,n_s}$, & $Mob(\mathbf{a})$,\\
    & $t_S$, &  & $Comp(\mathbf{a})$, \\
    & $v_S$.& $\mathbf{a}^{(z)}_{n_a,n_s}$, $\mathbf{T}_{n_t,n_s}$. & $[\bar{\mathbf{T}},\nabla \bar{\mathbf{T}}]$, \\
    &&& $Lcorr(\mathbf{a})]$.\\
    \hline
    Static contact & $d_C$, &  $\mathbf{F}_{n_f,n_s}$,  & $\bar{F}$, \\
    & $t_C$. & $\mathbf{T}_{n_t,n_s}$. & $[\bar{\mathbf{T}},\nabla \bar{\mathbf{T}}]$.\\
    \hhline{======}
    \end{tabular}   
    \caption[Exploratory actions, sensory feedbacks and corresponding tactile features.]{Exploratory actions, sensory feedbacks and corresponding tactile features.}
    \label{tab:action+signal}
\end{table}
\cleardoublepage
\let\textcircled=\pgftextcircled
\chapter{Leveraging Prior Tactile Exploratory Action Experiences}\label{chapter:method}
\begin{figure}[!htp]
	\centering
	\begin{minipage}{1\linewidth}
		\centering
		\includegraphics[width=0.75\textwidth]{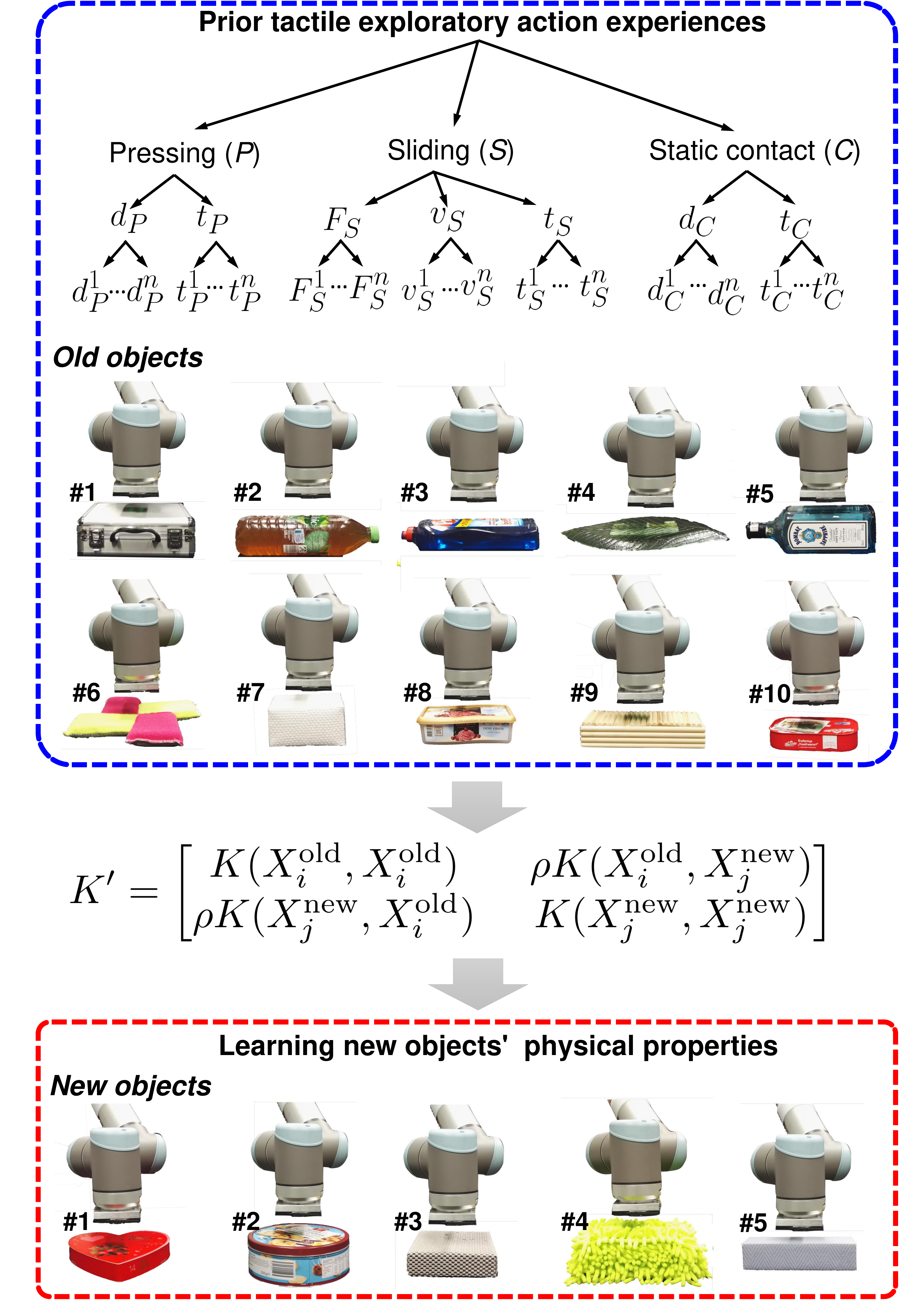}
	\end{minipage}
	\caption[Visualization of our proposed transfer learning algorithm.]{Visualization of our transfer learning algorithm that leverages the prior tactile exploratory action experiences to learn the physical properties of new objects.} \label{fig:TL-visualization}
\end{figure}

\initial{T}his chapter describes our proposed transfer learning method (Active Tactile Instance Knowledge Transfer: ATIKT) in detail. First, we formulate our problem in Sec.~\ref{sec::method::TL-problem}. Then, we discuss what to transfer (Sec.~\ref{sec:method:sensor_fusion}), how to transfer (Sec.~\ref{sec::method::how_to_transfer}), from where to transfer and how much to transfer (Sec.~\ref{sec::method::where_to_transfer}). The motivation of our method is demonstrated in Fig.~\ref{fig:TL-visualization}.
\section{Problem Formulation} \label{sec::method::TL-problem}
Assume that a robotic system has gained the prior tactile knowledge (or prior tactile exploratory action experiences) of some objects which we refer to as \textit{old} objects, when it applied different exploratory actions with different action parameters on them. The tactile knowledge consists of the feature observations perceived by the multiple sensors (tactile instance knowledge) and the observation models of the old objects (tactile model knowledge). Now, the robot is tasked to learn about a set of \textit{new} objects. Since the old objects might share some similar physical properties with the new objects, the robot can learn about new objects more efficiently by leveraging prior tactile knowledge. 

We define $N_{\text{new}}$ number of new objects ($O^{\text{new}}=\{o_i^{\text{new}}\}_{i=1}^{N_{\text{new}}}$) the robot is tasked to explore through different exploratory actions  $A=\{\mathbf{\alpha}_n^{\bm{\theta}_n}\}_{n=1}^{N_{\mathbf{\alpha}}}$ (For simplicity, we will denote $\mathbf{\alpha}$ as an exploratory action in the rest of the paper). The robot should actively attain object feature observations ($X_{\mathbf{\alpha}}^{\text{new}}=\{X^{\text{new}}_{o_1},X^{\text{new}}_{o_2},...,X^{\text{new}}_{o_{N_{\text{new}}}}\}$) for each exploratory action $\mathbf{\alpha}$ and construct reliable observation models $X_{\mathbf{\alpha}}^{\text{new}} \xrightarrow{\mathbf{f}_{\mathbf{\alpha}}^{\text{new}}}O^{\text{new}}$. We further define prior tactile experiences of an exploratory action $\mathbf{\alpha}$ for $N_{\text{old}}$ number of old objects ($O^{\text{old}}=\{o_i^{\text{old}}\}_{i=1}^{N_{\text{old}}}$) as their feature observations $X_{\mathbf{\alpha}}^{\text{old}}=\{X^{\text{old}}_{o_1},X^{\text{old}}_{o_2},...,X^{\text{old}}_{o_{N_{\text{old}}}}\}$ (prior tactile instance knowledge) and their observation models $X_{\mathbf{\alpha}}^{\text{old}} \xrightarrow{\mathbf{f}_{\mathbf{\alpha}}^{\text{old}}}O^{\text{old}}$ (prior tactile model knowledge). The feature observations are collected by the multiple tactile sensors from the artificial robotic skin. 

We hereby formulate our problem as a transfer learning in the Gaussian Process Classification (GPC) framework (\cite{rasmussen2006gaussian}), where each object is regarded as a class, and for each exploratory action, an one-vs-all GPC model is built as the observation model. The robot iteratively applies exploratory actions and leverages \textit{prior tactile instance knowledge} to improve the GPC observation models of new objects.
\section{Process} \label{sec:method:TL_process}
The robot following ATIKT first applies each of the exploratory actions one time on each new object, in order to collect a small number of feature observations $X^{\text{new}} = \{X^{\text{new}}_{\mathbf{\alpha}_n}\}_{n=1}^{N_{\mathbf{\alpha}}}$ (Initial data collection). Then, the robot reuses its prior tactile instance knowledge to improve the observation models of \textit{each} new object (Initial prior knowledge transfer). During this process, the robot compares the relatedness between the old objects and the new objects (Sec.~\ref{sec::method::where_to_transfer}), and chooses the most related one to transfer its prior feature observations $X^{\text{old}}$  (Sec.~\ref{sec::method::how_to_transfer}).

Afterwards, the robot begins to iteratively collect and combine the feature observations and update the prior tactile knowledge in order to improve the new objects' observation models. At each iteration of prior tactile knowledge updating, the robot (1). actively selects the next object and the next exploratory action in order to attain a new feature observation (Active feature observations collection in Sec.~\ref{sec::method::active_learning}), and (2). updates the prior tactile instance knowledge \textit{only} for the selected exploratory action (Prior knowledge update in Sec.~\ref{sec::method::active_learning}). The iteration terminates when there is no improvement in the observation models of new objects. Our algorithm is demonstrated by Alg.~\ref{alg:ATTL}. 

In the sequel, we first explain how the robot combines feature observations from the multi-modal artificial skin to build the prior tactile instance knowledge (what to transfer). Then, we describe how it transfers the knowledge to the new objects (how to transfer). Afterwards, we illustrate several ways of finding the most related prior knowledge (from where to transfer) and how much to transfer it. In the end, we show how the robot updates the prior tactile knowledge.

\begin{algorithm}[!htp]
	\caption{Proposed transfer learning}
	\label{alg:ATTL}
	\SetKwInOut{Input}{Input}
	\SetKwInOut{Output}{Output}	
	
	\Input{$O^{\text{new}}=\{o^{\text{new}}_j\}_{j=1}^{N_{\text{new}}}$ \Comment{$N_{\text{new}}$ new objects, each object is regarded as a class.}\;
	
	$A=\{\mathbf{\alpha}_n\}_{n=1}^{N_{\mathbf{\alpha}}}$ \Comment{$N_{\mathbf{\alpha}}$ number of exploratory actions with different action parameters}\;
		
    $X^{\text{old}}=\{X^{\text{old}}_{o_1},X^{\text{old}}_{o_2},...,X^{\text{old}}_{o_{N_{\text{old}}}}\}$, $X^{\text{old}} \xrightarrow{\mathbf{f}^{\text{old}}}O^{\text{old}}$ \Comment{\textbf{what to transfer}: old objects' feature observations and observation models.}}\;
	
	\Output{$X^{\text{new}} \xrightarrow{\mathbf{f}^{\text{new}}}O^{\text{new}}$\Comment{new objects' observation models.}\;
	
	$X^{\text{new}}=\{X^{\text{new}}_{o_1},X^{\text{new}}_{o_2},...,X^{\text{new}}_{o_{N_{\text{new}}}}\}$\Comment{new objects' feature observations.}
	}
	\textbf{Initialization}:\ $X^{\text{new}}$\Comment{initial data collection.}\;

    \ \;
    
	\ \ \ \ \ \ \ \ \ \ \ \ \ \ \ \ \ \ \ \ \ \ \ \ \ \ \ \ \ \ \ \ \ \ \ \ \ \textbf{\textit{Initial prior knowledge transfer}}\;

	\For{$\mathbf{\alpha} \in A$}
	{
		\For{$j=1:N_{\text{new}}$}
		{
		$o_{\mathbf{\alpha}}^{\text{old}^{\ast}} \gets priorsSelection(X_{\mathbf{\alpha}}^{\text{old}},X_{\mathbf{\alpha},o_j}^{\text{new}})$  \Comment{\textbf{Where to transfer} Sec.~\ref{sec::method::where_to_transfer}}\;
		
        $\rho_{\mathbf{\alpha}}^{\ast} \gets relatednessEstimate(o_{\mathbf{\alpha}}^{\text{old}^{\ast}},X_{\mathbf{\alpha}}^{\text{old}},X_{\mathbf{\alpha},o_j}^{\text{new}})$  \Comment{\textbf{How much to transfer} Sec.~\ref{sec::method::where_to_transfer}}\;
        
		$\mathbf{\gamma}_{\mathbf{\alpha}} \gets weightsEstimate( K^{(1)},K^{(2)}...,K^{(M_{\mathbf{\alpha}})})$\Comment{\textbf{what to transfer} Sec.~\ref{sec:method:sensor_fusion}}\;
		
		$\mathbf{f}_{\mathbf{\alpha},o_j}^{\text{new}}(\cdot) \gets updateGPC(X^{\text{new}}_{\mathbf{\alpha},o_j},\rho_{\mathbf{\alpha}}^{\ast},\mathbf{\gamma}_{\mathbf{\alpha}})$ \Comment{\textbf{How to transfer} Sec.~\ref{sec::method::how_to_transfer}} \;	
		}    
	}
    $X^{\text{new}} = \{X^{\text{new}}_{\mathbf{\alpha}_n, o_j}\}_{n=1,\ j=1}^{N_{\mathbf{\alpha}},\ N_{\text{new}}}$, $\mathbf{f}^{\text{new}}(\cdot)=\{\mathbf{f}_{\mathbf{\alpha}_n,o_j}^{\text{new}}(\cdot)\}_{n=1,\ j=1}^{N_{\mathbf{\alpha}},\ N_{\text{new}}}$\;

    \ \;
    
	\ \ \ \ \ \ \ \ \ \ \ \ \ \ \ \ \textbf{\textit{Iterative feature observation collection and prior knowledge update}}\;

    \ \;
    
	\While{ not $\ stopCondition()$} 
	{
		\textbf{\textit{Active feature observation collection}}\;
		\\
		$\text{UNC}(\mathbf{\alpha}_n,o_j^{\text{new}}) \gets uncertaintyEstimate(\mathbf{f}^{\text{new}}(\cdot),X^{\text{new}})$  \Comment{Eq. \ref{eq:H_estimate}}\;
		
		$ o^{\text{new}^{\ast}}, \mathbf{\alpha}^{\ast} \gets objectActionSelection(\text{UNC}(\mathbf{\alpha}_n,o_j^{\text{new}}))$
		\Comment{Eq.~ \ref{eq:al_action_selection}}\;
		
		$ \mathbf{x}^{\text{new}^{\ast}} \gets actionExecution(o^{\text{new}^{\ast}}, \mathbf{\alpha}^{\ast})$ \Comment{percieve new feature observation}\;
		
		$X^{\text{new}} \gets X^{\text{new}} \bigcup \mathbf{x}^{\text{new}^{\ast}}$ \Comment{update dataset}\;
		
		\textbf{\textit{Prior tactile experience update}}\;
		
		\For{$j=1:N_{\text{new}}$}
		{	$o_{\mathbf{\alpha}^{\ast}}^{\text{old}^{\ast}} \gets priorsSelection(X_{\mathbf{\alpha}^{\ast}}^{\text{old}},X_{\mathbf{\alpha}^{\ast},o_j}^{\text{new}})$ \;
		
        $\rho_{\mathbf{\alpha}^{\ast}}^{\ast} \gets relatednessEstimate(o_{\mathbf{\alpha}^{\ast}}^{\text{old}^{\ast}},X_{\mathbf{\alpha}^{\ast}}^{\text{old}},X_{\mathbf{\alpha}^{\ast},o_j}^{\text{new}})$\;
        
        $\mathbf{\gamma}_{\mathbf{\alpha}^{\ast}} \gets weightsEstimate( K^{(1)},K^{(2)}...,K^{(M_{\mathbf{\alpha}})})$
		
		$\mathbf{f}_{\mathbf{\alpha}^{\ast},o_j}^{\text{new}}(\cdot) \gets updateGPC(X^{\text{new}}_{\mathbf{\alpha}^{\ast},o_j},\rho_{\mathbf{\alpha}^{\ast}}^{\ast},\mathbf{\gamma}_{\mathbf{\alpha}})$\;	   
		}    
	}
\end{algorithm}
\section{What to Transfer} \label{sec:method:sensor_fusion}
When a robotic system applies an exploratory action on objects, it perceives multiple feature observations (e.g. by the pressing movement, it can perceive the object stiffness and thermal conductivity). The prior tactile instance knowledge are built using the prior objects feature observations from multiple sensory modalities that are combined together.  

The informativeness of each sensory output is dissimilar to each other: some are discriminant among objects and thus more informative, while some can be easily confused, i.e. they are less informative. Therefore, the robot should be able to intelligently select which sensor modality is more reliable. We formulate the task of multiple feature observations combination as a multi-kernel learning problem under the GPC framework, where each sensor modality has a kernel function to describe the similarity between its feature observations. By linearly combining the kernels, the robotic system can learn about objects with higher accuracy than using single sensor modality.

In the following, we first describe our feature observations combination method in detail. To support the understanding of our method, we show a visualization of its behavior using a toy dataset and the pseudo-code by Alg.~\ref{alg:sensor_fusion}.
 
\subsection{Combining Multiple Feature Observations} \label{subsec:method:method:sensor_fusion}
Following the notation from Sec.~\ref{sec::method::TL-problem}, we define $\mathbf{\alpha}$ as an exploratory action. A feature observation perceived by a robot after applying $\mathbf{\alpha}$ consists of multiple observations from different sensors and can be described as: 
\begin{equation}
    \mathbf{x}_{\mathbf{\alpha}} = [\mathbf{x}^{(1)}_{\mathbf{\alpha}},\mathbf{x}^{(2)}_{\mathbf{\alpha}},...,\mathbf{x}^{(m_{\mathbf{\alpha}})}_{\mathbf{\alpha}},...,\mathbf{x}^{(M_{\mathbf{\alpha}})}_{\mathbf{\alpha}}]
\end{equation}
where $\mathbf{x}^{(m_{\mathbf{\alpha}})}_{\mathbf{\alpha}}$ is a feature observation from the sensor modality $m_{\mathbf{\alpha}}$. For the pressing and static contact movements, we use the normal force and temperature sensing, for the sliding movement the accelerometer and temperature sensing (Tab.~\ref{tab:action+signal}).  

Now we assume that for a sensor modality $m_{\mathbf{\alpha}}$, a kernel function $K^{(m_{\mathbf{\alpha}})}$ is given. To combine multiple feature observations and to exploit the information from all sensors after applying the exploratory action $\mathbf{\alpha}$, we linearly combine the kernels:
\begin{equation}
K_{\mathbf{\alpha}}' = \gamma_{\mathbf{\alpha}}^{(1)} K^{(1)} + \gamma_{\mathbf{\alpha}}^{(2)} K^{(2)} + ... + \gamma_{\mathbf{\alpha}}^{(m_{\mathbf{\alpha}})} K^{(m_{\mathbf{\alpha}})} + \gamma_{\mathbf{\alpha}}^{(M_{\mathbf{\alpha}})} K^{(M_{\mathbf{\alpha}})}
\end{equation}
where $\gamma_{\mathbf{\alpha}}^{(m_{\mathbf{\alpha}})} \geq 0$. This hyper-parameter controls how much a robot can rely on the sensor modality $m_{\mathbf{\alpha}}$. It ranges between $0$ and $1$, with $\gamma_{\mathbf{\alpha}}^{(m_{\mathbf{\alpha}})} = 0$ indicating that the sensor feedback is not informative, and $\gamma_{\mathbf{\alpha}}^{(m_{\mathbf{\alpha}})} = 1$ highly useful. We further constrain these hyper-parameters with $L_1$ norm:
\begin{equation}
\sum_{m_{\mathbf{\alpha}=1}}^{M_{\mathbf{\alpha}}} |\gamma_{\mathbf{\alpha}}^{(m_{\mathbf{\alpha}})}| = 1.
\end{equation}

As discussed in Sec.~\ref{subsec:background:kernel_contruction}, the linearly weighted kernel satisfies the Mercer Condition, and thus is valid for training the GPC model.

\begin{figure}[!htp]
	\centering
	\begin{minipage}{1\linewidth}
		\centering
	\includegraphics[width=0.83\textwidth]{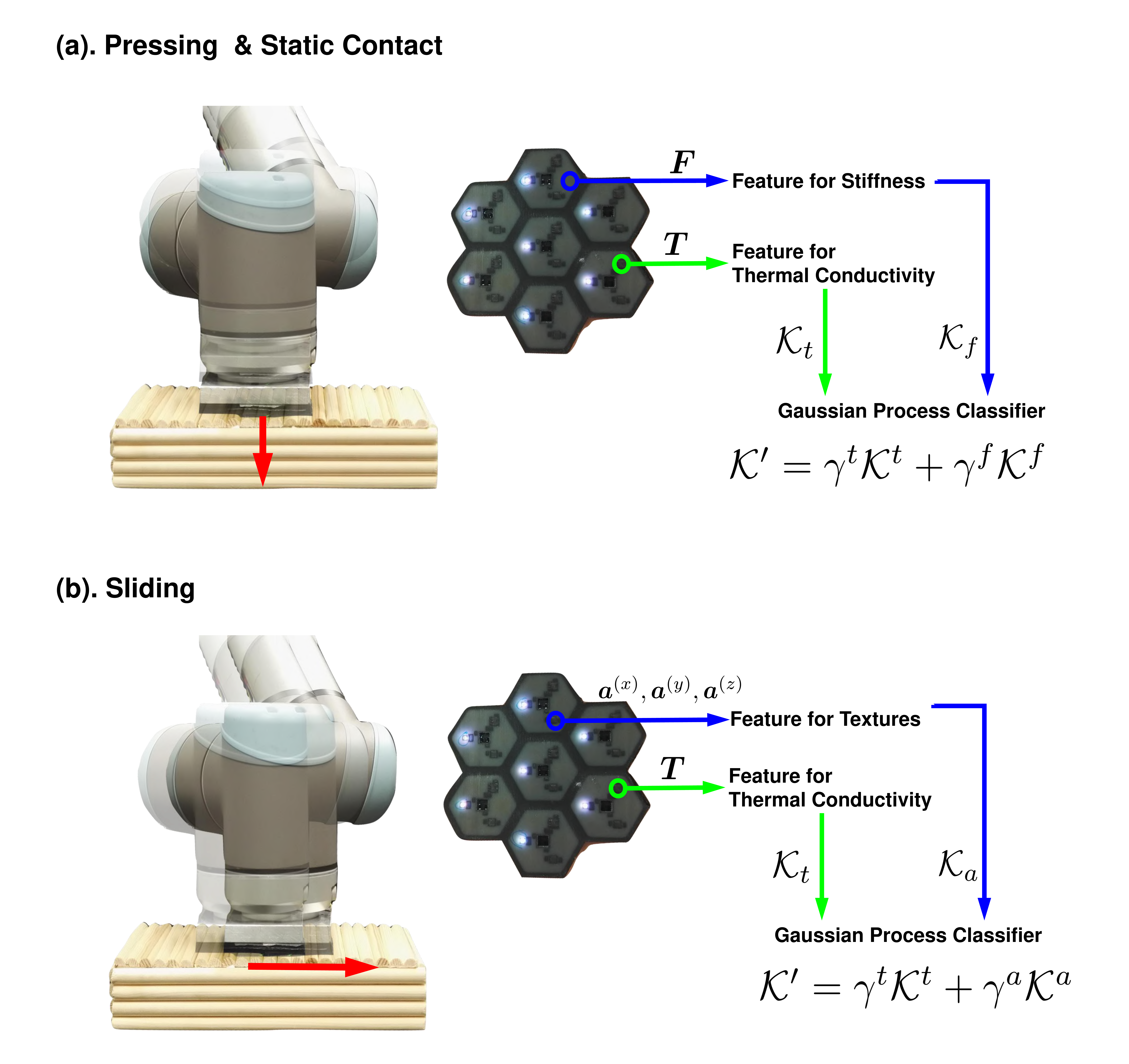}
	\end{minipage}
	\caption[Illustration of multiple feature observations combination method.]{Illustration of multiple feature observations combination method. The figures are reproduced and modified based on \cite{rodner2010one}. \textbf{(a)} The robotic system combined normal force sensing and temperature sensing to learn about objects by applying pressing and static contact movements. \textbf{(b)} The robot slides on the object surface to sense its textural property and thermal conductivity.}\label{fig:sensor-fusion}
\end{figure}

For each exploratory action, the GPC observation model is built using $K'$. $\gamma$ and the parameters in kernels are determined by maximizing the log marginal likelihood (Sec.~\ref{subsec:background:gpc_optimization}). In this way, the robotic system can actively exploit the information for multiple sensors to build the prior tacitle instance knowledge. Fig.~\ref{fig:sensor-fusion} illustrates our multiple feature observations combination method.
\subsection{An Example}
\begin{figure}[!htp]
	\centering
	\begin{minipage}{1\linewidth}
		\centering
		\includegraphics[width=0.85\textwidth]{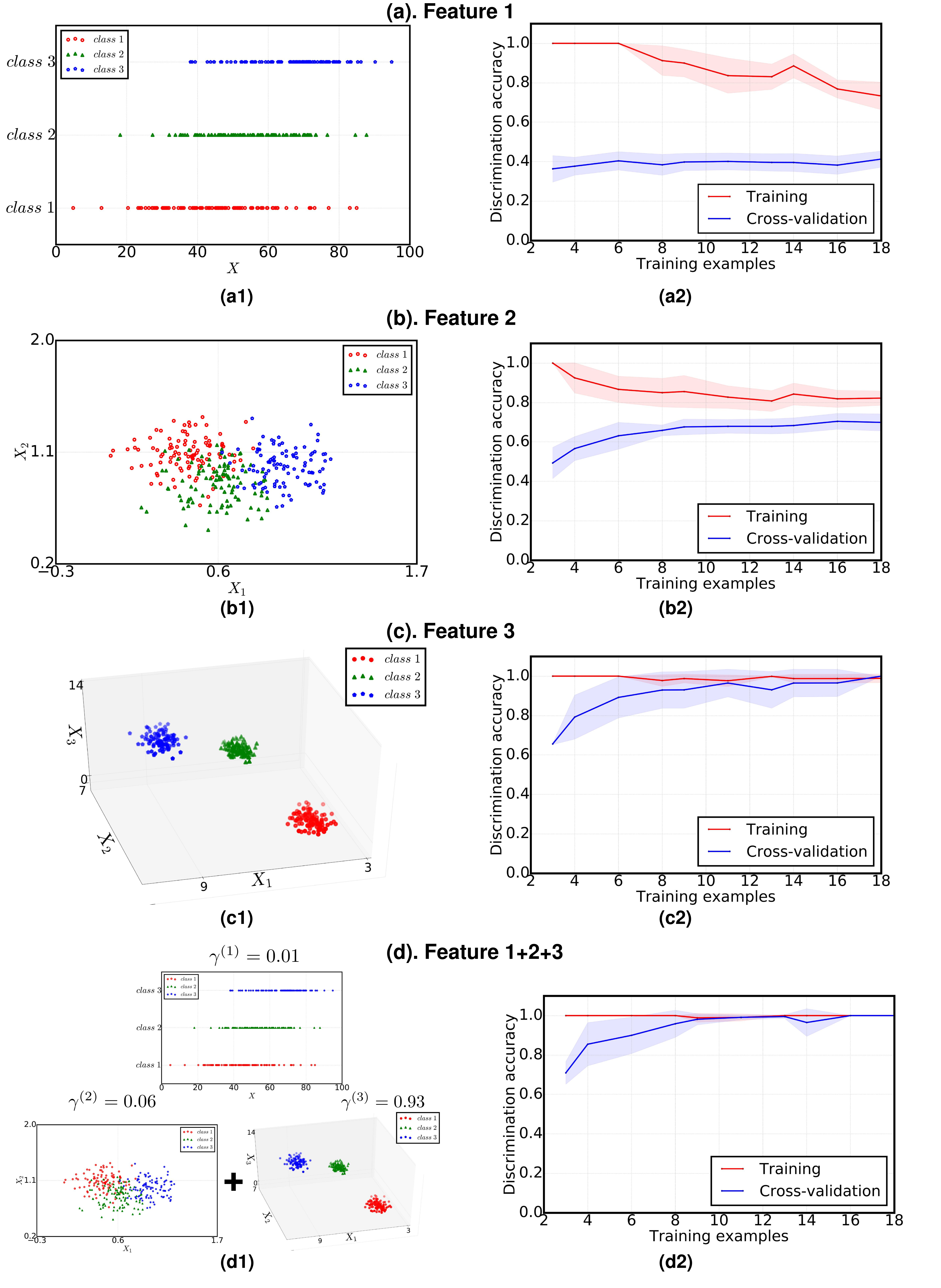}
	\end{minipage}
	\caption[Illustration of how our multiple sensor combination algorithm works.]{A toy example to illustrate how our multiple sensor combination algorithm works. \textbf{(a1), (b1), (c1)}: the feature(s) distribution of three different data. \textbf{(a2), (b2), (c2)}: the learning curve for building the GPC model when using only one feature. \textbf{(d1)}: using three features together to build GPC by weighted kernel combination. The RBF kernels were employed. The averaged weights for each feature are shown by $\gamma$. \textbf{(d2)}: the learning curve using all features.} \label{fig:sensor-fusion-example}
\end{figure}

We explain how our algorithm works with a toy dataset shown by Fig.~\ref{fig:sensor-fusion-example}. 

Imagine a robot perceives three types of feature observations to discriminate among three objects, which we refer to as $class\ 1$, $class\ 2$, $class\ 3$. The features ranges from one dimension to three dimensions. Their observation distributions are shown by Fig.~\ref{fig:sensor-fusion-example} (a1), (b1), (c1). Based on Feature 1, the objects are highly confused from each other, whereas based on Feature 3, objects can be easily discriminated. Fig.~\ref{fig:sensor-fusion-example} (a2), (b2), (c2) demonstrate the learning curves each feature respectively. The training accuracy and test accuracy by $10$-fold cross-validation are plotted. These plots show that learning Feature 1 yields the lower training and test accuracy than via Feature 2 and 3. Using our proposed multiple sensor combination algorithm, the robot casts more weights on Feature 3, and less in Feature 2 and 1 (shown by the $\gamma$ values in Fig.~\ref{fig:sensor-fusion-example} (d1)). By exploiting the information from three different features together, the robot achieves the highest learning performance (Fig.~\ref{fig:sensor-fusion-example} (d2)).

\begin{algorithm}[!htp]
	\caption{multiple feature observations combination}
	\label{alg:sensor_fusion}
	\SetKwInOut{Input}{Input}
	\SetKwInOut{Output}{Output}	
	
	\Input{$O=\{o_j\}_{j=1}^{N_o}$ \Comment{$N_o$ number of  objects, each object is regarded as a class.}\;
	
	$A=\{\mathbf{\alpha}_n\}_{n=1}^{N_{\mathbf{\alpha}}}$ \Comment{$N_{\mathbf{\alpha}}$ number of exploratory actions with different action parameters}\;
	
	$X = \{X_{\mathbf{\alpha}_n, o_j}\}_{n=1,\ j=1}^{N_{\mathbf{\alpha}},\ N_o}$\Comment{feature observations}\;
	}\;
	
	\Output{$\mathbf{\gamma} = \{ \gamma_{\mathbf{\alpha}_n,o_j}^{(m_{\mathbf{\alpha}})} \}_{\mathbf{\alpha}_n=1, m_{\mathbf{\alpha}}=1,j=1}^{N_{\mathbf{\alpha}},M_{\mathbf{\alpha}},N_o}$
	\Comment{Estimated sensory feedback weights.}\;
	}
	
	\For{$j=1:N_o$}
	{
	    \For{$n=1:N_{\mathbf{\alpha}}$}
	    {
	        $K_{\mathbf{\alpha}_n}^{'} \gets \gamma_{\mathbf{\alpha}_n}^{(1)} K^{(1)} + ... + \gamma_{\mathbf{\alpha}_n}^{(M_{\mathbf{\alpha}_n})} K^{(M_{\mathbf{\alpha}_n})}$ \Comment{Linear kernel combination Sec.~\ref{subsec:method:method:sensor_fusion}.}
	        
	        $\{ \gamma_{\mathbf{\alpha}_n,o_j}^{(m_{\mathbf{\alpha}})} \}_{m_{\mathbf{\alpha}}=1}^{M_{\mathbf{\alpha}}}
	        \gets optimizeGPC(K_{\mathbf{\alpha}_n}^{'},X_{\mathbf{\alpha}_n, o_j})$ \Comment{Finding optimal weights.}
	    }
	}
\end{algorithm}

\section{How to Transfer} \label{sec::method::how_to_transfer}
We now describe how a robotic system transfers the prior tactile instance knowledge from an old object $o_i^{\text{old}}$ in order to learn the GPC observation model of the new object $o_j^{\text{new}}$, based on the exploratory action $\mathbf{\alpha}$ \footnote{How to determine which prior object to be selected will be explained in the following section. For simplicity, in Sec.~\ref{sec::method::how_to_transfer} and Sec.~\ref{sec::method::where_to_transfer} we only describe tactile knowledge transfer based on one exploratory action $\mathbf{\alpha}$, and refer to $i$ and $j$ as $o_i^{\text{old}}$ and $o_j^{\text{new}}$, respectively.}. 
Following similar notations in the section related to the introduction of Gaussian Process (Sec.~\ref{sec::background::GP_model}), we define $\mathbf{h}^{\text{old}}_i$ as the GP latent function values for the old object $o_i^{\text{old}}$ and $\mathbf{h}^{\text{new}}_j$ for the new object $o_j^{\text{new}}$. We assume that these two function values are not independent from each other, but are sampled together over a dependent Gaussian Prior. This dependent GP is then used to construct the GPC observation model of the new object. The latent function can be modified accordingly: 
\begin{equation}
\mathbf{h}^{\text{new'}}_{j} \gets [\mathbf{h}^{\text{old}}_{i}, \  \mathbf{h}^{\text{new}}_{j}]. 
\end{equation}

We further incorporate the relatedness between prior object and new object into the dependent GP model by introducing the following dependent kernel function:
\begin{align} \label{eq:dependent_GP_kernel}
	K' = 
	\begin{bmatrix}
		K(X^{\text{old}}_i,X^{\text{old}}_i) \ \ \ \ \ &                   \rho K(X^{\text{old}}_i,X^{\text{new}}_j) \\
		\rho K(X^{\text{new}}_j,X^{\text{old}}_i) \ \ \ \ \ &
		K(X^{\text{new}}_j,X^{\text{new}}_j) \\
	\end{bmatrix}
\end{align}

$K(\cdot,\cdot)$ serves as the basic kernel function that measures the similarity between the feature observations from the same objects. $\rho K(\cdot,\cdot)$ measures the similarity between a feature observation from the old object and the one from the new object. $\rho$ controls the relatedness of  $o^{\text{old}}_{i}$ and $o^{\text{new}}_{i}$. Chai \textit{et al.} \cite{chai2009generalization} systematically evaluated the influence of $\rho$ with the range $[-1,1]$, with $\rho < 0$ indicating that the feature observations from two data sources are negative related, and $\rho > 0$ positive related. Furthermore, Cao \textit{et al.} \cite{cao2010adaptive} proved that the dependent kernel function $K'$ is semi-positive definite, given $|\rho|\leq 1$, and thus satisfies the Mercer Condition. In our work, we are not interested in transferring negative-related knowledge. Therefore, we constrain the range of $\rho$ within $[0,1]$.  

In the following, we discuss in detail the influence of $\rho$ values on the GPC performance, with a visualization on a toy dataset.\\

\noindent \textbf{Influence of different $\rho$ values}

\noindent When $\rho=0$, it indicates that there is no relatedness between the old object and new object. No feature observations transfer occurs. However, since the hyper-parameters of $K$ are tuned by the dependent kernel function, these values are still influenced by the old object. 

$0<\rho<1$ indicates that there is a relationship between old object and new object. The larger $\rho$ is, the more information from the old object can be transferred.

$\rho=1$ refers that two objects are regarded to be the same. In this case, the GPC observation model of the new object is learned by combining its feature observations with the old object under a basic kernel function $K$.

\begin{figure}[!htp]
	\centering
	\begin{minipage}{1\linewidth}
		\centering
		\includegraphics[width=1\textwidth]{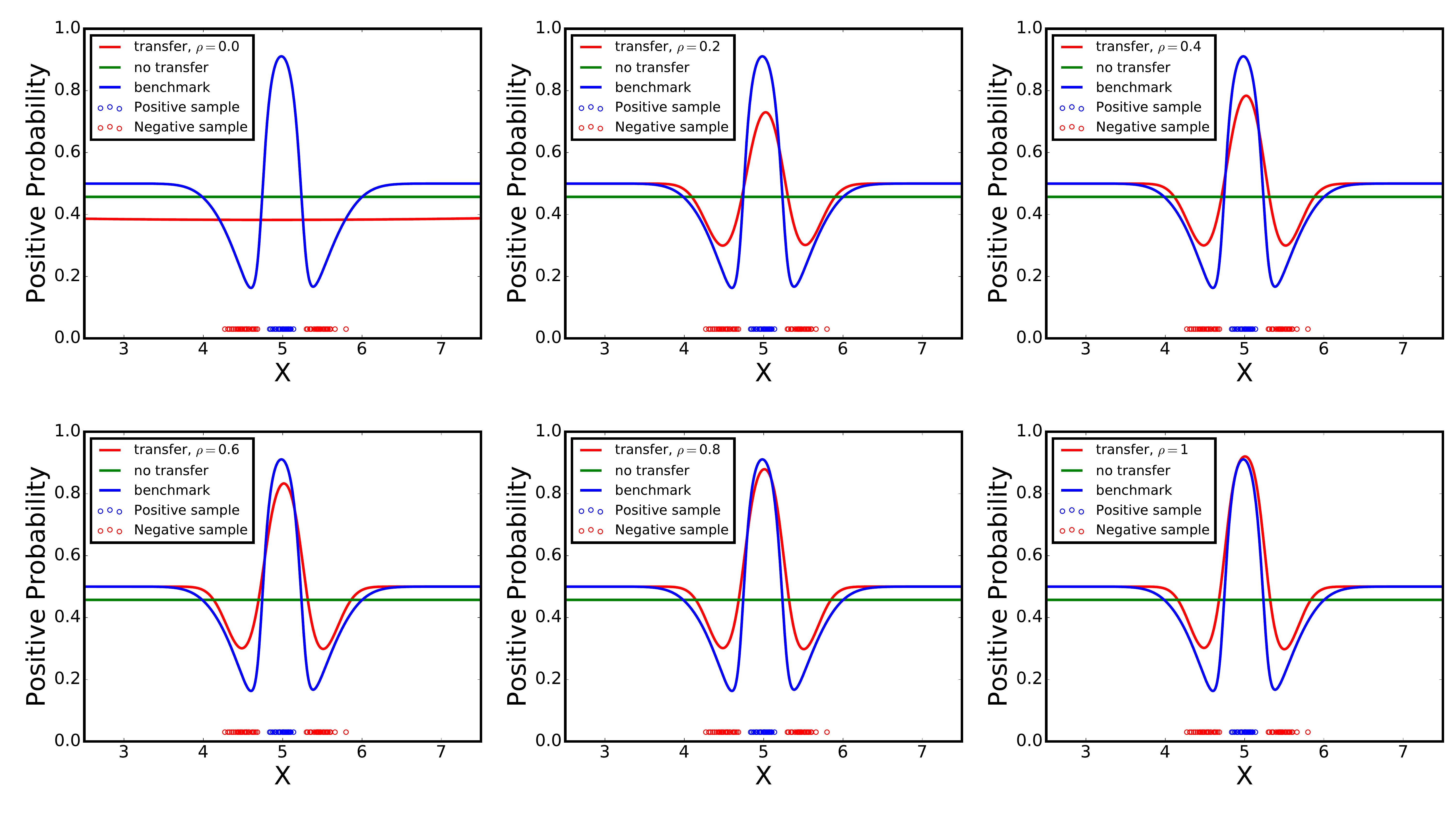}
	\end{minipage}
	\caption[A toy example to explain the influence of $\rho$.]{A toy example to explain the influence of parameter $\rho$.} \label{fig:rho_explaination}
\end{figure}

We give an example to visualize the influence of $\rho$ on a binary GPC classification problem (Fig.~\ref{fig:rho_explaination}). Suppose we have one dimensional feature distributions for three objects, as shown by the scatters in Fig.~\ref{fig:rho_explaination} (the feature value is represented in $x$ axis). We want to distinguish the object whose feature values lie at $5$ (blue scatters) from the objects at $4.5$ and $5.5$ (red scatters). Therefore, we consider this scenario as a binary classification problem, where the observations from the target object is assigned to be $+1$, and the rest $-1$. When using all feature observations to train a complete GPC model, the object posterior distribution (i.e. $p(y=+1|x)$) can be illustrated by the blue curve. Green curve shows the posterior distribution by the GPC model trained with merely $12$ feature observations ($4$ for positive labels, and $8$ for negative). As can be seen in Fig.~\ref{fig:rho_explaination}, the posterior distribution is flat, indicating that the training samples are insufficient. Now we suppose that there is an old object whose feature distribution is the same as the new object. We use its feature observations together with the $12$ samples mentioned above to train a dependent GPC with different $\rho$ values. The object posteriors are demonstrated by the red curves (Fig.~\ref{fig:rho_explaination}). By gradually increasing the object relatedness $\rho$ from $0$ till $1$, the object posterior distribution tends to be more similar to the benchmark (blue curve). This example not only indicates that introducing auxiliary feature observations helps improve the GPC model, but also addresses the importance of finding a good old object and a correct estimation of $\rho$. 

\section{From Where and How Much to Transfer} \label{sec::method::where_to_transfer}
Sec.~\ref{sec::method::how_to_transfer} describes how to transfer the prior tactile instance knowledge to learn about new objects. This section illustrates how a robotic system selects the \textit{most related} old object (from where to transfer) and how to determine the relatedness ($\rho$) between two objects (how much to transfer). Here, we propose two approaches.\\

\subsection{Model Prediction-Based Approach}
This method determines the most related old object and its relatedness to the new object $\rho^{\ast}$ by taking advantage of the prediction from the old objects' observation models (i.e. prior tactile model knowledge). Let $p(o^{\text{old}}_i|\mathbf{x}_j^{\text{new}})
$ be the prediction probability that a feature observation from the new object $\mathbf{x}_j^{\text{new}}$ is assigned to the old object $o^{\text{old}}_i$. We measure the expected prediction to all the observations $\mathbf{x}_j^{\text{new}} \in X_j^{\text{new}} $ that belong to the new object: $\bar{p}(o^{\text{old}}_i|X_j^{\text{new}})=\frac{1}{N_j^{\text{new}}}\sum p(o^{\text{old}}_i|\mathbf{x}_j^{\text{new}})$

\noindent where $N_j^{\text{new}}$ being the number of new object feature observations. the average prediction value estimates the similarity between the old object $o^{\text{old}}_i$ and the new object $o^{\text{new}}_i$. A larger value indicates that these two objects are highly similar. Therefore, we can use it to select the most related old object (denoted as $o^{ \text{old}^{\ast}}$) for a new object regarding on the exploratory action $\mathbf{\alpha}$. Furthermore, to avoid transferring irrelevant tactile information, we add a threshold $\epsilon_{neg_1}$ which prevents the robot from selecting any old object, if the prediction value is smaller than $\epsilon_{neg_1}$ \footnote{Since we use the binary GPC, the predictions from the observation model of a \textit{relevant} old object should be at least larger than $0.5$. This indicates that the condition $\epsilon_{neg_1}>0.5$ should be satisfied.}. Therefore, we have: $o^{\text{old}^{\ast}}= \underset{o_i^{ \text{old}} \in O^{\text{old}}}{\operatorname{arg\ max}}\ \bar{p}(o^{\text{old}}_i|X_j^{\text{new}}) $, if $\bar{p}(o^{ \text{old}^{\ast}}|X^{\text{new}}) \geq \epsilon_{neg_1}$. Once we select $o^{\text{old}^{\ast}}$, we further estimate $\rho^{\ast}s$ by $\rho^{\ast}$: $\rho^{\ast} = \bar{p}(o^{ \text{old}^{\ast}}|X^{\text{new}})$.\\

\subsection{Model Optimization Approach}
We can also consider $\rho$ to be a model parameter in the dependent GPC model, and use the model optimization method (introduced in Sec.~\ref{subsec:background:gpc_optimization}) to tune its value. In this regard, for each old object $o_i^{\text{old}}$, we train its dependent GPC model by maximizing the log-likelihood (Sec.~\ref{subsec:background:gpc_optimization}), and obtain the hyper-parameter $\rho_i$. The old object with the largest $\rho$ value is selected, and $\rho^{\ast}$ is determined accordingly: $o^{\text{old}^{\ast}} = \underset{o_i^{\text{old}} \in O^{\text{old}}}{\operatorname{arg\ max}}\ (\rho_i)$, 
$\rho^{\ast} = \underset{o_i^{\text{old}} \in O^{\text{old}}}{\operatorname{max}}\ (\rho_i)$. To avoid negative transfer, we also set a threshold $\epsilon_{neg_2}$ for preventing negative transfer (i.e. if $\rho^{\ast}<\epsilon_{neg_2}$, the robot does not select any old objects). 

\section{Prior Tactile Knowledge Update}
\label{sec::method::active_learning}
When the robot iteratively builds the new object dataset $X^{\text{new}}$ and reliable observation model $X^{\text{new}} \xrightarrow{\mathbf{f}^{\text{new}}}O^{\text{new}}$ for \textit{each} exploratory action, it actively decides which new object to explore and which exploratory action to apply, and correspondingly updates the prior tactile instance knowledge.

\subsection{Active Feature Observation Collection.} 
We follow the method proposed from \cite{di2017iros} for next object and next exploratory action selection strategy: at each iteration, the robot first updates the GPC model for each exploratory action with the feature observations collected hitherto, and estimates the uncertainty in the observation models. The robot hereof measures the Shannon entropy of the object posterior for each feature observation $\mathbf{x}^{\text{new}} \in X^{\text{new}}$:

\begin{align} \label{eq:shannon} H(o^{\text{new}}|\mathbf{x}^{\text{new}})=-\sum_{o_i^{\text{new}} \in O^{\text{new}}} p(o_i^{\text{new}}|\mathbf{x}^{\text{new}})\log(p(o_i^{\text{new}}|\mathbf{x}^{\text{new}})).
\end{align}

Then the new objects' feature observations are categorized according to the exploraotry action and the new object class, i.e. $X^{\text{new}} = \{X^{\text{new}}_{\mathbf{\alpha}_n, o_j}\}_{n=1,\ j=1}^{N_{\mathbf{\alpha}},\ N_{new}}$. Each category contains $N^{\text{new}}_{n,j}$ number of feature observations. The uncertainty in the GPC model $\text{UNC}(\mathbf{\alpha}_n,o_j)$ is estimated as the mean value of the Shannon entropy:
\begin{align} \label{eq:H_estimate}
	\text{UNC}(\mathbf{\alpha}_n,o_j^{\text{new}})=\dfrac{1}{N^{\text{new}}_{n,j}} \sum_{x^{\text{new}}_{n,j} \in X^{\text{new}}_{\mathbf{\alpha}_n, o_j^{\text{new}}}} H(o_j^{\text{new}}|\mathbf{x}_{n,j}^{\text{new}})
\end{align}
A large $\text{UNC}(\mathbf{\alpha}_n,o_j)$ indicates that the robot is uncertain about the object feature observations from the exploratory action $\mathbf{\alpha}_n$. 

As \cite{di2017iros} describes, the next object and the next action selection process should consider the exploration-exploitation trade-off. In this regard, the next exploratory action $\mathbf{\alpha}^{\ast}$ and the next object $o^{\text{new}^{\ast}}$ are determined by:

\begin{align} \label{eq:al_action_selection}
    \left\{
    \arraycolsep=1.5pt\def\arraystretch{0.6}
	\begin{array}{@{}ll@{}}
		o^{\text{new}^{\ast}}, \mathbf{\alpha}^{\ast}  = \underset{\mathbf{\alpha}_n \in A;\ o_j^{\text{new}} \in O^{\text{new}}}{\operatorname{arg\ max}} \text{UNC}(\mathbf{\alpha}_n,o_j)\ \ \  & \text{if}\ p_{\text{rand}} \geq \epsilon_{\text{explor}} \\
		\\
		o^{\text{new}^{\ast}} = \mathcal{U}\{o_1^{\text{new}},o_2^{\text{new}},...,o_{N_{\text{new}}}^{\text{new}}\}, \mathbf{\alpha}^{\ast} = \mathcal{U}\{\mathbf{\alpha}_1,\mathbf{\alpha}_2,...\mathbf{\alpha}_{N_{\mathbf{\alpha}}} \}
		&\text{otherwise} \\		
	\end{array}\right.
\end{align}
where $\epsilon_{\text{explor}}$ is a pre-defined value that determines how much the robot collects a new feature observation based on the estimate of the GPC uncertainty. $p_{\text{rand}}$ is a randomly generated value at each feature collection step, following the uniform distribution $\mathcal{U}(0,1)$. \\

\subsection{Knowledge Update.} 
Once the robot collects a new feature observation by applying the exploratory action $\mathbf{\alpha}^{\ast}$, it updates the prior tactile instance knowledge only for this exploratory action. This process includes updating the multiple feature observations combination, updating the object relatedness $\rho$, and transfer the prior feature observations to the new objects' observation models. 

\cleardoublepage
\let\textcircled=\pgftextcircled
\chapter{Experimental Validation and Results}\label{chapter:experiment}

\initial{T}his chapter shows the experimental evaluation of our proposed ATIKT method. First, we show what experimental objects were used (Sec~\ref{sec:experiment:object}) and how the robot applied exploratory actions with different action parameters on them (Sec.~\ref{sec:experiment:object}). Afterwards, we systematically evaluated our proposed transfer learning algorithm, including (1). evaluating our proposed multi-sensor feature observations combination technique (Sec.~\ref{sec:experiment:sensor_fusion}); (2). analyzing ATIKT with different combinations and number of prior objects (Sec.~\ref{sec:experiment:TL_evaluation}); (3). testing the algorithm's robustness against negative transfer (Sec.~\ref{sec:experiment:Negative_transfer}).

\section{Experimental Objects}\label{sec:experiment:object}
\begin{figure}[!htpb]
	\centering
	\begin{minipage}{1\linewidth}
		\centering
		\includegraphics[width=1\textwidth]{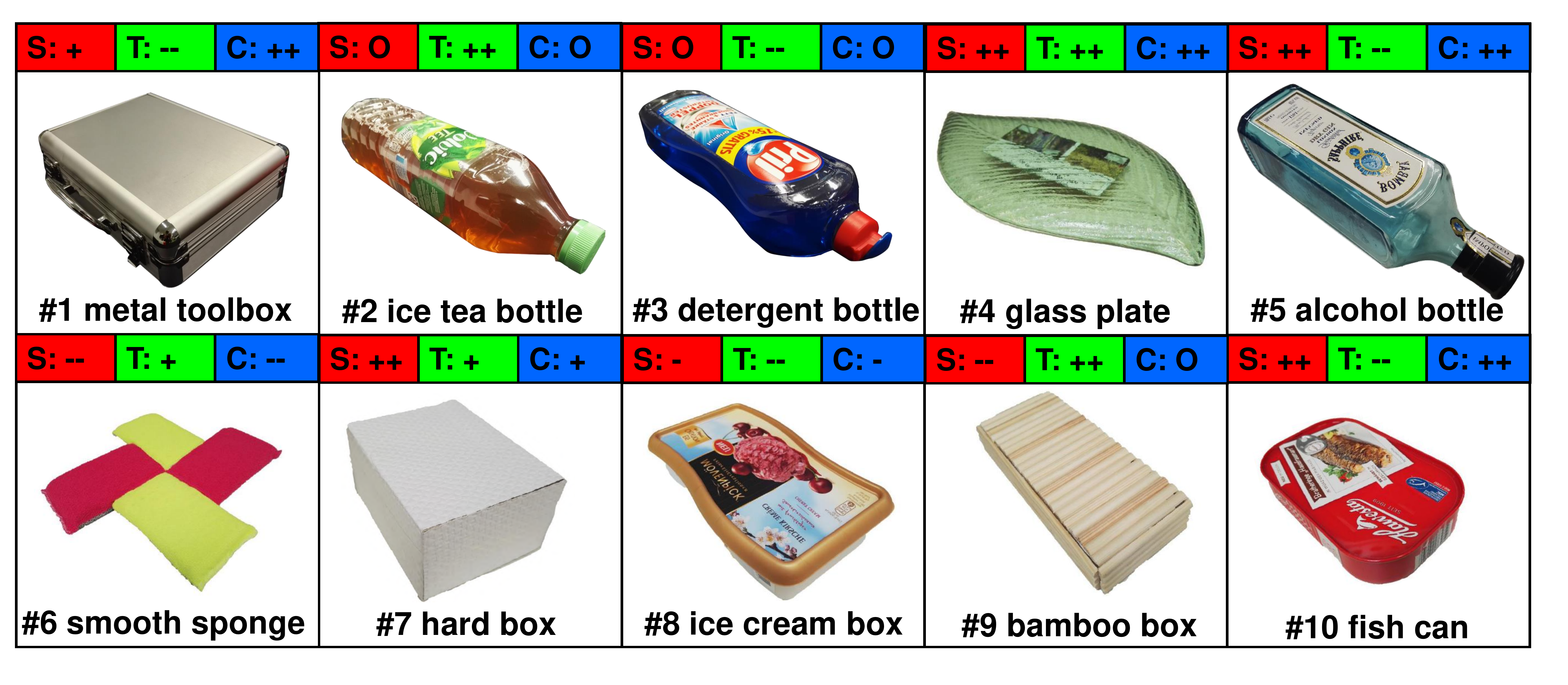}
	\end{minipage}
	\caption[Prior object list.]{Prior object list. The object physical properties were evaluated by the human subjects. Notice that by changing the exploratory action parameters, the perceived object physical properties may change. S: stiffness, T: roughness of surface textures, C: thermal conductivity. ++: very high; +: high; O: middle; -:low; --: very low.}\label{fig:object_list_old}
\end{figure}

\begin{figure}[!htpb]
	\centering
	\begin{minipage}{1\linewidth}
		\centering
		\includegraphics[width=1\textwidth]{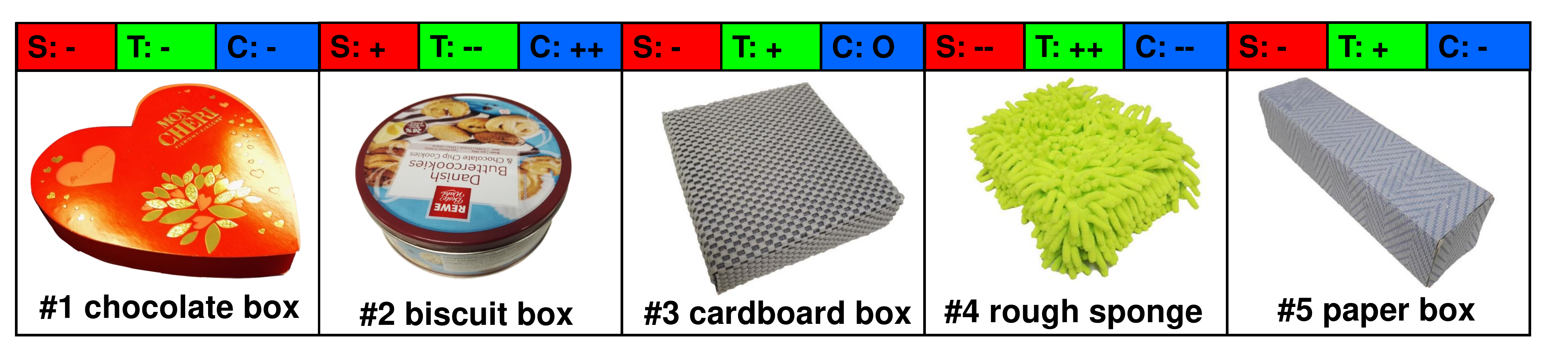}
	\end{minipage}
	\caption[New object list.]{New object list. The object physical properties were evaluated by the human subjects. The new object index starts from $1$. In the rest of the thesis, when we combine the new objects with the old objects, the new object index starts from $11$.}\label{fig:object_list_new}
\end{figure}

We deliberately selected $10$ daily objects with different shapes and different physical properties as prior objects which served to build the robotic prior tactile exploratory action experiences, i.e. prior tactile instance knowledge and model knowledge (Fig.~\ref{fig:object_list_old}). We further selected $5$ objects as new objects, shown by Fig.~\ref{fig:object_list_new}. For each new object, there existed one or more old objects that shared similar physical properties. For example, both rough sponge and smooth sponge are soft; paper box and hard box have similar surface textures; metal toolbox and biscuit box have high thermal conductivity. In this way, when learning about new objects based on their physical properties, the robot can leverage the related prior tactile instance knowledge.

Due to the constraints of the UR10 robotic system, all objects were chosen to have flat surfaces and have larger area than the artificial skin, so that they can be fully connected with the tactile sensors. Furthermore, as the transparent object can hardly be detected by the proximity sensors, resulting in an inaccurate light-touch detection, we deliberately added non-transparent tab on the backside of object contact areas (e.g. glass plate was tapped with a white paper).

\section{Action Parameters Determination and Test Data Collection}\label{sec:experiment:data_collection}
\subsection{Exploratory Actions Determination}
As shown in the section related to the exploratory action definition (Sec.~\ref{sec:action_perception:exploratory_action}), the robot can press, slide or build a static contact on objects to perceive their physical properties. The robot that applies the same actions but with different action parameters can attain different feature observations, e.g. pressing on an object surface with depth $d_P=1$ mm or $d_P=2$ mm. In our experiment, we defined $7$ exploratory actions for pressing, sliding, and static contact with various action parameters shown in Tab.~\ref{tab:experimental_actions}. These action parameters were deliberately selected that satisfied the experimental setup constraints. For instance, sliding on the surface of objects with contact force larger than $F_S=0.2$ N caused unexpected robot vibration; Pressing the objects with $d_P>2$ mm resulted in large normal force feedback for hard objects (wood, metal etc.), which made the UR10 robot emergency stop. Furthermore, we found that the pressing depth did not influence the temperature measurement by the static contact movement, when $d_C > 1$mm. This is due to the fact that the artificial skin could fully contact with object surfaces. Therefore, in the experiment we only considered one static contact exploratory action (C1).

\begin{table}[!htp]
    \centering
    \begin{tabular}{ c c } 
     \hhline{==}
     \textbf{Exploratory actions}  & \textbf{Action parameters} \\
     \textbf{(Notations)} & \\ 
     \hline
     Pressing (P1)     & $d_P=1$ mm, $t_P=3$ s. \\
     Pressing (P2)   & $d_P=2$ mm, $t_P=3$ s. \\
     Sliding (S1)   & $F_S=0.1$ N, $t_S=1$ s, $v_S=1$ cm/s. \\
     Sliding (S2)   & $F_S=0.1$ N, $t_S=1$ s, $v_S=5$ cm/s.\\
     Sliding (S3)   & $F_S=0.2$ N, $t_S=1$ s, $v_S=1$ cm/s.\\
     Sliding (S4)   & $F_S=0.2$ N, $t_S=1$ s, $v_S=5$ cm/s.\\
     Static Contact (C1) & $d_C=2$ mm, $t_C=15$ s. \\
     \hhline{==}
    \end{tabular}   
    \caption[Exploratory actions determination and action parameters used in the experiments.]{Exploratory actions determination and the action parameters used in the experiments.}
    \label{tab:experimental_actions}
\end{table}

Before applying any $7$ exploratory actions, the robot built a light contact with objects. A light contact was detected, once the total normal force on the the artificial skin increased above $0.05$ N. In order to efficiently build an accurate light contact, the robotic system should actively adjust its moving speed when it approaches to an object. When the distance between the artificial skin and the object is large, slowly approaching the robot to the object takes unnecessarily more time; on the other hand, when the distance is small, approaching the object with a large speed makes it difficult to control the light contact force. We hereof combined the proximity and normal force sensing to realize an efficient light contact detection. Based on the observations from all proximity sensors in the skin, the distance $d$ between the artificial skin and the object can be estimated. The robot adjusted its moving speed and moving distance according to $d$. The smaller $d$ is, the slower the robot should approach to the object. In this way, a light contact could be detected with an error less than $0.01$ N for hard objects (wood, metal etc.) and $0.005$ N for soft objects (paper box, sponge etc.) in our experiments. 

When perceiving the object thermal conductivity, the robot requires a similar initial temperature condition before it touches its artificial skin with objects. Therefore, after executing an action, the robot was controlled to raise its end-effector for $30$ s such that the temperature sensors could be restored to the ambient temperature. It is noteworthy to mention that all the tactile data were collected in the lab, which built a closed environment so that the fluctuation of the ambient temperature was controlled to be less than $1^{\circ}$C. This variance is acceptable for our feature extraction method of thermal conductivity. For a detailed evaluation of the influence of the initial temperature, please refer to the work from Bhattacharjee \textit{et al.} \cite{bhattacharjee2015material}.
\subsection{Data Collection}
\begin{figure}[!htp]
	\centering
	\begin{minipage}{1\linewidth}
		\centering
		\includegraphics[width=0.9\textwidth]{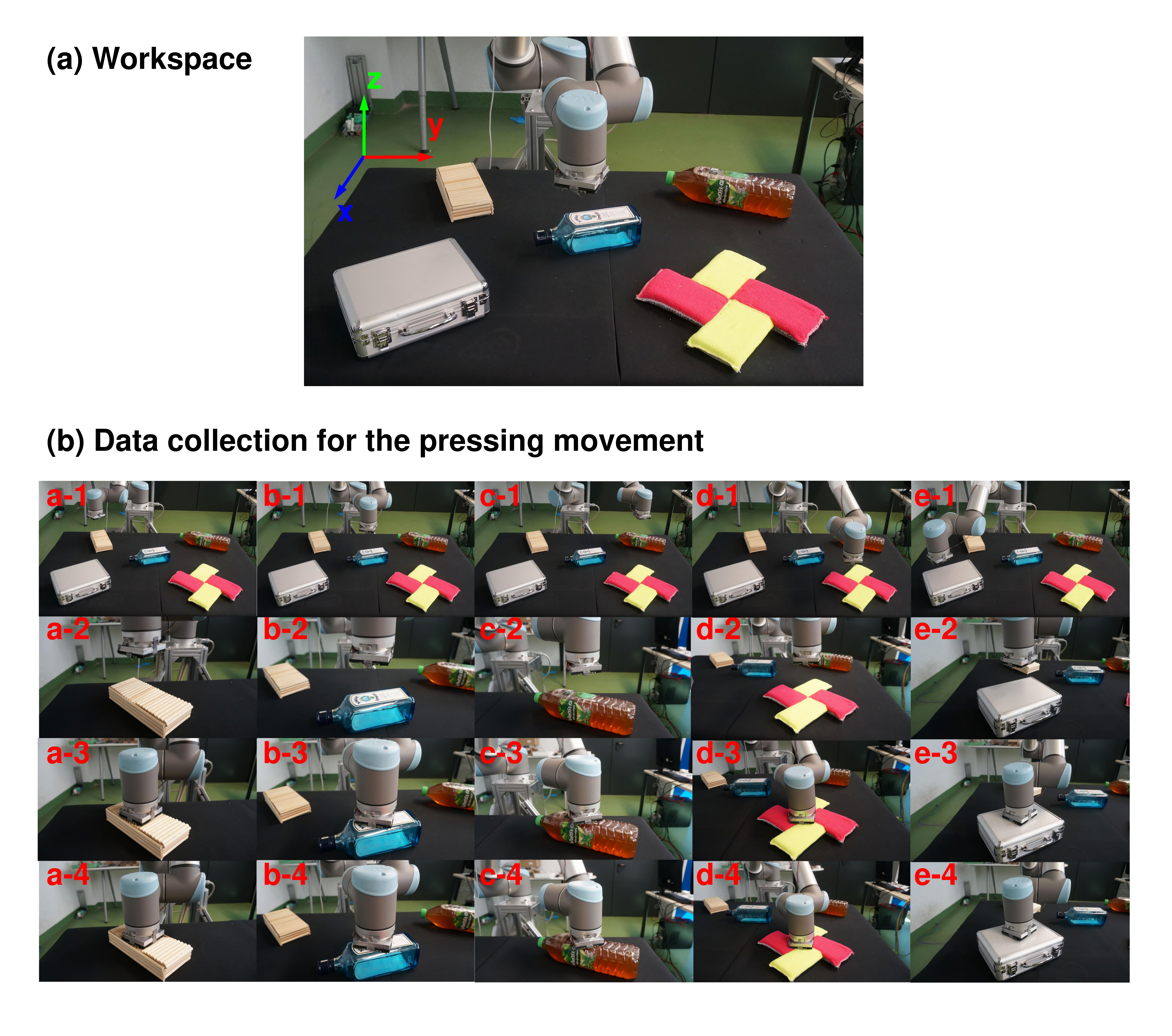}
	\end{minipage}
	\caption[An example of data collection for the pressing movement (P2).]{An example of data collection for the pressing movement (P2). \textbf{(a)} the workspace for data collection, where the objects were placed with different orientations. \textbf{(b)} the data collection process for the pressing movement P2. a-1(a-2) - e-1(e-2): the robot positioned its artificial skin above the objects. a-3 - e-3: the robot built a light contact with the objects. a-4 - e-4: the robot pressed the objects with $d_S=2$ mm and held for $3$ s.} \label{fig:data_collection_pressing}
\end{figure}

We evaluated the performance of our proposed method based on a test dataset. This dataset was built by the robot by iteratively collecting feature observations from the objects that were placed in a workspace (see Fig.~\ref{fig:data_collection_pressing}). At each round of data collection, objects were manually shifted and rotated, and the robot sequentially applied one of the seven exploratory actions once on each object. This process was repeated $10$ times for the static contact movement (C1), and $20$ times for the rest exploratory actions (P1, P2, S1, S2, S3, S4). In this way, our dataset was robust against the variations in the object contact locations and the physical change of our artificial skin. In the end, we had a test dataset with \textit{15 objects$\times$6 exploratory actions$\times$20 trials +  15 objects$\times$1 exploratory actions$\times$10 trials = 1950} samples. 

\begin{figure}[!htp]
	\centering
	\begin{minipage}{1\linewidth}
		\centering
		\includegraphics[width=0.9\textwidth]{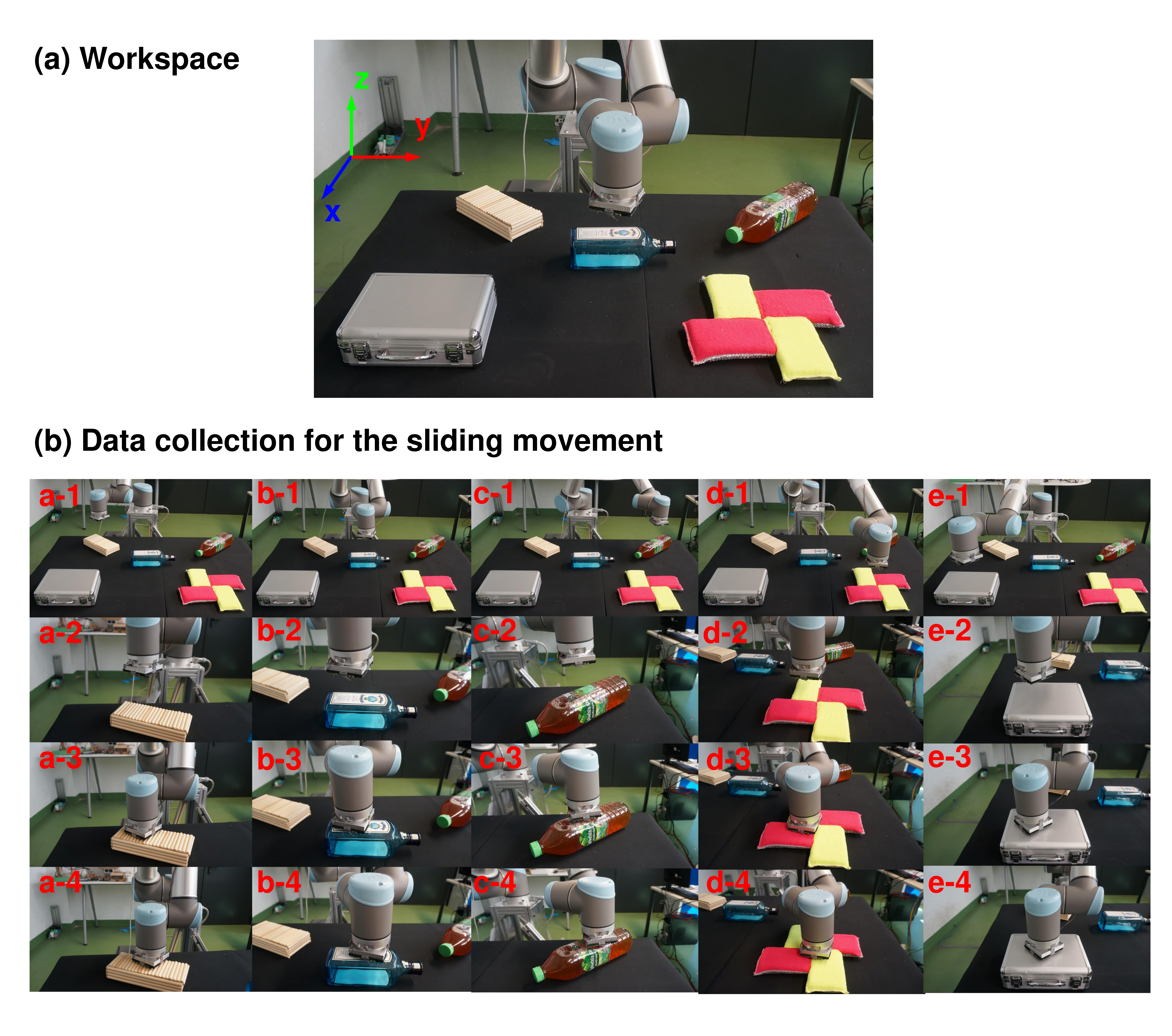}
	\end{minipage}
	\caption[An example of data collection for the sliding movement (S1).]{An example of data collection for the sliding movement (S1).}\label{fig:data_collection_sliding}
\end{figure}

Fig.~\ref{fig:data_collection_pressing} and Fig.~\ref{fig:data_collection_sliding} show examples of the data collection process. In Fig.~\ref{fig:data_collection_pressing}, the robotic system was collecting the feature observations by the pressing movement P2 for objects \#1, \#2, \#5, \#6 and \#9. The robot pressed each object once, as can be seen from the photos a-1 till e-1 in Fig.~\ref{fig:data_collection_pressing}. When the robot applied a pressing movement on an object, it first positioned its sensory part above the object (e.g. photos a-1 and a-2 in Fig.~\ref{fig:data_collection_pressing}), and then built a light contact using the sensory feedbacks from the proximity sensors and normal force sensors, as can be seen by photos a-3 till e-3 in Fig.~\ref{fig:data_collection_pressing}. In Fig.~\ref{fig:data_collection_sliding}, the object orientations were manually changed in order to ensure the data variance in our test dataset. The robot applied the sliding movement (S1) for each object once (sliding horizontally, see photos b-3 and b-4 in Fig.~\ref{fig:data_collection_sliding} as an example).  

\section{Experimental Setting}\label{sec:experiment:setting}
We designed three experiments to assess the ATIKT algorithm. In the first experiment (Sec.~\ref{sec:experiment:sensor_fusion}), we tested our multiple feature observations combination algorithm, when the robot applied each of the seven exploratory actions (P1, P2, C1, S1, S2, S3, S4) to build GPC models. In Sec.~\ref{sec:experiment:TL_evaluation}, we evaluated the transfer learning algorithm with different combinations and number of prior objects. The performance ATIKT was compared with learning without prior tactile knowledge. Finally, we evaluated the robustness of our method against negative transfer, i.e. when the old objects were irrelevant to the new objects (Sec.~\ref{sec:experiment:Negative_transfer}).

The GPC models and kernel functions were developed on the basis of the \textit{scikit-learn} package (version 0.18.1, \textit{http://scikit-learn.org/stable/}). To find the optimized hyper-parameters by maximizing the marginal log-likelihood function, we used Sequential Least SQuares Programming method (SLSQP) (see \textit{https://docs.scipy.org/doc/scipy-0.13.0/reference/generated/scipy.optimize.fmin\_slsqp.html}).

\section{Evaluation of Combining Multiple Feature Observations}\label{sec:experiment:sensor_fusion}

In this experiment, we evaluated the performance of our proposed robotic multiple feature observations combination algorithm. The robot selected $10$ groups of objects from the old object list (Fig.~\ref{fig:object_list_old}) and new object list (Fig.~\ref{fig:object_list_new}) \footnote{The indices of the new objects were re-arranged to start from $11$, following the indices of prior objects ($1$ - $10$).} to learn about their physical properties (i.e. building the GPC observation models) by applying all seven exploratory actions separately. Each group contained five objects that were randomly selected following the uniform distribution. 

We compared the learning performance of our proposed method with using only a single sensor modality as baseline methods. For actions P1, P2, C1, the combination of temperature and normal force sensing was compared with using temperature feedbacks and normal force feedbacks individually; for actions S1, S2, S3, S4 the combination of textural properties and temperature was compared with using accelerometers feedbacks and temperature sensors feedbacks. 

For each group of objects, the dataset was split randomly into training set and test set. The training set was used to train the GPC models which predicted the discrimination accuracy of the test dataset. By increasing the number of training samples, the learning curves could be obtained as an indication of the GPC performance. The experiments were conducted $10$ times for each object group. The averaged learning curves were plotted. For a fair comparison we used RBF kernel for each sensor modality.

\begin{figure}[!htp]
	\centering
	\begin{minipage}{1\linewidth}
		\centering
		\includegraphics[width=0.88\textwidth]{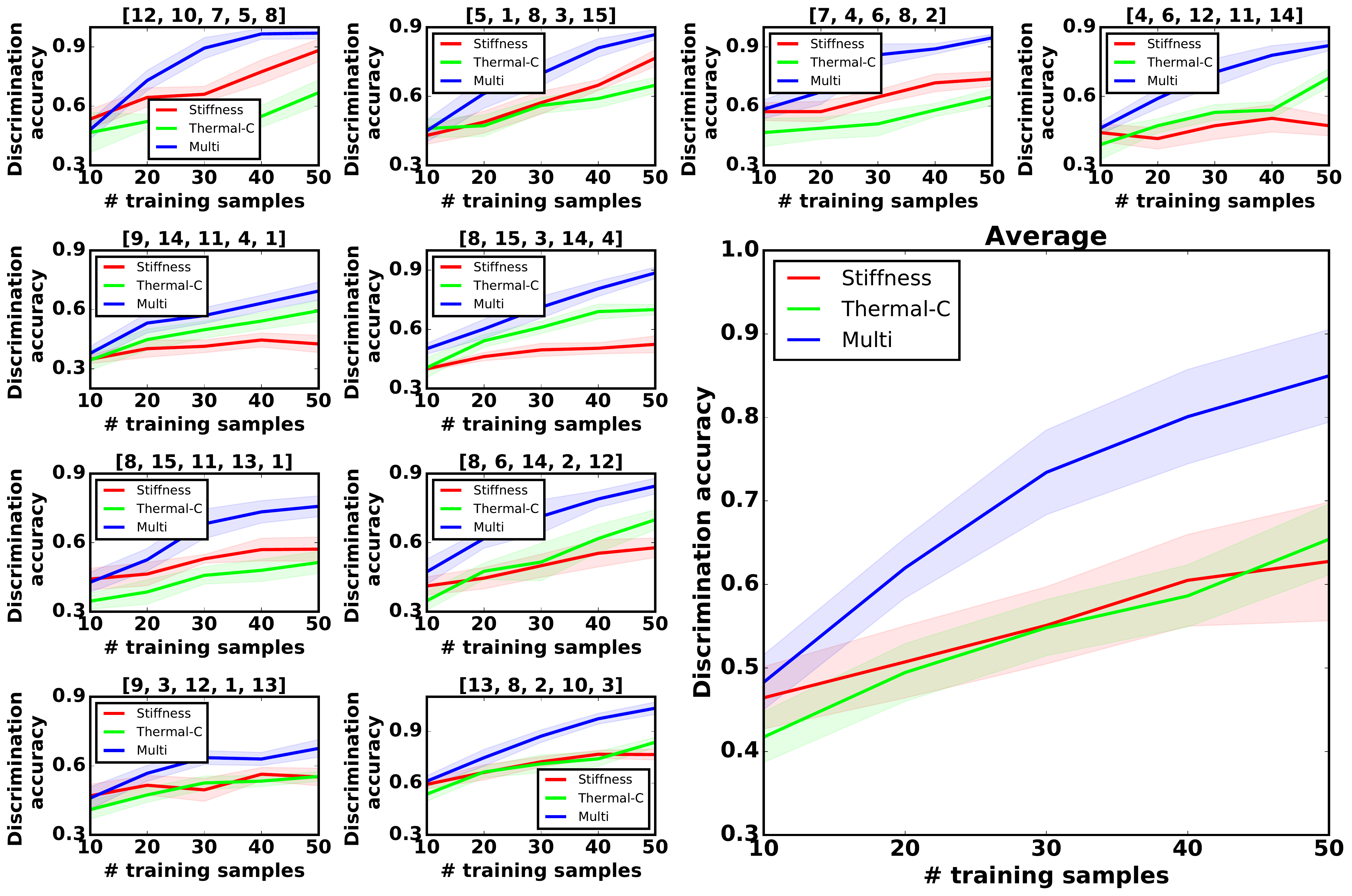}
	\end{minipage}
	\caption[The results for multiple feature observations combination of the movement P1.]{The results for multiple feature observations combination of the movement P1. Smaller plots show the learning curves from 10 groups of objects. Each group contained five objects, whose object indices were illustrated in the title of sub-figures. The bigger plot on the bottom right shows the averaged learning curves based on the stiffness observations (Stiffness), thermal conductivity observations (Thermal-C) or a combination of both (Multi).  }\label{fig:multiK_pressing_d=1}
\end{figure}

\begin{figure}[!htp]
	\centering
	\begin{minipage}{1\linewidth}
		\centering
		\includegraphics[width=0.88\textwidth]{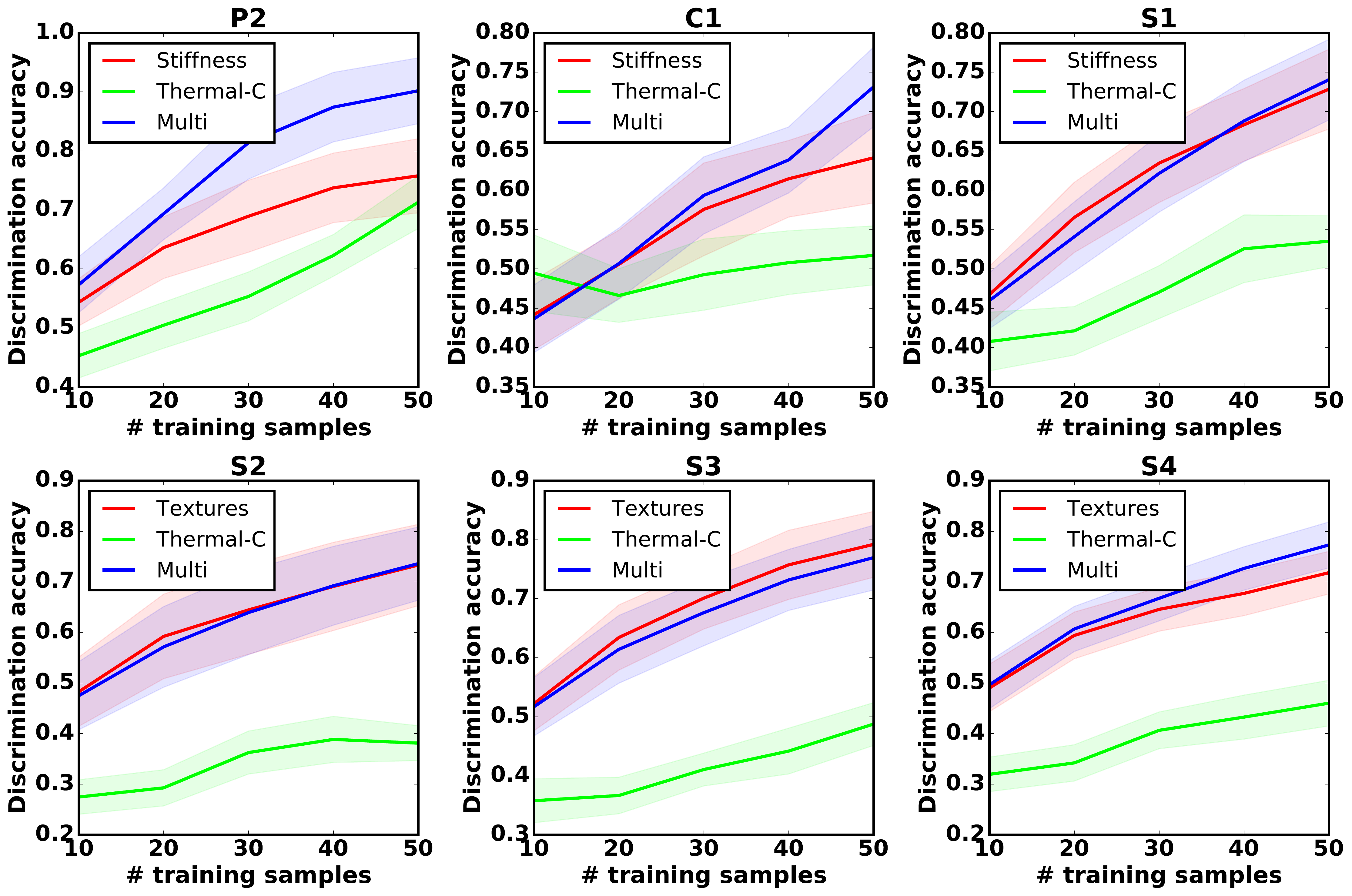}
	\end{minipage}
	\caption[Averaged multiple feature observations combination results.]{Averaged multiple feature observations combination results for actions P2, C1, S1 - S4.}\label{fig:multiK_average}
\end{figure}

Fig.~\ref{fig:multiK_pressing_d=1} shows the results of the exploratory action P1 for each object group and their averaged performance. Fig.~\ref{fig:multiK_average} shows the averaged results for P2, C1, S1, S2, S3, and S4. First, it can be seen that for different exploratory actions, the informativeness of the sensor modality was different. For instance, the object learning curves using textural properties were consistently better than using object thermal conductivity (Fig.~\ref{fig:multiK_average} S1-S4), indicating that by the sliding movement, vibro-tactile signals brought about more discriminate information than temperature signals. By the pressing and static contact movements, however, the temperature sensing could be more informative than the normal force sensing. Second, the informativeness of different sensory feedbacks also varied with regard to objects the robotic system learned about. For example, the thermal conductivity performed better than normal force sensing for the object group $\{8,15,11,13,1\}$, but worse for the object group $\{12,10,7,5,8\}$, when the robot applied the exploratory movement P1 (Fig.~\ref{fig:multiK_pressing_d=1}).

In all scenarios mentioned above (i.e. different exploratory actions or different object combinations), our proposed algorithm can either performed similar to the best single-sensor result, indicating that the robot actively selected the most informative sensory feedback, or took advantage from different sensor modalities to reach the best learning curves. The improvement of the discrimination accuracy using our proposed method can reach as much as $20\%$ (e.g. see averaged learning curves in Fig.~\ref{fig:multiK_pressing_d=1}).

\section{Evaluation of the Proposed Tactile Transfer Learning Method}\label{sec:experiment:TL_evaluation}
\subsection{Transferring the Prior Tactile Knowledge from Different Groups of Objects.} \label{subsec:experiment:TL_different_groups} 
In this experiment, we evaluated the performance of the ATIKT algorithm with different groups of prior objects. As an initialization, the robot first applied each of the seven actions (P1, P2, C1, S1, S2, S3, S4) once on each new object, in order to build initial feature observations. Then it actively combined multiple sensory feedbacks and transferred the prior tactile instance knowledge for each new object and exploratory action. When the robot iteratively learned the physical properties of new objects, it updated the prior tactile knowledge and the dependent GPC models with all feature observations collected hitherto. At each iteration, we measured the discrimination accuracy of the new objects' observation models to the test dataset. The transfer learning performance was compared with the learning process without prior knowledge which served as the baseline method, i.e. the robot can only actively collect feature observations.

We first evaluated ATIKT with ten groups of prior objects. Each group contains three objects that were randomly selected following the uniform distribution. We conducted the experiment five trials for each group. In each trial, the robot followed the transfer learning approach and no-transfer approach to collect $40$ feature observations. In this way, we had a fair comparison between different learning strategies. We used the model prediction-based approach to estimate object relatedness $\rho$. In Sec.~\ref{sec::method::where_to_transfer}, we also proposed model optimization approach, i.e. to obtain $\rho$ by maximizing the log-likelihood. However, in the experiment we found that following this strategy, the $\rho$ estimation was biased towards the feature observations of old objects, when there are only a few feature observations of new objects. As a result, the object relatedness from many old objects were "over-estimated" to be nearly $1$. Reusing an old object which should be less related to the new object degraded the learning performance. Therefore, in all sequential experiments, we only used model-prediction based approach. We further set the exploration rate $\epsilon_{\text{explore}}$ to be $0.3$, i.e. the robot randomly selected the object and exploratory action with a probability of $30\%$.

\begin{figure}[!htpb]
	\centering
	\begin{minipage}{1\linewidth}
		\centering
		\includegraphics[width=0.88\textwidth]{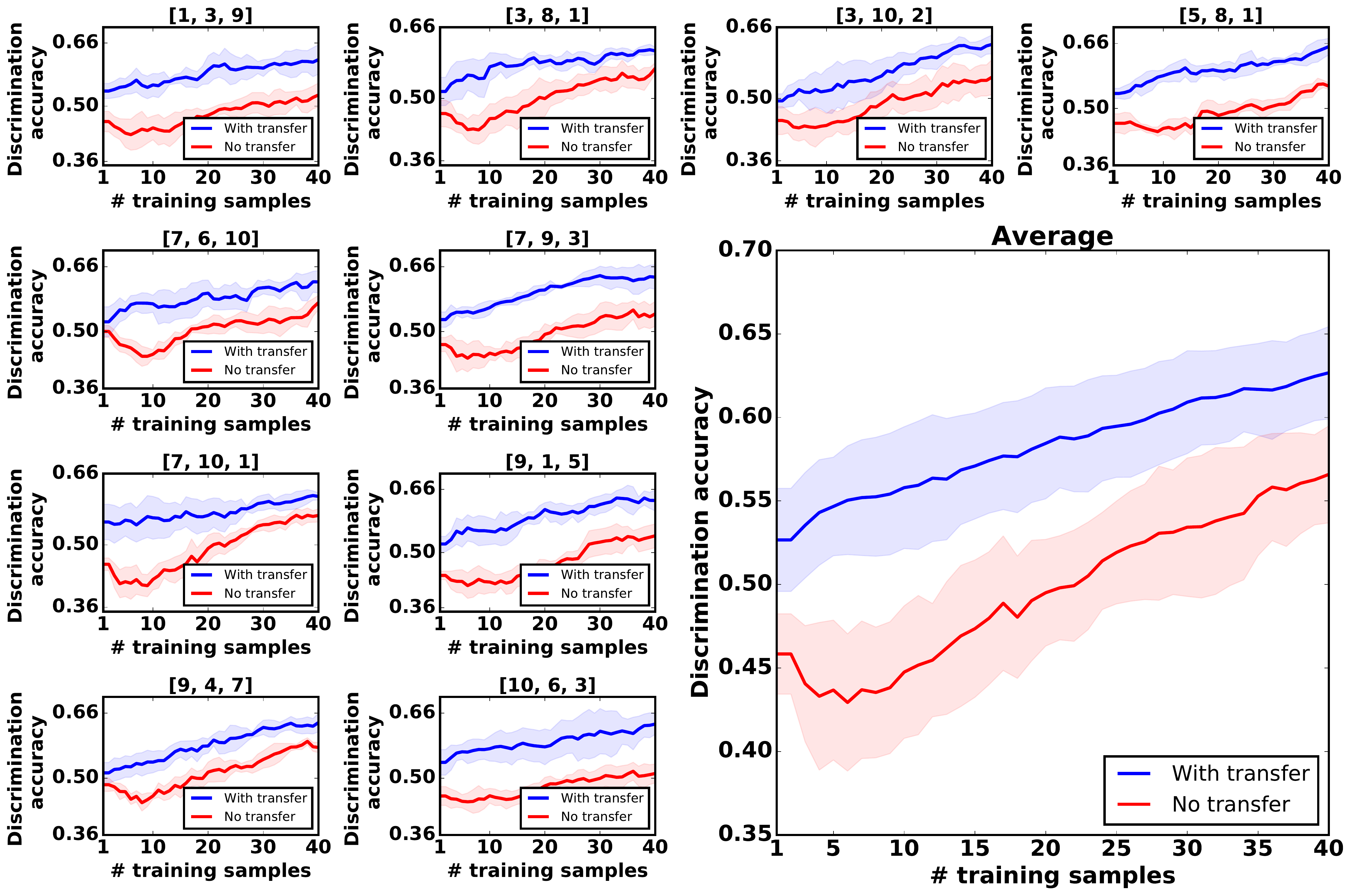}
	\end{minipage}
	\caption[Transfer learning performance with the prior tactile exploratory action experiences from three old objects.]{Transferring the prior tactile exploratory action experiences from three old objects to improve the process of learning the physical properties of five new objects. The transfer learning algorithm is compared with no transfer strategy. Small plots on the left show the learning curves with $10$ groups of prior tactile knowledge. Each group contains three old objects, whose object indices are illustrated in the title of each subplot. The figure on the bottom right shows the averaged learning performance. The horizontal axis represents the growing number of feature observations the robot has collected hitherto; The vertical axis represents the averaged value of discrimination accuracy of the test dataset. }\label{fig:TL_all}
\end{figure}

Fig.~\ref{fig:TL_all} illustrates the learning performance for each group of prior objects and the averaged performance. In all groups, the robot that used ATIKT consistently outperformed the learning process without prior knowledge. For instance, the robot achieved in average $8\%$ higher discrimination accuracy than the baseline method, when only \textit{one} new feature observation was collected, showing the one-shot learning behaviour (Fig.~\ref{fig:TL_all}). When the robot collected feature observations from $1$ to $40$, it achieved $65\%$ discrimination accuracy, by actively leveraging the prior tactile instance knowledge. Conversely, the robot without knowledge can discriminate among objects with an accuracy of only $55\%$.

\subsection{Learning About Objects via One Exploratory Action} 
\begin{figure}[H]
	\centering
	\begin{minipage}{1\linewidth}
		\centering
		\includegraphics[width=0.96\textwidth]{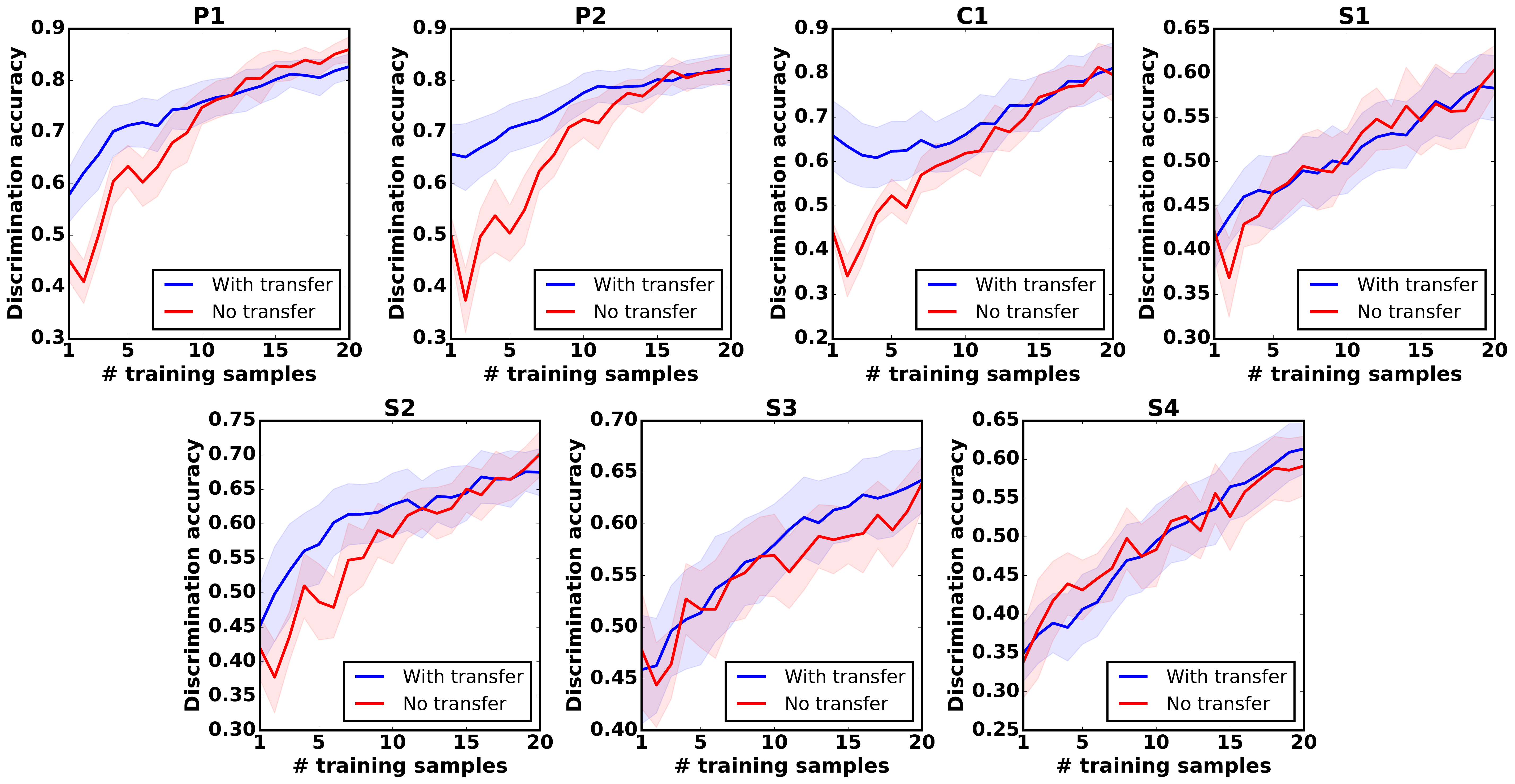}
	\end{minipage}
	\caption[Transfer learning using only one exploratory action.]{Transfer learning using only one exploratory action.}\label{fig:TL_separate}
\end{figure}

In order to further evaluate the robustness of our transfer learning algorithm, the robot was tasked to learn about objects via applying only \textit{one} of the exploratory actions. The experimental procedure was the same as the one described above. The results are shown in Fig.~\ref{fig:TL_separate}. As can be seen, by actions P1, P2 and C1, The robot had a larger improvement than actions S1, S2, S3 and S4. For example, the robot increased the discrimination accuracy by $25\%$, when it reused the prior tactile instance knowledge from the movement P2. However, when learning about objects by actions S1 and S4, little improvement was seen. This was due to the fact that different exploratory actions produced different object feature observations. For action P2, there existed higher related prior tactile knowledge than S1 and S4, and the robot could benefit more on it.

In all scenarios, using our proposed transfer learning algorithm, the robot could achieve a higher discrimination accuracy than the baseline method with the same number of feature observations. Therefore, we can conclude that the proposed method helps the robot build reliable observation models of new objects with fewer training samples, even when only one kind of the exploratory action is applied.

\subsection{Increasing the Number of Prior Objects}\label{subsec:experiment:TL_many_priors}
\begin{figure}[H]
	\centering
	\begin{minipage}{1\linewidth}
		\centering
		\includegraphics[width=1\textwidth]{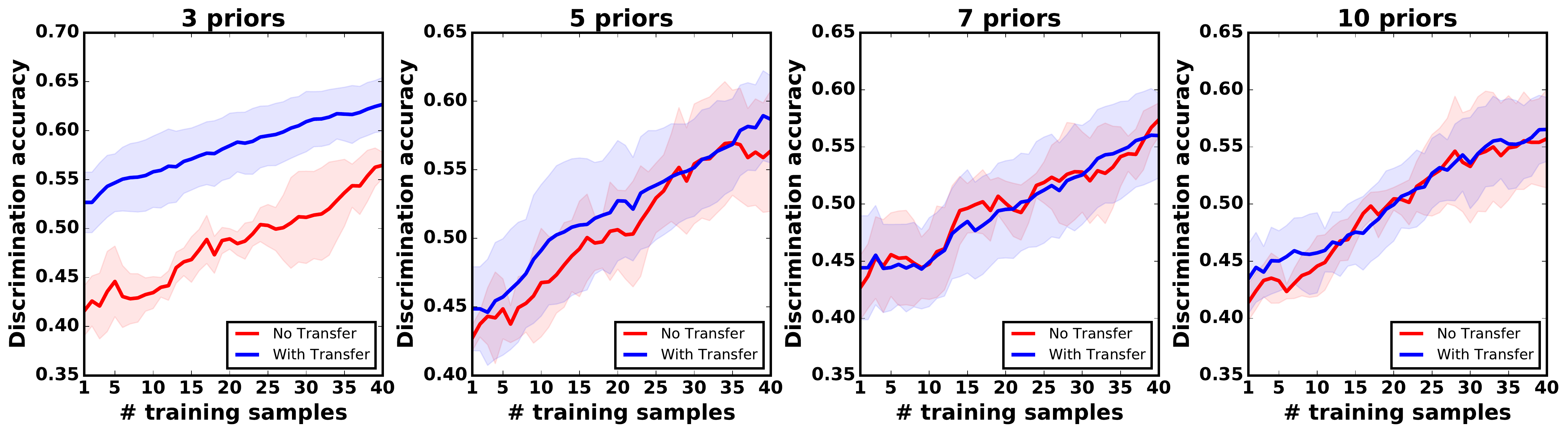}
	\end{minipage}
	\caption[Transfer learning performance with different number of prior objects for all exploratory actions.]{Evaluation of the transfer learning algorithm with different number of prior objects for all exploratory actions. }\label{fig:TL_all_comparison}
\end{figure}

\begin{figure}[!htp]
	\centering
	\begin{minipage}{1\linewidth}
		\centering
		\includegraphics[width=0.96\textwidth]{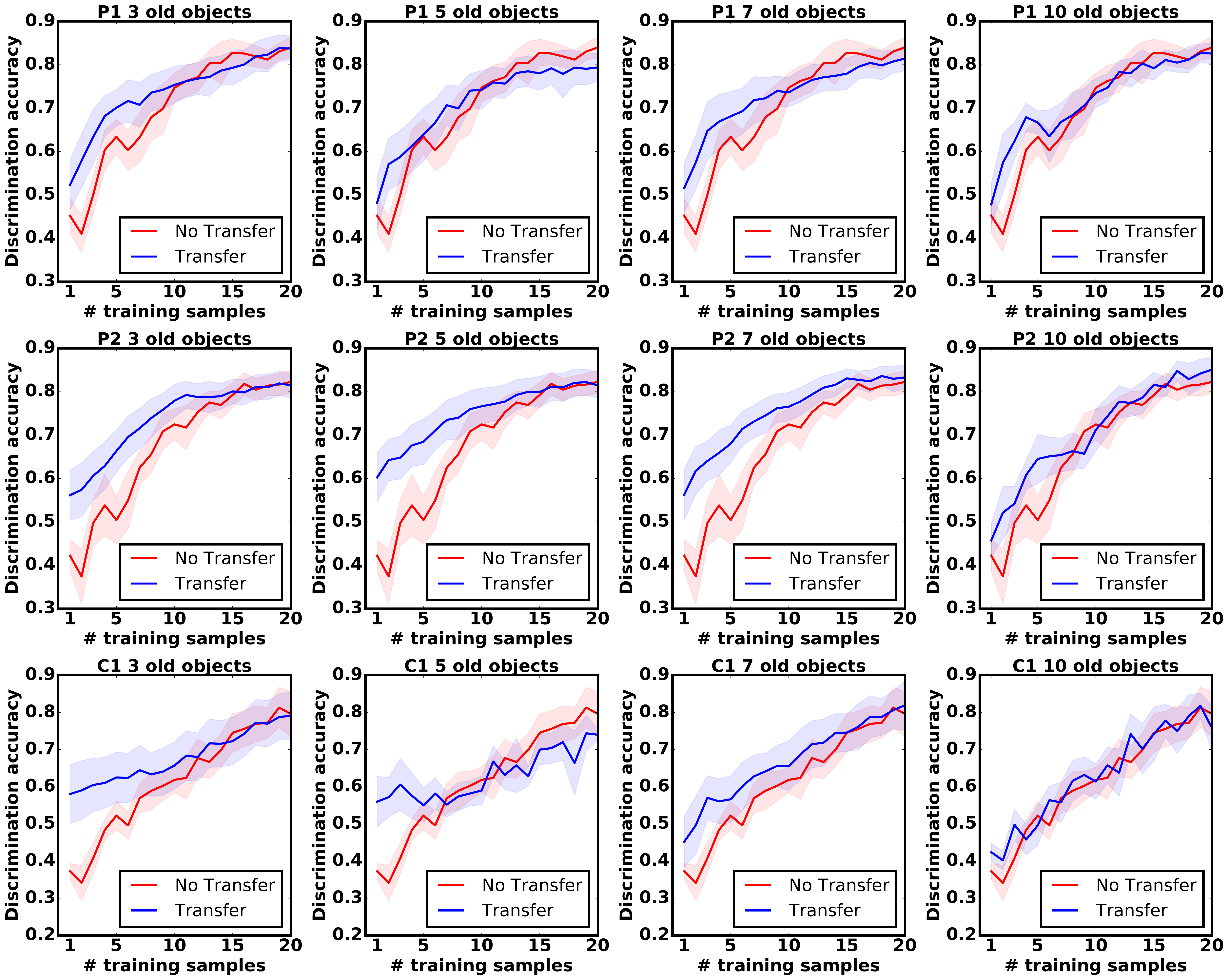}
	\end{minipage}
	\caption[Evaluation of transfer learning algorithm with different number of old objects for exploratory actions P1, P2, C1.]{Evaluation of transfer learning algorithm with different number of old objects for exploratory actions P1, P2, C1. Each row represents the learning process from an exploratory action: P1, P2, C1; Each column shows the number of old objects: 3, 5, 7, 10.}\label{fig:TL_results_separate1}
\end{figure}

\begin{figure}[!htp]
	\centering
	\begin{minipage}{1\linewidth}
		\centering
		\includegraphics[width=0.96\textwidth]{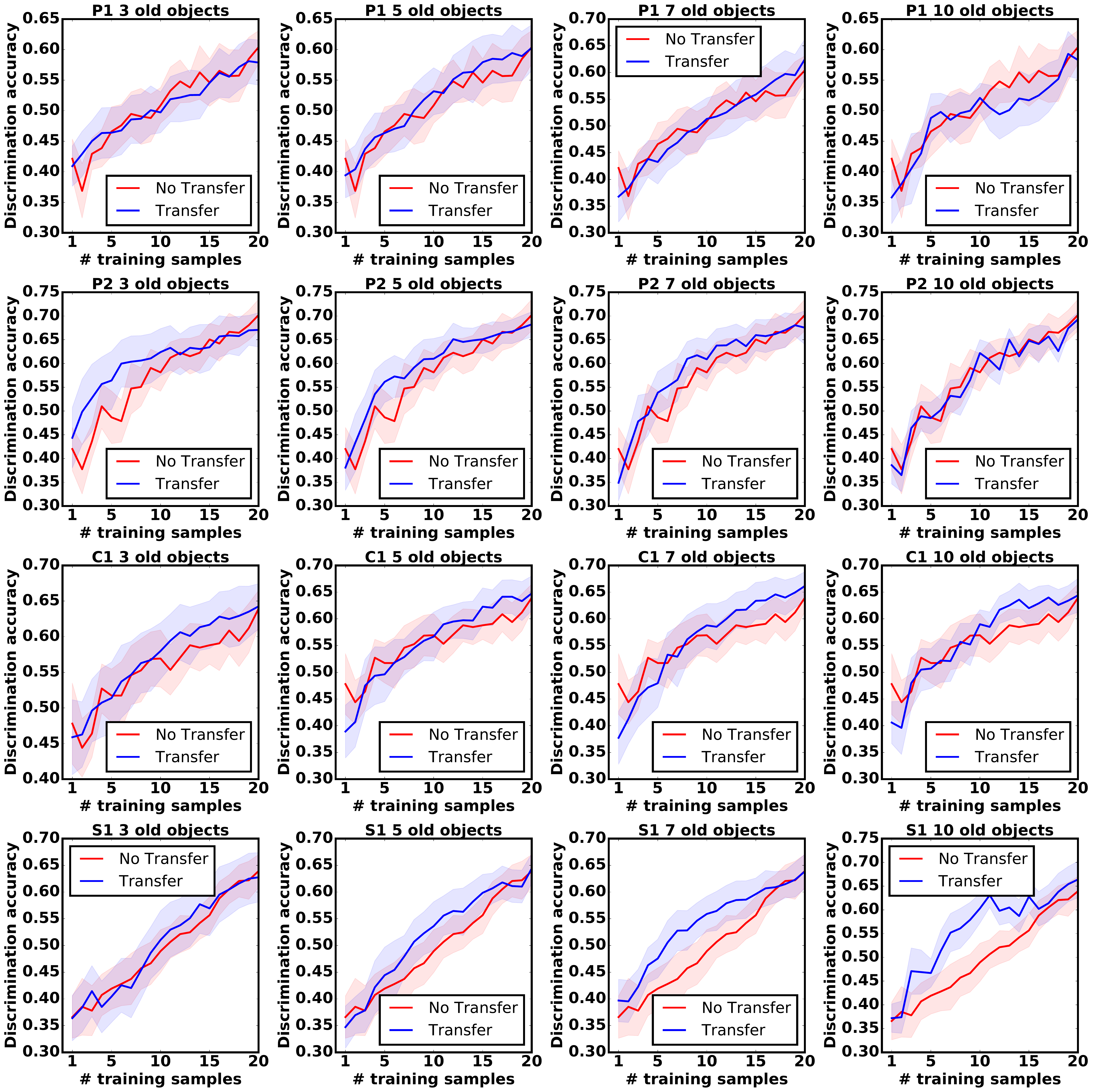}
	\end{minipage}
	\caption[Evaluation of transfer learning algorithm with different number of old objects for exploratory actions S1, S2, S3, S4.]{Evaluation of transfer learning algorithm with different number of old objects for exploratory actions S1, S2, S3, S4. Each row represents the learning process from an exploratory action: S1, S2, S3, S4; Each column shows the number of old objects, starting from 3, 5, 7, up to 10. The horizontal axis represents the growing number of training samples (feature observations) collected by the robotic system. The vertical axis shows the discrimination accuracy by the dependent GPC model to the test database.}\label{fig:TL_results_separate2}
\end{figure}

In this experiment, we evaluated the performance of our proposed transfer learning algorithm with an increasing number of prior objects. Intuitively, as the number of old objects grows, it is more likely that the robot can find highly-related prior tactile knowledge, and correspondingly the learning performance could continue to be improved. In this regard, following the same experimental procedure described in Sec.~\ref{subsec:experiment:TL_different_groups}, we randomly selected $5$ and $7$ old objects $10$ times and tested the performance when the robot learned about new objects via only one action or all seven actions. We also used all $10$ prior objects to conduct the experiment $5$ trials. The results are shown in Fig.~\ref{fig:TL_all_comparison} (for all exploratory actions), Fig.~\ref{fig:TL_results_separate1} (for exploratory action P1, P2 and C1), and Fig.~\ref{fig:TL_results_separate2} (for exploratory action S1, S2, S3, and S4).

Unexpectedly, except the results from action $S4$ (the last row of Fig.~\ref{fig:TL_results_separate2}), all results show that the transfer learning performance depreciated as the number of priors grew. As an example, using the prior tactile instance knowledge from $3$ old objects, the robot that applied action $C1$ improved the discrimination accuracy by $25\%$, whereas when there were $10$ old objects available, the transfer learning performed similar to the baseline method (the third row of Fig.~\ref{fig:TL_results_separate1}). Looking the learning process into detail, we found that the relatedness estimation $\rho$ tended to be smaller than $0.5$, as the number of old objects grew. Since this prediction was smaller than the threshold value $\epsilon_{neg_1}$, the robot stopped transferring knowledge. This phenomena was due to the fact that when training the old objects' GPC with a larger number of objects, the borders to discriminate among objects become sharper, making the objects "more dissimilar" to each other. This made it more difficult to find the related prior objects.

\section{Negative Prior Tactile Knowledge Transfer Testing}\label{sec:experiment:Negative_transfer}

\begin{figure}[h]
	\centering
	\begin{minipage}{1\linewidth}
		\centering
		\includegraphics[width=1\textwidth]{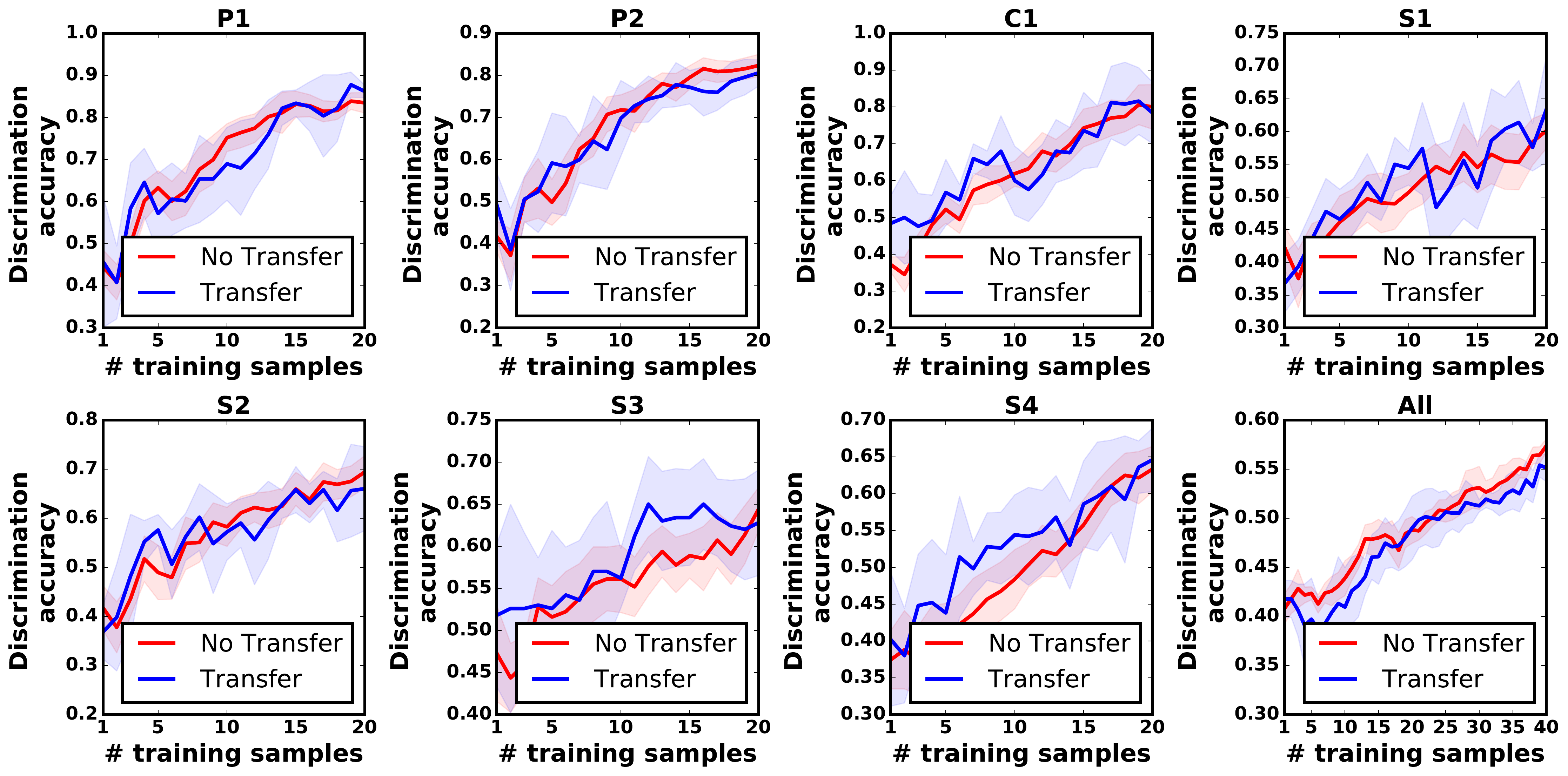}
	\end{minipage}
	\caption[Negative prior tactile knowledge transfer testing.]{Negative prior tactile knowledge transfer testing. The prior objects were deliberately selected that were unrelated to the new objects.}\label{fig:TL_negative}
\end{figure}
In the section related to the introduction of transfer learning (Sec.~\ref{sec:background:TL}), we mentioned that prior objects may not always relate to new objects. A brute-force transfer may degrade the learning performance, resulting in the negative knowledge transfer phenomena. In this case, a transfer learning algorithm should stop leveraging irrelevant prior knowledge.

\begin{figure}[!htpb]
	\centering
	\begin{minipage}{1\linewidth}
		\centering
		\includegraphics[width=0.7\textwidth]{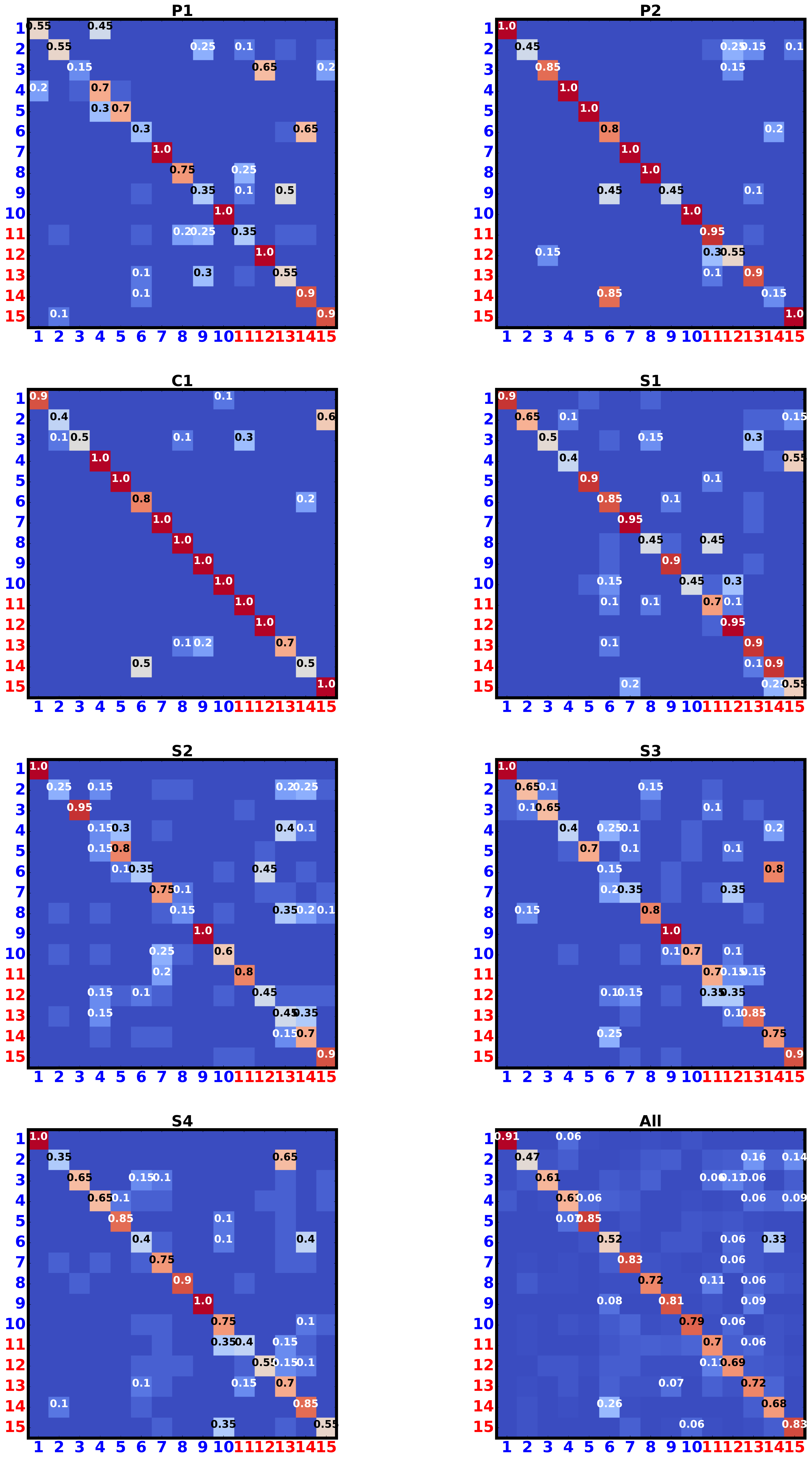}
	\end{minipage}
	\caption[Object confusion matrices.]{Single confusion matrix for each exploratory action (P1, P2, C1, S1, S2, S3, S4) and the averaged matrix. The value is normalized between $0-1$. Blue index indicates the objects that are selected as the prior objects. The new objects are reordered with indices starting from $11$ and are shown in red.}\label{fig:confusion_matrices}
\end{figure}

In order to evaluate ATIKT against the negative tactile knowledge transfer, we deliberately selected irrelevant prior objects and compared the transfer learning performance with the baseline method, following the same experiment process described in Sec.~\ref{subsec:experiment:TL_different_groups}. When finding which objects were relevant (or irrelevant) to each other, we built object confusion matrices to roughly evaluate the object similarity. To do this, for each of the seven exploratory actions, we trained a Gaussian Mixture Model (GMM) and calculated the object confusion matrix. We further calculated the confusion matrix averaged over all exploratory actions. The results are shown in Fig.~\ref{fig:confusion_matrices}. 

According to Fig.~\ref{fig:confusion_matrices}, objects \{$1$, $5$, $7$\} were dissimilar to the new objects (objects \{$11$ - $15$\}) regarding the exploratory movement P1, objects \{$1$, $4$, $7$\} for P2, objects \{$4$, $7$, $10$\} for C1, objects \{$1$, $6$, $9$\} for S1, objects \{$1$, $7$, $10$\} for S2, objects \{$1$, $3$, $9$\} for S3, and objects \{$1$, $3$, $8$\} for S4. We thus used these objects as prior objects to test the transfer learning performance via the single exploratory action. We further selected objects \{$1$, $5$, $10$\} to test the learning process via all exploratory actions, since these three objects shared relative small similarity to the new objects. 

The results in Fig.~\ref{fig:TL_negative} illustrate that the discrimination accuracy achieved by ATIKT was similar to the baseline method, when the robot applied either one or all seven exploratory actions. The results indicate that our proposed algorithm stopped transferring negative prior tactile instance knowledge.

\cleardoublepage
\let\textcircled=\pgftextcircled
\chapter{Summary and Discussion}\label{chapter:conclusion}
\initial{I}n this master thesis, we proposed a transfer learning method (Active Tactile Instance Knowledge Transfer: ATIKT) for a robot equipped with multi-modal artificial skin to actively reuse the prior tactile exploratory action experiences when learning about new objects. These action experiences consisted of the feature observations (prior tactile instance knowledge) and observation models of prior objects (prior tactile model knowledge). They were built when the robot applied different exploratory actions with different action parameters on objects. Using ATIKT, the robot combined the observations from multiple sensory modalities to build the prior tactile instance knowledge and used it to successfully improve the process of learning the detailed tactile physical properties of new objects.

As the tactile observations (or instance) that a robotic system can perceive are dependent on how it applies an exploratory action, the robot can build a detailed knowledge of the objects' physical properties by applying exploratory actions with various action parameters (thinking about pressing an object with different normal forces). In this master thesis, we considered two pressing, four sliding, and one static contact movements controlled by various action parameters. By leveraging the prior tactile exploratory action experiences from these seven exploratory actions, the robot was able to efficiently learn the objects' stiffness, surface textures, and thermal conductivity. 

The robot can receive several sensory feedbacks when it interacts with objects, due to the multi-model artificial skin. In our case, pressing and static contact movements produced normal force and temperature sensing; sliding movement produced vibro-tactile signals and temperature sensing. The multiple feature observations after applying an exploratory action were combined such that the robot could rely more on the informative observations and less on the uninformative ones. To do this, a general kernel was built using the linearly weighted combination of the basic kernel for each type of feature, with the weights determined by maximizing the marginal log-likelihood of the GPC observation models. Experimental evaluation showed that by combining different sensor modalities, the robot could increase the discrimination accuracy by $20\%$, compared with using only one sensor modality.

When the robot transferred the prior tactile instance knowledge, it iteratively selected the most relevant prior object, and used its knowledge to boost the new object learning process. This was achieved by introducing the dependent GPC that incorporated the feature observations of prior objects in the observation models of new objects. Experimental results showed that with the help of transfer learning, the robot consistently achieved $10\%$ higher discrimination accuracy than without transfer learning. With a small number or even one feature observation(s), the robot could improve the discrimination accuracy by $25\%$, showing the one-shot learning behaviour. The results also showed that our method could stop transferring irrelevant prior tactile knowledge which may degrade the learning performance. \\

\noindent \textbf{Future works.} There are several aspects of this thesis that could be extended:
\begin{itemize}
    \item \textit{More types of exploratory actions}. In this work, we considered pressing, sliding and static contact movements with different action parameters, resulting in seven exploratory actions. In the real world, we humans interact with objects using more exploratory actions, such as tapping, poking, lifting, etc. As a future work, a robot could be facilitated to perform more exploratory actions to learn more physical properties of an object, such as auditory properties, center of mass, etc.
    
    \item \textit{Combination of more complex kernels.} In this work, we used the basic RBF kernel for each type of sensor modality, and built a general kernel using linearly weighted basic kernels combination. However, since the feature vectors that describe different physical properties are also different (e.g. different dimensions and distributions), it would be more sophisticated to use a domain-specific kernel for each sensor modality. Furthermore, combining kernels in a more complex way (e.g. Hadamard product) could exploit more data information, and thus results in a higher discrimination accuracy. These are interesting topics that can be investigated in the future.
    
    \item \textit{Heterogeneous tactile knowledge transfer.} Here, we dealt with homogeneous knowledge transfer (see the introduction in Sec.~\ref{Sec:TL-methodology}). It could be a very interesting future work to extend our approach to heterogeneous knowledge transfer, where the source domains and the target domains are different. For example, the robot could transfer the prior tactile experiences between pressing and tapping movements (cross-exploratory-action knowledge transfer), or a robotic arm could transfer its prior tactile knowledge to a humanoid robot (cross-robot knowledge transfer). 
\end{itemize}
%
%
%
\backmatter
\bibliographystyle{ieeetr}
\refstepcounter{chapter}

\clearemptydoublepage
%
%
\end{document}